\documentclass[preprint,12pt]{elsarticle}

\usepackage{amssymb}

\usepackage{amsmath}

\usepackage{todonotes}
\usepackage{xcolor}
\usepackage{graphicx}

\usepackage{rotating}
\usepackage{tabularx}
\usepackage{amssymb}
\usepackage{pifont}
\usepackage{adjustbox}
\usepackage{caption}
\usepackage{booktabs}
\usepackage{subcaption}
\usepackage{url}

\usepackage{array} % 必须添加：启用 m{ } 列格式
\graphicspath{ {./image/} }

\usepackage[edges]{forest}
\usepackage{tikz} % Added for TikZ drawings

\usepackage{ragged2e} 
\usepackage{booktabs,makecell, multirow, tabularx}
\usepackage{threeparttable}

\usepackage{enumitem}
\newlist{titemize}{itemize}{1}
\setlist[titemize]{leftmargin=*, nosep, topsep=2pt, partopsep=2pt, parsep=2pt, itemsep=2pt}

\usepackage{float}  %设置图片浮动位置的宏包

\newcommand{\Color}{magenta}  % (236,0,140)
\newcommand{\cod}[1]{\textcolor{\Color}{#1}}

\usepackage{hyperref}
\hypersetup{
	colorlinks=true, %链接颜色
	linkcolor=red,  %内部链接
	filecolor=red,  %本地文档
	urlcolor=blue,  %网址链接
	pdftitle={ZHOU Heng},
	pdfpagemode=FullScreen,
}

\journal{ISPRS}

\begin{document}

\begin{frontmatter}

\title{
Remote Sensing Image Dehazing: A Systematic Review of Progress, Challenges, and Prospects
} %% Article title

\author[1,2]{Heng Zhou}
\author[1,2]{Xiaoxiong Liu}
\author[3]{Zhenxi Zhang\texorpdfstring{\corref{cor1}}{*}}
\author[1,2]{Jieheng Yun}
\author[4]{Chengyang Li}
\author[7]{Yunchu Yang}
\author[6]{Dongyi Xia}
\author[3]{Chunna Tian}
\author[1,2]{Xiao-Jun Wu}

\cortext[cor1]{Corresponding author. Email: zhangzhenxi@xidian.edu.cn (Zhenxi Zhang)}

\address[1]{School of Artificial Intelligence and Computer Science, Jiangnan University, Wuxi 214122, China}
\address[2]{Josef Kittler Research Institute on Artificial Intelligence, Jiangnan University, Wuxi 214122, China}
\address[3]{School of Electronic Engineering, Xidian University, Xi’an 710071, China}
\address[4]{College of Artificial Intelligence, China University of Petroleum (Beijing), Beijing 102249, China}
\address[7]{Aerospace Information Research Institute, Chinese Academy of Sciences, Beijing 100094, China}
\address[6]{Department of Computer Science, University of California, Santa Barbara, CA 93106, USA}

%% Abstract
\begin{abstract}
Remote sensing images (RSIs) are frequently degraded by haze, fog, and thin clouds, which obscure surface reflectance and hinder downstream applications. 
This study presents the first systematic and unified survey of RSIs dehazing, integrating methodological evolution, benchmark assessment, and physical consistency analysis. 
We categorize existing approaches into a three-stage progression: from handcrafted physical priors, to data-driven deep restoration, and finally to hybrid physical–intelligent generation, and summarize more than 30 representative methods across CNNs, GANs, Transformers, and diffusion models. 
To provide a reliable empirical reference, we conduct large-scale quantitative experiments on five public datasets using 12 metrics, including PSNR, SSIM, CIEDE, LPIPS, FID, SAM, ERGAS, UIQI, QNR, NIQE, and HIST. 
Cross-domain comparison reveals that recent Transformer- and diffusion-based models improve SSIM by 12\%–18\% and reduce perceptual errors by 20\%–35\% on average, while hybrid physics-guided designs achieve higher radiometric stability. 
A dedicated physical radiometric consistency experiment further demonstrates that models with explicit transmission or airlight constraints reduce color bias by up to 27\%. 
Based on these findings, we summarize open challenges: dynamic atmospheric modeling, multimodal fusion, lightweight deployment, data scarcity, and joint degradations, and outline promising research directions for future development of trustworthy, controllable, and efficient (TCE) dehazing systems.
All reviewed resources, including source code, benchmark datasets, evaluation metrics, and reproduction configurations are publicly available at 
\url{https://github.com/VisionVerse/RemoteSensing-Restoration-Survey}.

\end{abstract}

%% Keywords
\begin{keyword}
Remote sensing;
Image dehazing;
Image restoration;
Diffusion models;
Deep learning
%  \sep

\end{keyword}

\end{frontmatter}

%% main text

\section{Introduction}
\label{Sec.introduction}
Remote sensing satellite systems~\cite{gray2020remote,jiang2025combining,adebiyi2025fallowed} have evolved into sophisticated, multi-dimensional platforms that provide enhanced three-dimensional observation, broader spatiotemporal coverage, and increasingly intelligent data processing capabilities. These technological advancements underpin the extraction of rich geospatial information from remote sensing images (RSIs)~\cite{chen2023large,castellanos2024mineral}, driving significant progress across Earth observation, environmental monitoring, and resource exploration.
However, the utility of RSIs is frequently compromised by atmospheric phenomena such as haze, fog, and cloud cover, which obscure surface details and degrade the fidelity of subsequent analyses~\cite{juneja2022systematic,zhang2023perception}. 
Unlike traditional dehazing techniques that primarily address contrast and color distortions induced by aerosol scattering in natural scene images, 
RSIs dehazing targets the comprehensive restoration of Earth’s surface reflectance~\cite{agrawal2022comprehensive,xu2019thin}, demanding the removal of a broader range of atmospheric interferences, including haze, fog, and thin clouds~\cite{wang2023remote,chi2023trinity,pan2024hdrsa}.
As shown in~\tablename~\ref{tab:definition_scope}, this review focuses on remote sensing dehazing and thin-cloud removal associated with aerosol-dominated scattering.
RSIs dehazing task aims to address information attenuation rather than information loss and primarily relies on single-image restoration techniques. 
Opaque cloud occlusion and surface reconstruction are excluded from this work, and atmospheric correction is also beyond the scope.

\begin{table}[t]
\centering
\caption{
The difference analysis of haze, thin clouds, discussed in this review, and thick clouds, atmospheric correction.}
\label{tab:definition_scope}
\setlength{\tabcolsep}{2pt}
\renewcommand{\arraystretch}{1.25}
\resizebox{\textwidth}{!}{%
\begin{tabular}{
m{3.2cm}
@{\hspace{10pt}}   % ← 列间距（可调）
m{6.2cm}
@{\hspace{10pt}}   % ← 列间距（可调）
m{6.2cm}
}
\toprule
\textbf{Term} & \textbf{Physical Cause} & \textbf{Relation to This Review} \\
\midrule
\textbf{Haze} &
Aerosol scattering and absorption veil surface radiance and reduce contrast. &
\textbf{Core focus}. This survey fully covers haze as the primary visibility degradation. \\[4pt]
\textbf{Thin Cloud} &
Weak scattering by small droplets forms a semi-transparent layer. &
\textbf{Included}. Treated as haze-like when optical thickness is low. \\[4pt]  \midrule
\textbf{Thick Cloud, Cloud Shadows} &
Strong occlusion from dense clouds or shadows causes irreversible radiance loss. &
\textbf{Excluded}. Belongs to cloud removal rather than dehazing. \\[4pt]
\textbf{Atmospheric Correction} &
Gas absorption and Rayleigh or Mie scattering alter top-of-atmosphere radiance. &
\textbf{Related but not covered}. A radiometric calibration process rather than visibility restoration. \\
\bottomrule
\end{tabular}
}
\end{table}

\begin{figure}[t]
	\centering 	
	\includegraphics[scale=0.6]{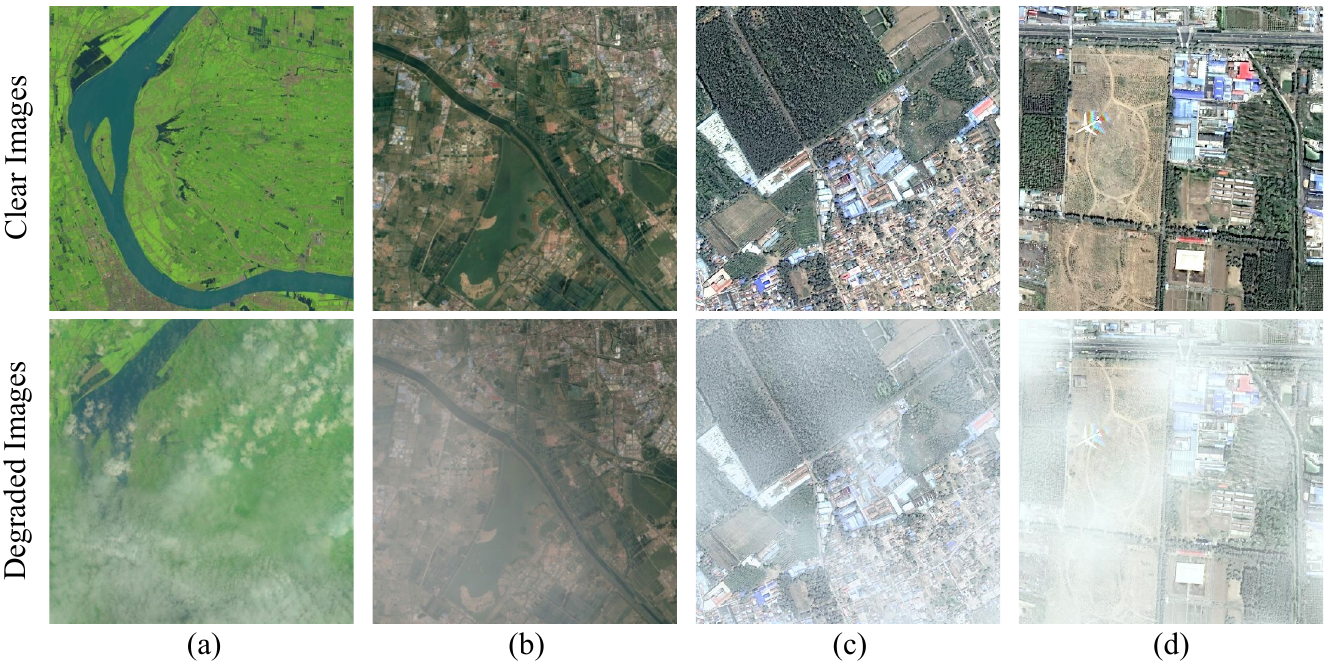}
	\caption{
		Illustrative scenarios for RSIs dehazing. 
        Each column (a)–(d) shows a different land-cover type from riverine–agricultural to industrial–urban areas, with haze density increasing from light (a) to heavy (d). 
	}
	\label{Fig.dehaze_example}
\end{figure}

The atmospheric scattering model is widely used in RS imaging to characterize the physical processes underlying the formation of degraded observations~\cite{tan2008visibility,bian2024enhancing,lihe2024phdnet}. 
Mathematically, the observed hazy image $\boldsymbol{I}_{hazy}$ at pixel location $p$ is modeled as:
\begin{equation}
	\boldsymbol{I}_{{hazy}}(p) = 
	\underbrace{\boldsymbol{I}_{{clear}}(p) \cdot \boldsymbol{t}(p)}_{{\text{Direct Attenuation}}}
	+ 
	\underbrace{\boldsymbol{A}(1 - \boldsymbol{t}(p))}_{{\text{Airlight}}}
\end{equation}
where $\boldsymbol{I}_{clear}$ denotes the clear reference image (scene radiance), $\boldsymbol{A}$ is the global atmospheric light, and $\boldsymbol{t}(p) = e^{-\boldsymbol{\beta} d(p)}$ represents the transmittance describing the fraction of unscattered light reaching the sensor~\cite{he2010single}. Here, $\boldsymbol{\beta}$ is the atmospheric scattering coefficient and $d(p)$ is the scene depth, such that radiance decays exponentially with increasing distance.
The direct attenuation $\boldsymbol{I}_{clear}(p)\cdot\boldsymbol{t}(p)$~\cite{tan2008visibility} models the decay of scene radiance as it passes through the scattering medium. 
The airlight $\boldsymbol{A} (1 - \boldsymbol{t}(p))$~\cite{tan2008visibility} accounts for the scattered atmospheric light that adds a veiling effect and alters color perception. 

As shown in \figurename~\ref{Fig.dehaze_example}, (a)\--{}(d) exhibit increasing levels of atmospheric degradation, where haze and thin clouds gradually obscure ground details and reduce contrast. 
In contrast, the top row shows their clear counterparts with sharp textures and accurate surface information.
By eliminating degradation artifacts and recovering structural details, RSIs restoration provides high-quality data foundations for subsequent computer vision applications, such as 
RS object detection~\cite{li2023detection,duan2024mdcnet,fu2024s,pan2025locate}, 
segmentation~\cite{wang2022unetformer,cheng2024methods,kang2024fusion}, and 
change detection~\cite{cheng2024harmony,li2024stade,zhou2025exploring}.

\tikzstyle{my-box}=[
rectangle,
draw=gray!50,
rounded corners,
text opacity=1,
minimum height=2em, % 小幅增加
minimum width=5em,
inner sep=3pt,
inner ysep=6pt, % 重点：上下边距更大
align=center,
fill opacity=0.15,
line width=0.5pt,
]

\tikzstyle{leaf}=[my-box, minimum height=2em,
fill=gray!5, text=black, align=left, font=\normalsize,
inner xsep=3pt,
inner ysep=6pt, % 叶子节点也要同步
line width=0.5pt,
]

\definecolor{c1}{RGB}{102,178,255} % fresh light blue
\definecolor{c2}{RGB}{255,153,153} % soft coral red
\definecolor{c3}{RGB}{255,204,102} % mellow orange
\definecolor{c4}{RGB}{153,221,153} % soft mint green
\definecolor{c5}{RGB}{204,179,255} % soft lavender purple
\definecolor{c7}{RGB}{153,221,214} % soft turquoise (更柔和)
\definecolor{c8}{RGB}{221,160,221} % plum purple
\definecolor{c9}{RGB}{255,179,207} % pastel pink

% \vspace*{-15mm}%尝试上移
%!pt
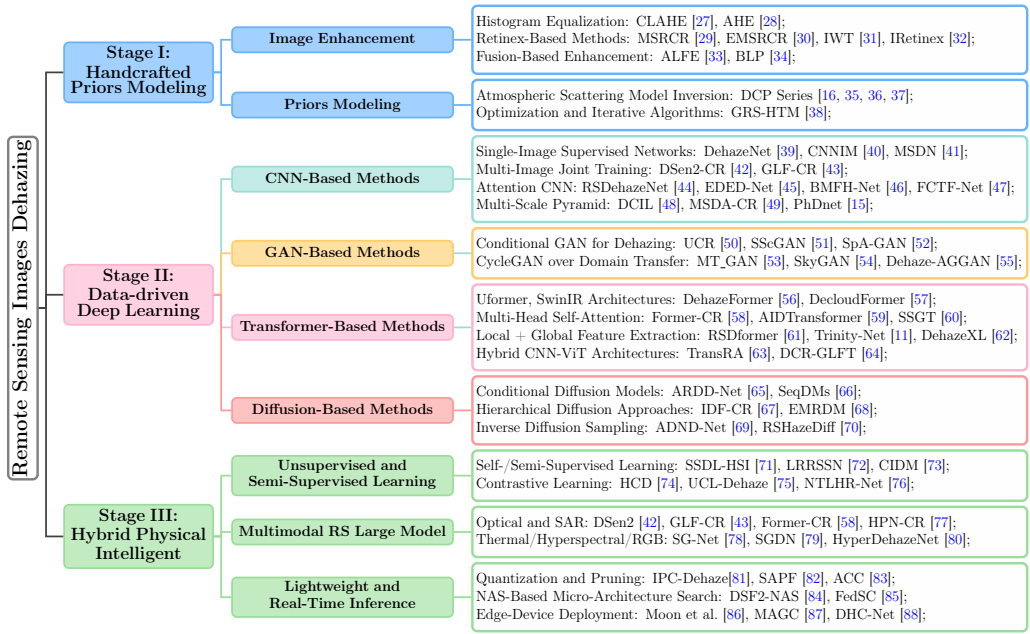
\begin{figure*}[t]
	\centering
	\resizebox{\textwidth}{!}{
		\begin{forest}
			forked edges,
			for tree={
				grow=east,
				reversed=true,
				anchor=base west,
				parent anchor=east,
				child anchor=west,
				base=center,
				font=\large,
				rectangle,
				% draw=hidden-draw,
				draw=gray,
				rounded corners,
				align=left,
				text centered,
				minimum width=4em,
				edge+={darkgray, line width=0.5mm},
				s sep=3pt,
				inner xsep=2pt,
				inner ysep=3pt,
				line width=0.8pt,
				ver/.style={rotate=90, child anchor=north, parent anchor=south, anchor=center},
			},
			where level=1{text width=10em,font=\normalsize,}{},
			where level=2{text width=15em,font=\normalsize,}{},
			where level=3{text width=10em,font=\normalsize,}{},
			where level=4{text width=20em,font=\normalsize,}{}, %，从35em改为38em
			where level=5{text width=10em,font=\normalsize,}{},
			[
			\Large \textbf{Remote Sensing Images Dehazing}, ver, line width=0.7mm
			[
			\large \shortstack{\textbf{Stage I:} \\ \textbf{Handcrafted} \\ \textbf{Priors Modeling}}, fill=c1!60, draw=c1, line width=0.5mm
			[
			\shortstack{\textbf{Image Enhancement}}, 
			fill=c1!60, draw=c1, line width=0.5mm, edge={c1}
			[
			Histogram Equalization{: CLAHE~\cite{5301485}, AHE~\cite{thanh2019single}};\\
			Retinex-Based Methods{: MSRCR~\cite{xue2016video}, EMSRCR~\cite{gao2014enhancement}, IWT~\cite{rong2014improved}, IRetinex~\cite{fan2017improved}};\\
			Fusion-Based Enhancement{: ALFE~\cite{liang2024remote}, BLP~\cite{shen2014exposure}};
			, leaf, text width=38em, draw=c1, line width=0.7mm, edge={c1} %
			]
			]
			[
			\shortstack{\textbf{Priors Modeling}}, 
			fill=c1!60, draw=c1, line width=0.5mm, edge={c1}
			[
			Atmospheric Scattering Model Inversion{: DCP Series~\cite{he2010single,yuan2017effective,li2018haze,shi2021novel}};\\
			Optimization and Iterative Algorithms{: GRS-HTM~\cite{liu2017haze}};
			, leaf, text width=38em, draw=c1, line width=0.7mm, edge={c1} %
			]
			]
			]% level1
			[   
			\large \shortstack{\textbf{Stage II:} \\ \textbf{Data-driven} \\ \textbf{Deep Learning}}, fill=c9!60, draw=c9, line width=0.5mm
			[
			\textbf{CNN-Based Methods}, align=center, fill=c7!60, draw=c7, line width=0.5mm, edge={c7}
			[
			Single-Image Supervised Networks{: DehazeNet~\cite{cai2016dehazenet}, CNNIM~\cite{zi2021thin}, MSDN~\cite{ren2016single}};\\
			Multi-Image Joint Training{: DSen2-CR~\cite{meraner2020cloud}, GLF-CR~\cite{xu2022glf}};\\
			Attention CNN{: RSDehazeNet~\cite{guo2020rsdehazenet}, EDED-Net~\cite{dong2024end}, BMFH-Net~\cite{sun2025bidirectional}, 
            % ICL-Net~\cite{dong2024icl}
            FCTF-Net~\cite{li2020coarse}}; \\
			Multi-Scale Pyramid{: DCIL~\cite{zhang2022dense}, MSDA-CR~\cite{yu2022cloud}, PhDnet~\cite{lihe2024phdnet}};
			,leaf, text width=38em, draw=c7, line width=0.7mm, edge={c7} %
			]
			]
			[
			\textbf{GAN-Based Methods}, fill=c3!60, draw=c3, line width=0.5mm, edge={c3}
			[
			Conditional GAN for Dehazing{: 
            UCR~\cite{zheng2020single}, SScGAN~\cite{huang2020single}, 
            SpA-GAN~\cite{pan2020cloud}}; \\
			CycleGAN over Domain Transfer{:
            MT\underline{~}GAN~\cite{wang2025mt_gan},
             SkyGAN~\cite{mehta2021domain},  
            Dehaze-AGGAN~\cite{zheng2022dehaze}};
			,leaf, text width=38em, draw=c3, line width=0.7mm, edge={c7} %
			]
			]
			[
			\shortstack{\textbf{Transformer-Based Methods}}
			, fill=c9!60, draw=c9, line width=0.5mm, edge={c9}
			[
			{Uformer, SwinIR Architectures}{: DehazeFormer~\cite{song2023vision}, DecloudFormer~\cite{li2025decloudformer}};\\
			Multi-Head Self-Attention{: Former-CR~\cite{han2023former}, AIDTransformer~\cite{kulkarni2023aerial}, 
            SSGT~\cite{du2024ssgt}};\\
	            Local + Global Feature Extraction{: RSDformer~\cite{song2023learning}, Trinity-Net~\cite{chi2023trinity}, DehazeXL~\cite{chen2025tokenize}};\\
	            Hybrid CNN-ViT Architectures{: TransRA~\cite{dong2022transra}, DCR-GLFT~\cite{quan2024density}};
			,leaf, text width=38em, draw=c9, line width=0.7mm, edge={c7} %
			]
			]
			[
			\shortstack{\textbf{Diffusion-Based Methods}}
			, fill=c2!60, draw=c2, line width=0.5mm, edge={c2}
			[
			Conditional Diffusion Models{: ARDD-Net~\cite{huang2023remote}, SeqDMs~\cite{zhao2023cloud}};\\
	            Hierarchical Diffusion Approaches{: IDF-CR~\cite{wang2024idf}, EMRDM~\cite{liu2025effective}};\\
	            Inverse Diffusion Sampling{: ADND-Net~\cite{huang2024diffusion}, RSHazeDiff~\cite{xiong2024rshazediff}};
			,leaf, text width=38em, draw=c2, line width=0.7mm, edge={c2} %
			]
			]
			] 
			[   
			\large \shortstack{\textbf{Stage III:} \\ \textbf{Hybrid Physical} \\ \textbf{Intelligent}}
			, fill=c4!60, draw=c4, line width=0.5mm
			[
			\shortstack{\textbf{Unsupervised and} \\ \textbf{Semi-Supervised Learning}}, 
			align=center, fill=c4!60, draw=c4, line width=0.5mm, edge={c4}                   
			[
			Self-/Semi-Supervised Learning{: SSDL-HSI~\cite{li2024supervise},
             LRRSSN~\cite{chen2024thick},
             CIDM~\cite{zhang2025cidm}};\\
			Contrastive Learning{: HCD~\cite{wang2024restoring},
            UCL-Dehaze~\cite{wang2024ucl},
            NTLHR-Net~\cite{ma2025nighttime}};\\
			,leaf, text width=38em, draw=c4, line width=0.7mm, edge={c4} %
			]
			]
			[
			\textbf{Multimodal RS Large Model}, fill=c4!60, draw=c4, line width=0.5mm, edge={c4}
			[
			Optical and SAR{:  
            DSen2~\cite{meraner2020cloud}, GLF-CR~\cite{xu2022glf}, Former-CR~\cite{han2023former}, HPN-CR~\cite{gu2025hpn}};\\
			{Thermal/Hyperspectral/RGB}{: SG-Net~\cite{ma2022spectral},
            SGDN~\cite{fang2025guided},
            HyperDehazeNet~\cite{fu2024hyperdehazing}};\\
			,leaf, text width=38em, draw=c4, line width=0.7mm, edge={c4} %
			]
			]
			[
			\shortstack{\textbf{Lightweight and} \\ \textbf{Real-Time Inference}}, fill=c4!60, draw=c4, line width=0.5mm, edge={c4}
			[
			Quantization and Pruning{: IPC-Dehaze\cite{fu2025iterative},
            SAPF~\cite{hu2025scale},
            ACC~\cite{li2025dual}};\\
			NAS-Based Micro-Architecture Search{: DSF2-NAS~\cite{feng2025dsf2},
            FedSC~\cite{li2025towards}};\\
			Edge-Device Deployment{: Moon et al.~\cite{moon2025development},
            MAGC~\cite{ye2025map},
            DHC-Net~\cite{li2025dhc}};
			,leaf, text width=38em, draw=c4, line width=0.7mm, edge={c4} %
			]
			]
			] 
			]
		\end{forest}
	}
	% \vspace{-0mm}
	\caption{Taxonomy of Remote Sensing Image Dehazing Methods.}
	% \label{fig:taxonomy}
	\label{Fig.Framwork}
	% \vspace{0mm}
\end{figure*}

\figurename~\ref{Fig.Framwork} provides a systematic illustration of representative frameworks categorized into traditional and deep learning-based approaches for RSIs dehazing.
Early methodologies predominantly leveraged physical priors~\cite{long2013single,wang2017single} and statistical analysis techniques~\cite{wang2017fast,bui2017single,makarau2014haze}. 
The image enhancement strategies, including contrast stretching and Retinex-based illumination reflectance separation, were frequently employed~\cite{ju2017single,ju2021ide}.
Accurately decoupling multiple degradation factors and adapting to diverse scenes remains challenging for physically consistent, high-quality restoration, motivating a shift from explicit physics-based models to data-driven learning frameworks.
This transition marks \textbf{the first stage} in the methodological evolution of RSIs dehazing: from handcrafted physical priors to adaptive representation learning.

Conversely, deep learning techniques harness semantic priors and contextual features to model complex atmospheric degradation through data-driven approaches~\cite{zhou2023position,yin2025cssf}. 
Initially, classical convolutional neural network (CNN) architectures,
such as AlexNet~\cite{krizhevsky2012imagenet}, VGGNet~\cite{simonyan2014very}, and ResNet~\cite{he2016deep},
were adapted to RS dehazing tasks. 
While CNN-based models achieved promising restoration accuracy, their local receptive fields and dependence on paired datasets limited robustness under real-world atmospheric variations~\cite{dong2020multi}. 
This limitation prompted the emergence of generative paradigms such as GANs~\cite{singh2018cloud}, which introduced adversarial supervision to enhance perceptual realism and compensate for the oversmoothing of convolutional outputs.
However, GAN-based frameworks often suffer from training instability and lack of explicit physical constraints, producing visually convincing but physically inconsistent results.
To overcome these shortcomings, Vision Transformers (ViTs)~\cite{dosovitskiy2020image} were adopted to capture global dependencies and semantic consistency through self-attention mechanisms, bridging local detail recovery and large-scale contextual reasoning.
More recently, diffusion-based generative models~\cite{ho2020denoising,huang2024stfdiff,zhong2024ssdiff} have extended this trajectory by introducing stochastic denoising processes capable of reconstructing fine-grained structures~\cite{liu2024diffusion,sui2024diffusion}.
Extensive computational demands and iterative inference processes currently constrain their practical applicability.
These limitations have driven an evolutionary shift from deterministic representation learning toward generative restoration frameworks. 
Early CNN- and GAN-based models improved perceptual realism and nonlinear feature modeling, yet their dependence on large paired datasets, weak physical constraints, and training instability restricted their scalability and physical fidelity. 
The introduction of Transformer and Diffusion models further extended this paradigm by integrating global attention and probabilistic generation, enabling more consistent scene-level reasoning but at the cost of high computational complexity. 
This transition marks \textbf{the second stage} in the methodological evolution of RS image dehazing: from adaptive representation learning to generative restoration.

Contemporary research increasingly emphasizes hybrid model-physics frameworks~\cite{chen2024social,cui2024physics,lihe2024phdnet,liang2024remote,zhang2025hydrogen}, multi-source data fusion strategies~\cite{narayanan2023multi,zhou2024frequency,jia2025airborne}, and lightweight architectural designs~\cite{ullah2021light,jin2023lfd,ma2024deepcache,song2024lightweight}.
Integrating atmospheric physical priors with deep learning has enhanced interpretability and physical realism in restoration outcomes. Simultaneously, attention-driven fusion methods~\cite{li2024m2fnet,wang2025gat} significantly enhance the preservation of spectral and structural integrity within multimodal datasets. 
Given the reliance on extensive degraded-clear image pairs for deep learning models~\cite{liu2024boosting,wang2025deep,zhou2025deformation}, there is a growing emphasis on meticulous datasets creation, particularly ensuring diversity in degradation patterns, comprehensive spatial coverage, and fidelity to physical atmospheric conditions.
In the latest research, by embedding atmospheric priors into neural and generative frameworks, these hybrid paradigms enhance physical realism and spectral fidelity while alleviating data dependence through cross-modal learning and synthetic simulation.
As shown in \figurename~\ref{Fig.Framwork}, this transition marks \textbf{the third stage} in the methodological evolution of RSIs dehazing: from generative restoration to hybrid physical–intelligent integration.

Consequently, the main contributions of this review are structured as follows:
\begin{itemize}
\item \textbf{A unified evolutionary framework.}
We establish, for the first time, a stage-wise taxonomy that traces the methodological progression of RSIs dehazing from handcrafted physical priors to deep data-driven restoration and, more recently, to hybrid physical–intelligent paradigms. 
Within this framework, six representative methodological families: image enhancement, atmospheric modeling, convolutional restoration, adversarial generation, Transformer architectures, and diffusion models, are systematically analyzed in terms of their theoretical foundations and technical characteristics.
\\ 
\item \textbf{Comprehensive methodological assessment.}
A critical examination is conducted for each methodological class, covering modeling assumptions, architectural properties, restoration behavior under heterogeneous degradation, physical interpretability, generalization capacity, and computational considerations. 
This analysis provides a balanced understanding of the strengths and limitations across different generations of techniques.
\\
\item \textbf{Large-scale benchmark evaluation.}
More than 30 representative methods are quantitatively compared on five widely used datasets using twelve full-reference, perceptual, and radiometric metrics. 
The results offer a transparent and reproducible benchmark, revealing consistent performance gains of recent Transformer- and diffusion-based approaches and highlighting the radiometric advantages of physics-constrained designs.
\\
\item \textbf{Radiometric consistency investigation.}
Beyond standard benchmarks, we introduce the most comprehensive physical-consistency evaluation to examine atmospheric-parameter coherence. The analysis shows that methods incorporating explicit transmittance or airlight constraints achieve notably improved spectral and color stability, providing additional insights into the physical reliability of advanced models.
\\
\item \textbf{Clarification of research challenges and directions.}
Drawing from methodological and empirical evidence, we summarize five open challenges—dynamic atmospheric variability, multimodal fusion, lightweight deployment, data scarcity, and coupled degradations—and outline promising avenues toward more trustworthy, controllable, and efficient remote sensing dehazing systems.
\\
\item \textbf{Open-source resource consolidation.}
To support reproducibility and community development, all surveyed methods, datasets, evaluation scripts, and configuration files are compiled and made publicly accessible through a dedicated \href{https://github.com/VisionVerse/RemoteSensing-Restoration-Survey}{Github}.
\end{itemize}
% \href{https://github.com/VisionVerse/RemoteSensing-Restoration-Survey}{Github \emoji{star2}${150+}$}.

The remainder of this review is organized as follows. Section~\ref{Sec.Literature} analyzes the bibliometric trends and evolution of remote sensing image dehazing. Section~\ref{Sec.CurrentAdvances} reviews methodological advances through a stage-wise taxonomy spanning handcrafted priors, deep learning, and hybrid physical–intelligent models. Section~\ref{sec.Datasets} summarizes benchmark datasets and evaluation metrics. Section~\ref{sec.benchmark} presents a comprehensive benchmark analysis, including quantitative and qualitative evaluations, efficiency, radiometric consistency, generalization, and failure cases. Section~\ref{sec.challenges} discusses open challenges and emerging directions, and Section~\ref{sec.conclusion} concludes the review with key insights and future perspectives.

\begin{figure}[t]
	\centering 
	\includegraphics[scale=0.5]{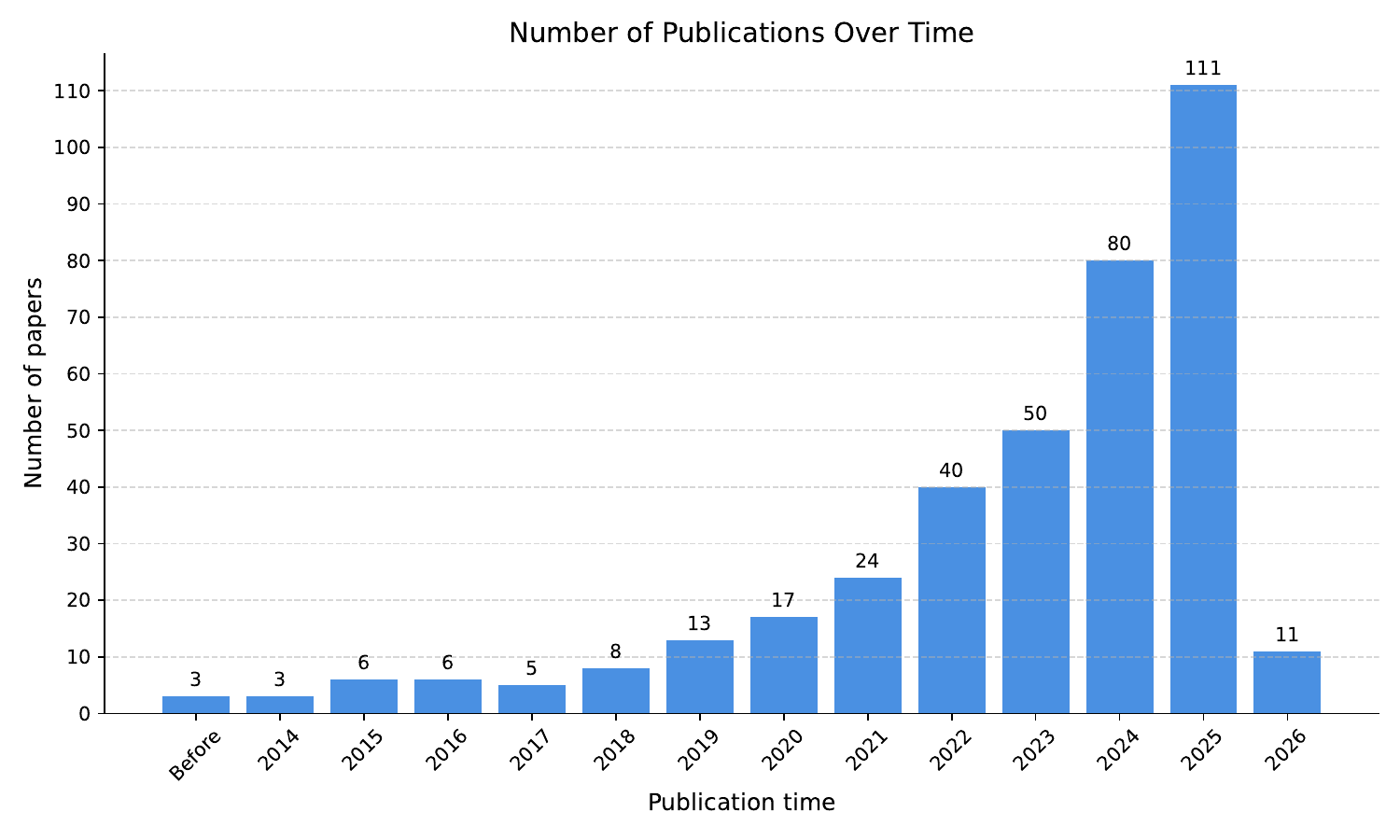}
	\caption{Number of papers published in RSIs dehazing up to January 2026.}
	\label{Fig.Number of paper}
\end{figure}

\begin{table}[t]
\centering
\caption{Impact Factor (IF2025), h5-index, and Paper Count$^\dagger$ of Selected Journals and Conferences.}

\label{tab:journal_if}
\scriptsize
\renewcommand{\arraystretch}{1.5} % 增加行距
\setlength{\tabcolsep}{1mm}       % 控制列间距
\begin{tabular}{p{9.1cm}>
{\centering\arraybackslash}p{0.9cm}>{\centering\arraybackslash}p{0.9cm}>{\centering\arraybackslash}p{1cm}}
\toprule
\textbf{Publication} & $\mathbf{IF_{2025}}$ & \textbf{h5} & \textbf{Num.} \\
\midrule
IEEE Trans. on Pattern Analysis and Machine Intelligence & 18.6 & 196 & {6} \\
Information Fusion & 15.5 & 134 & {6} \\
IEEE Trans. on Image Processing & 13.7 & 150 & {16} \\
ISPRS J. of Photogrammetry and Remote Sensing & 12.2 & 104 & 12 \\
IEEE Trans. on Circuits and Systems for Video Technology & 11.1 & 94 & {11} \\
IEEE Trans. on Geoscience and Remote Sensing  & 8.6 & 141 & {99} \\
IEEE Trans. on Intelligent Transportation Systems & 8.4 & 143 & {3} \\
Pattern Recognition & 7.6 & 118 & {6} \\
IEEE J. of Sel. Topics in Appl. Earth Obs. and Remote Sens. & 5.3 & 80 & {35} \\
IEEE Geoscience and Remote Sensing Letters  & 4.4 & 83 & {46} \\
Remote Sensing & 4.1 & 159 & {60} \\ 
Signal Processing & 3.6 & 69 & {7} \\ \midrule
IEEE/CVF Conf. on Computer Vision and Pattern Recognition & -- & 440 & {16}\\
European Conf. on Computer Vision & -- & 206 & 1 \\
IEEE/CVF Winter Conf. on Applications of Computer Vision  & -- & 109 & 4 \\
\bottomrule
\multicolumn{3}{l}{\scriptsize $\dagger$ Data as of January 2026.}\\
\end{tabular}

\end{table}

% 
% {\small 这是小号字体}。
% \tiny：极小的字体
% \scriptsize：比脚注更小的字体
% \footnotesize：脚注大小的字体
% \normalsize：正常字体大小
% \large：较大的字体 
% 

\section{Literature {and Existing Reviews} Analysis }
\label{Sec.Literature}

\subsection{Literature Analysis}
As shown in \figurename~\ref{Fig.Number of paper}, we employ {Web of Science (WoS\footnote{\url{https://support.clarivate.com/ScientificandAcademicResearch/s/article/}})} and {Database Systems and Logic Programming (DBLP\footnote{\url{https://dblp.org/}})} as the primary analytical tools.
The literature analysis was conducted through systematic retrieval using key terms:
remote sensing dehazing, haze removal, fog removal, thin-cloud removal, atmospheric scattering, aerosol degradation, visibility restoration.
{
Post-processing of the search results yielded 377 relevant publications for quantitative evaluation.}
{
\figurename~\ref{Fig.Number of paper}  shows that the annual number of remote sensing image dehazing publications has increased by more than 400\% since 2019, rising from fewer than 20 papers per year to over 100 by 2025, highlighting its emergence as a highly active research frontier in image restoration.
}
Before 2018, research activity remained limited, with fewer than 10 papers published annually. 
Starting in 2019 (13 papers), the number steadily increased to 24 by 2021.
As evidenced by \figurename~\ref{Fig.Number of paper}, from 2019 to the peak in 2024 (80 papers), the field experienced a growth of over 515\%, reflecting a significant surge in interest.
The counts in 2025 and 2026 demonstrate a sharp rise compared to pre-2019 levels.

\begin{figure}[t]
	\centering 
	\includegraphics[width=8cm]{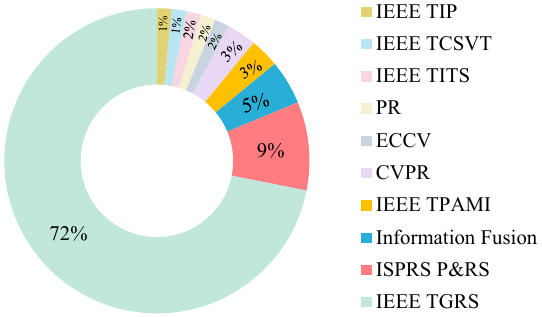}
	\caption{
    The distribution ratio of high-quality journals (impact factor greater than 7) and conference articles in the past three years.
    }
	\label{Fig.Publication}
\end{figure}

\tablename~\ref{tab:journal_if} and Figure~\ref{Fig.Publication} jointly present a comprehensive overview of the publication landscape in RSIs dehazing over the past three years, focusing on high-impact venues.
IEEE Transactions on Geoscience and Remote Sensing (TGRS) dominates both in quantity and influence, accounting for 72\% of all high-quality publications. 
This highlights its central role in disseminating methodological and applied advances in RSIs restoration. 
\textbf{The ISPRS Journal of Photogrammetry and Remote Sensing (ISPRS P\&RS) ranks second with a 9\% share, reflecting its increasing relevance, particularly at the intersection of photogrammetry, remote sensing, and computer vision.}
Other domain-specific journals, including Remote Sensing, IEEE Journal of Selected Topics in Applied Earth Observations and Remote Sensing (JSTARS), and IEEE Geoscience and Remote Sensing Letters (GRSL), also contribute significantly.

In contrast, high-impact venues such as Information Fusion and IEEE Transactions on Pattern Analysis and Machine Intelligence (TPAMI) publish relatively few RS dehazing papers, accounting for only 5\% of publications in Figure~\ref{Fig.Publication} and three papers, respectively. 
A similar trend is observed in major vision conferences such as CVPR and ECCV, which also feature relatively few RS dehazing studies, as shown in \tablename~\ref{tab:journal_if}, reflecting a preference for broader vision topics over domain-specific remote sensing applications.

\textbf{In summary, the field of RSIs dehazing is predominantly rooted in application-driven journals, with IEEE TGRS and ISPRS P\&RS serving as the primary publication platforms.} While interdisciplinary journals and vision conferences provide occasional contributions, their focus tends to favor methodological novelty over field-specific application. This distribution underscores the importance of domain expertise and specialized venues in advancing RSIs restoration research.

{
\subsection{Comparative Analysis of Existing Reviews}
In the early stage, Singh et al.~\cite{singh2018comprehensive} summarized the mathematical modeling and classification of traditional defogging methods. Thiruvikraman et al.~\cite{thiruvikraman2021survey} focused on the review of traditional image enhancement and filtering techniques. Sahu et al.~\cite{sahu2022trends} focused on traditional dehazed techniques from an image analytical perspective. Gao et al.~\cite{gao2009atmospheric}, facing the atmospheric correction problem in hyperspectral remote sensing, reviewed the physical correction methods based on the radiative transfer model and their improvement directions at the spectral level. 
Early work mainly relied on traditional pattern recognition methods based on handcrafted features, lacking the review of data-driven deep learning and generative reconstruction methods.
In recent years, Ayoub et al.~\cite{ayoub2025review} and Khan et al.~\cite{khan2022recent} conducted a review and experimental analysis of existing dehazing learning methods, such as physical models, CNNs, and GANs for natural images, but lacked the review of remote sensing images. 
Liu et al.~\cite{liu2021review} investigated the image enhancement, physical models, and data-driven methods in remote sensing image defogging, without covering the latest work such as Transformer and Diffusion model. 
Wang et al.~\cite{wang2024deep} and Ning et al.~\cite{ning2025cloud}, facing the optical remote sensing cloud detection, summarized the existing optical remote sensing cloud-removal methods from the perspectives of semantic segmentation and multi-source and multi-modal data, respectively. 
Sharma et al.~\cite{sharma2020review} also evaluated a variety of defogging methods and evaluation indicators based on a single real-image dataset for natural images. 
Rasti et al.~\cite{rasti2021image} reviewed the restoration methods for different types of noise and degradation in RSIs denoising, starting from the sensor type, while this paper further focuses on the specific imaging degradation problem of remote sensing dehazing.}

{Existing relevant reviews lack a unified evolutionary perspective, unable to connect handcrafted priors, deep models, and emerging large-model paradigms, not providing a standardized and repeatable large-scale benchmark, nor evaluating methods on synthetic and real-scene datasets. 
To present a clear and progressive view of the field, we organize remote sensing image defogging research into three successive paradigms, including physical-prior-driven enhancement, deep learning based restoration, and emerging physics and intelligence hybrid generation.
}

\section{Methodological Advances in Remote Sensing Dehazing}
\label{Sec.CurrentAdvances}
As illustrated in \figurename~\ref{Fig.Framwork}, this section provides a systematic exposition of current approaches from six methodological perspectives: 
image enhancement methods, physics model methods, deep convolutional methods, 
adversarial generation methods, vision transformer methods, and diffusion generation methods.
The \textbf{first stage} grounded restoration in physical priors, offering interpretability but limited adaptability to heterogeneous haze. 
The \textbf{second stage} shifted to deep and generative learning, replacing handcrafted assumptions with data-driven representations that improved realism yet sacrificed physical consistency and efficiency. 
The \textbf{third stage} now pursues hybrid physics–AI integration, embedding physical constraints into intelligent models to reconcile interpretability, generalization, and scalability. Collectively, these stages trace a coherent evolution from explicit physical modeling to adaptive learning and finally to unified physical–intelligent reasoning.

\subsection{Stage I: Handcrafted Priors Modeling}

\begin{figure}[t]
	\centering 
	\includegraphics[width=13.5cm]{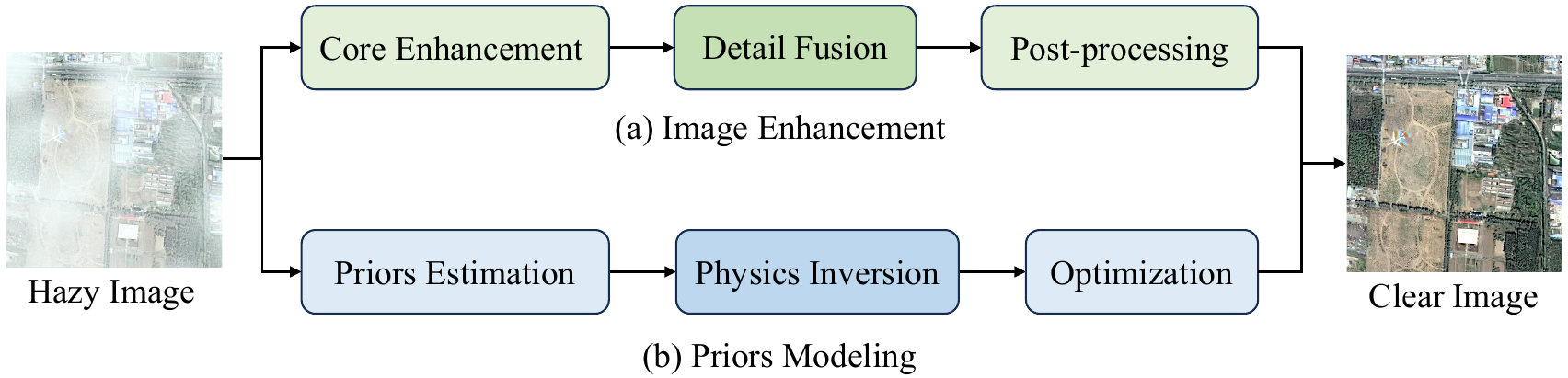}
	\caption{Diagram of Image Enhancement and Priors Modeling for RSIs Dehazing.}
	\label{Fig.Traditional methods}
\end{figure}

\subsubsection{Image Enhancement for Remote Sensing Image Dehazing}

As depicted in \figurename~\ref{Fig.Traditional methods} (a), image enhancement approaches operate without explicit modeling of the physical degradation processes, instead directly improving image visibility by manipulating pixel/frequency-domain distributions. 
HISTogram equalization improves visibility by expanding the dynamic range~\cite{5301485,thanh2019single}, while histogram matching aligns the intensity distribution of degraded images with that of haze-free references~\cite{helmer2005cloud}. 
These methods are computationally efficient and easy to implement, but rely on the assumption that cloudy and clear regions share similar spectral reflectance. 
When this condition is not met, performance degrades significantly, especially in heterogeneous scenes~\cite{shen2014effective}.
Moreover, their global nature often leads to radiometric distortion and detail loss.

Building on statistical methods, Retinex theory~\cite{land1971lightness} provides a more physically grounded approach by modeling the human visual system’s (HVS) ability to separate reflectance (object-intrinsic properties) from illumination (environmental lighting). 
Retinex theory assumes that human perception of brightness depends not only on absolute intensity but also on surrounding illumination. 
Accordingly, Retinex-based algorithms~\cite{xue2016video,gao2014enhancement,rong2014improved,fan2017improved} aim to remove ambient light effects and recover true surface reflectance.

Several Retinex variants have been developed by adjusting the scales of Gaussian filters. The Single-Scale Retinex (SSR)~\cite{liu2011image}, Multi-Scale Retinex (MSR)~\cite{zotin2018fast}, and MSR with Color Restoration (MSRCR)~\cite{parthasarathy2012automated} differ primarily in how they model illumination across spatial scales. 
MSR improves robustness under diverse illumination by aggregating responses across multiple spatial scales. The MSR output is given by:
\begin{equation}
	R_{\text{MSR}}(x) = \sum_{i=1}^{N} w_i \left[ \log \boldsymbol{I}_{hazy}(x) - \log \left( \boldsymbol{I}_{hazy}(x) \otimes G_{\sigma_i}(x) \right) \right],
	\label{eq:retinex_msr}
\end{equation}
where $G_{\sigma_i}(x)$ is a Gaussian kernel with a standard deviation $\sigma_i$, $\otimes$ denotes convolution, and $w_i$ represents normalized weights summing to 1.
Typically, three scales (\textit{e.g.}, $\sigma = 15, 80, 250$) in Eq.\eqref{eq:retinex_msr} are used to capture illumination variation at different spatial resolutions.

{By fusing fine- and coarse-scale illumination cues, MSR improves local contrast and structural consistency, making it well suited for complex remote sensing scenes affected by haze and shadows~\cite{xue2016video}. 
However, early MSR-based methods commonly exhibited halo artifacts and detail blurring, largely attributed to the inherent Gaussian smoothing in the Retinex formulation~\cite{choudhury2009perceptually}. Motivated by these limitations, later works introduced a series of refinements to better balance enhancement strength and detail preservation, including adaptive-scale MSRCR to alleviate over-smoothing~\cite{gao2014enhancement}, wavelet-domain SSR post-processing to selectively recover high-frequency structures~\cite{rong2014improved}, and hybrid HSV-RGB enhancement strategies to improve color fidelity while maintaining contrast~\cite{fan2017improved}. 
Along a complementary line, Shen et al.~\cite{shen2014exposure} proposed an exposure fusion framework with Laplacian pyramid boosting and exposure-guided weighting, explicitly targeting the texture loss issue of conventional MSR and producing more detail-preserving enhancement results.
}

\subsubsection{Priors Modeling for Remote Sensing Image Dehazing}
As shown in \figurename~\ref{Fig.Traditional methods} (b), 
The prior modeling method is based on atmospheric scattering models and radiative transfer theory, and aims to reconstruct the surface reflectance by inverting the degradation process introduced during image acquisition.
Early works focused on the atmospheric scattering model, which describes how light interacts with atmospheric particles during transmission. 
Dark Channel Prior (DCP)~\cite{he2010single} is a seminal method, built on the observation that at least one color channel often contains pixels with low intensity in haze-free outdoor scenes.
More recent advances have drawn inspiration from neural radiance fields (NeRF)~\cite{mildenhall2021nerf}, reinterpreting atmospheric scattering from a volumetric rendering perspective to enable continuous-depth haze modeling:
\begin{align}
	\boldsymbol{I}_{hazy} 
    &= \int_{0}^{D} T(s) \cdot \sigma(s) \cdot \boldsymbol{I}_{clear}(s) \, ds + (1 - T(D)) \cdot \boldsymbol{A} \\
	T(s) 
    &= \exp\left(-\int_{0}^{s} \sigma(u) \, du\right)
	\label{eq:nerf_style_model}
\end{align}
where $\boldsymbol{I}_{hazy}$ is the observed radiance, $\boldsymbol{I}_{clear}(s)$ is the clear radiance at depth $s$, $\sigma(s)$ is the haze extinction coefficient, $T(s)$ is the accumulated transmittance at depth $s$, $\boldsymbol{A}$ is the global atmospheric light, and $D$ is the farthest integration depth.
Eq.~\eqref{eq:nerf_style_model} provides a physically grounded and depth-continuous interpretation of haze formation, bridging traditional atmospheric models with differentiable rendering paradigms used in 3D reconstruction and neural field learning.

To address the failures of DCP in sky regions and overexposed white surfaces,
{Subsequent studies progressively introduced more robust priors and region-adaptive refinements. 
Zhu et al.~\cite{zhu2015fast} proposed the Color Attenuation Prior (CAP), leveraging brightness–saturation statistics to improve transmission estimation in cases where the dark-channel assumption breaks down. 
Motivated by DCP’s sensitivity to sky–ground ambiguity, Yuan et al.~\cite{yuan2017effective} further incorporated a Gaussian Mixture Model (GMM) to explicitly segment sky and non-sky regions, thereby enhancing DCP’s adaptability across heterogeneous scenes.
Along a complementary direction, Long et al.~\cite{long2013single} introduced Gaussian filtering to refine the transmission map, aiming to suppress artifacts and alleviate color distortion caused by noisy or over-sharp transmission estimates. 
Building upon these region-aware improvements, Liang et al.~\cite{liang2024remote} combined heterogeneous priors with superpixel segmentation to better capture structural boundaries and improve dehazing performance in remote sensing imagery. 
Similarly, to specifically address DCP’s degradation on deserts and snowfields, Li et al.~\cite{li2018haze} developed a two-stage framework that integrates homomorphic filtering with an improved DCP equipped with a sphere model.
% , explicitly targeting high-reflectance surfaces where DCP tends to underestimate transmission. 
Finally, to handle the scale variance of haze effects across large scenes, Shi et al.~\cite{shi2021novel} proposed a multiscale DCP formulation that couples multiscale transform-domain representations with optimized Laplacian sharpening, enabling more consistent restoration across spatial scales.
}

\begin{figure}[t]
	\centering 
	\includegraphics[width=13cm]{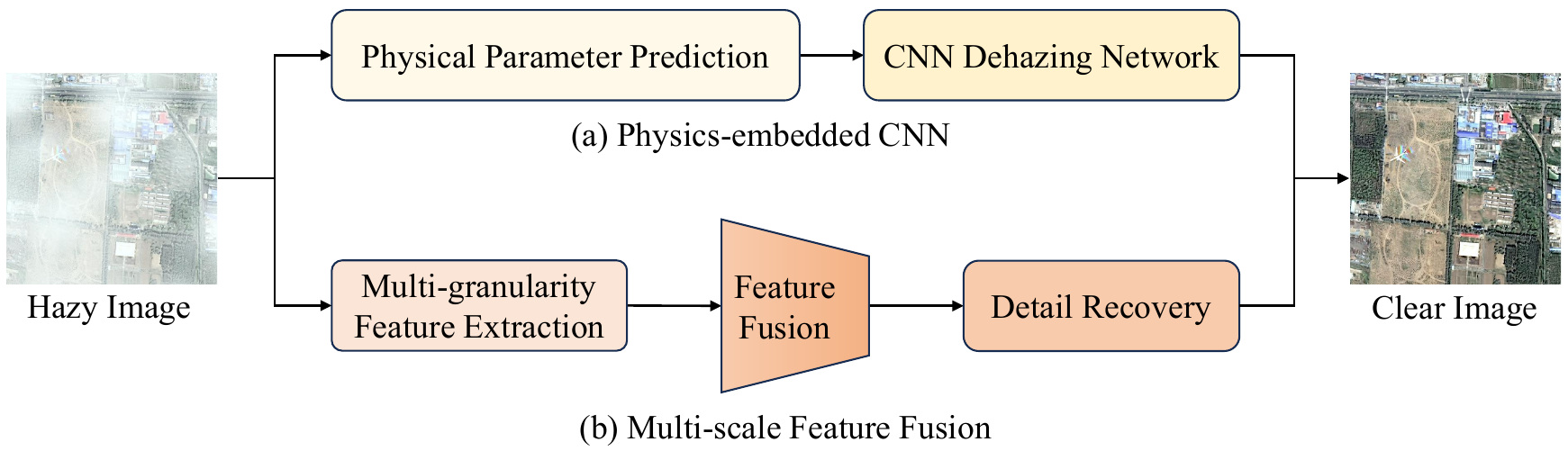}
	\caption{
    Diagram of Deep Convolution Methods for RSIs Dehazing.}
	\label{Fig.CNN}
\end{figure}

% ---- CNN -----------
\subsection{Stage II: Data-driven Deep Restoration}

\subsubsection{Deep Convolution for Remote Sensing Image Dehazing}
While traditional physics models laid a solid foundation for dehazing, they often rely on handcrafted priors and heuristic assumptions. With the rise of deep learning, data-driven CNN~\cite{lecun2015deep} have emerged as a powerful alternative. 
Deep convolution dehazing methods can be categorized into three classes based on technical principles: physics-embedded methods and multi-scale feature fusion methods.

As shown in \figurename~\ref{Fig.CNN} (a), physics-embedded frameworks represent a hybrid paradigm that integrates traditional physical degradation equations with deep learning~\cite{lihe2025ada4dir}, aiming to alleviate the limited interpretability and physical inconsistency of purely data-driven CNN models. 
{As an early representative, AOD-Net~\cite{li2017aod} embedded the atmospheric scattering model into the network architecture to jointly estimate transmittance and atmospheric light, reducing the error accumulation of decoupled parameter prediction. 
Building upon this parameter-embedded design, Zi et al.~\cite{zi2021thin} further integrated U-Net with the imaging physics model to estimate band-wise cloud thickness coefficients in multi-spectral imagery and generate cloud thickness maps. 
Motivated by the remaining gap between physical constraints and feature learning, Lihe et al.~\cite{lihe2024phdnet} combined the atmospheric scattering model with residual learning, improving both interpretability and performance by enforcing physically consistent restoration. 
More recently, to better exploit physical priors in deep feature space, Sun et al.~\cite{sun2025bidirectional} proposed a differential-expert guided bidirectional modulation module that injects atmospheric-model knowledge via physical inversion experts. By explicitly modeling the complementarity of CNN--Transformer features with differential experts, the module captures latent haze-related physical characteristics and achieves more effective bidirectional alignment.
}

As illustrated in \figurename~\ref{Fig.CNN} (b), recent advances in RSIs dehazing have emphasized three structured stages: multi-granularity feature extraction, feature fusion, and detail recovery, moving beyond simple end-to-end mappings towards more refined frameworks.
{In the feature extraction stage, early CNN-based DehazeNet~\cite{cai2016dehazenet} method employed multi-layer convolutions to regress transmission maps, reducing the heavy dependence on hand-crafted local priors (DCP). Nevertheless, these early designs often showed limited adaptability to complex scenes and multi-/hyper-spectral inputs. 
Motivated by this limitation, Ren et al.~\cite{ren2016single} introduced a coarse-to-fine architecture to progressively refine transmission estimation, better balancing global structures and local details. 
Building on the need to further exploit cross-band dependencies, Guo et al.~\cite{guo2020rsdehazenet} enhanced feature extraction with channel refinement blocks and attention mechanisms.
In parallel, feature fusion strategies in FCTF-Net~\cite{li2020coarse} were developed to better integrate multi-scale and multi-source information. 
For the final detail recovery stage, later works~\cite{dong2024icl} were driven by the observation that residual haze and subtle structures are easily smoothed out.
To explicitly strengthen fine-detail restoration, 
Yu et al.~\cite{yu2022cloud} incorporated cloud-distortion-aware representation modules into a multiscale grid architecture to handle complex cloud distortion patterns, while Dong et al.~\cite{dong2024end} introduced multi-scale and stepped-attention detail enhancement units to reinforce delicate textures. 
}

% --------

\begin{figure}[t]
	\centering 
	\includegraphics[width=13cm]{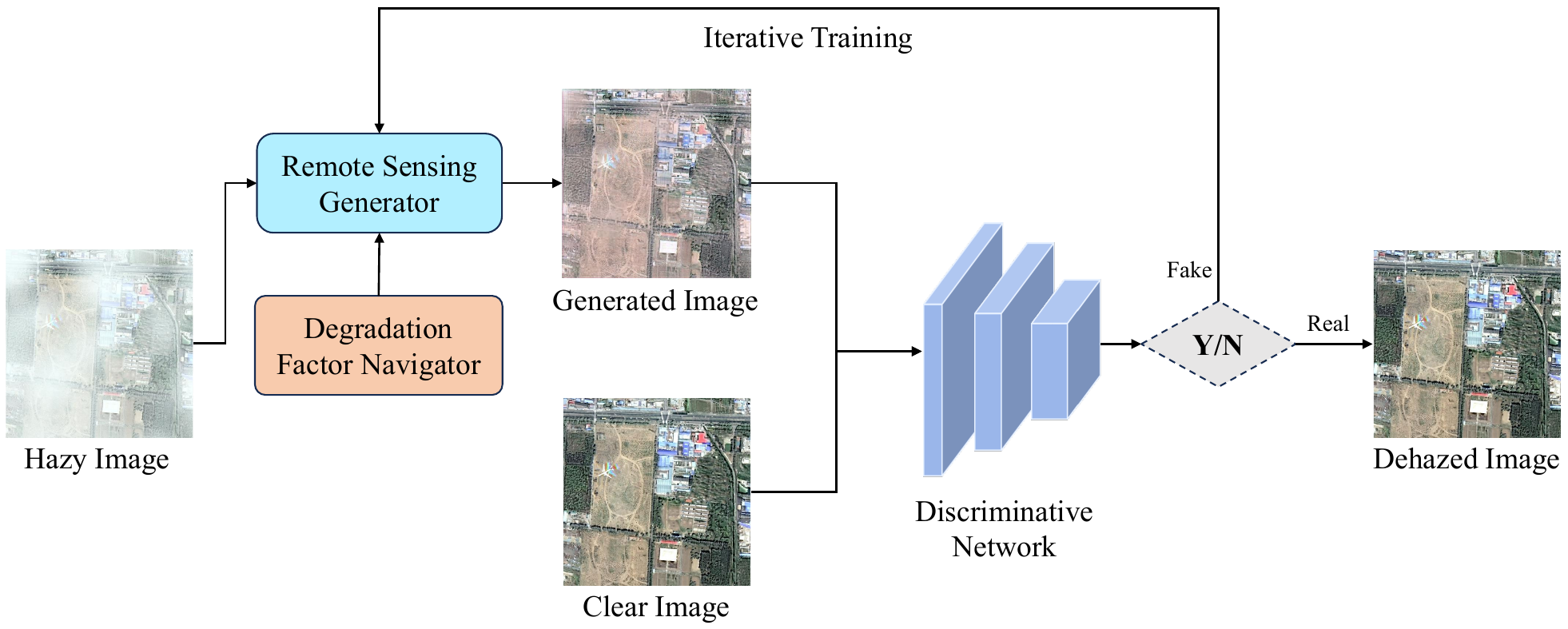}
	\caption{Diagram of Adversarial Generation Methods for RSIs Dehazing.}
	\label{Fig.GAN}
\end{figure}

\subsubsection{Adversarial Generation for Remote Sensing Image Dehazing}
Adversarial generation methods learn data distributions through an adversarial training mechanism, with its core lies in establishing a dynamic game process between the generator and discriminator, aiming to achieve efficient fitting of complex data distributions~\cite{goodfellow2020generative}.
As shown in \figurename~\ref{Fig.GAN}, adversarial learning has been widely explored for end-to-end dehazing. 
An adversarial generation objective for RSIs dehazing is formulated as:
\begin{equation}
	\min_{G} \max_{D} \mathcal{L}(G, D) = \mathbb{E}_{\boldsymbol{I}_{clear}}[\log D(\boldsymbol{I}_{clear})] + \mathbb{E}_{\boldsymbol{I}_{hazy}}[\log(1 - D(G(\boldsymbol{I}_{hazy})))]
	\label{eq:gan_objective}
\end{equation}
where $G$ denotes the generator that predicts the dehazed image $\boldsymbol{I}_{dehazed} = G(\boldsymbol{I}_{hazy})$ from an input hazy image, and $D$ is the discriminator trained to distinguish between real clear images and generated results.

Cloud-GAN~\cite{singh2018cloud} pioneered GAN-based cloud removal by using a generator to translate cloudy images into cloud-free ones, while a discriminator distinguishes real from restored images. 
However, early GAN formulations often lacked robustness and physical interpretability, inheriting the limitations of traditional hand-crafted features and low-rank assumptions. 
Motivated by this issue, Li et al.~\cite{li2020thin} proposed a semi-supervised thin cloud removal method, CR-GAN-PM, which couples GAN learning with a physical model of cloud distortion, jointly modeling bidirectional transmittance, reflectance, and absorption to explicitly capture cloud–multispectral interactions. 
To further reduce restoration ambiguity caused by mixed cloud and background regions, Zheng et al.~\cite{zheng2020single} developed a two-stage pipeline that first performs cloud segmentation with U-Net and then applies GAN-based restoration. 
Along another direction, Huang et al.~\cite{huang2020single} introduced a conditional GAN that fuses SAR priors to strengthen dehazing under heavy cloud occlusion, exploiting cross-modal complementarity. 
Extending this idea toward fully unsupervised multimodal translation, Wang et al.~\cite{wang2025mt_gan} proposed an end-to-end SAR-to-optical framework with multi-scale feature fusion, a despeckling module, and cycle-consistency constraints to generate cloud-free optical images without paired supervision. 
Similarly, Mehta et al.~\cite{mehta2021domain} presented Sky-GAN for aerial image dehazing by incorporating HSI guidance and multi-cue color inputs to improve generalization across domains.
In parallel, Darbaghshahi et al.~\cite{darbaghshahi2021cloud} designed a two-stage GAN framework where SAR-to-optical translation is first performed, and the translated results are then combined with optical data for cloud removal, further enhancing robustness in challenging cloudy scenarios.

Some researchers have integrated attention mechanisms to enhance model performance.  
Pan~\cite{pan2020cloud} introduced the spatial attention mechanism into the RSIs cloud removal task, dynamically identifying cloud regions through the local-to-global spatial attentive module and optimizing feature restoration by incorporating attention-guided residual blocks.
{To address the challenge of difficulty in large amounts of paired real-world data,}
Dehaze-AGGAN~\cite{zheng2022dehaze} employs an enhanced attention-guided GAN, combined with cycle consistency loss and total variation loss, to achieve unsupervised training.

\subsubsection{Vision Transformer for Remote Sensing Image Dehazing}
The vision Transformer architectures~\cite{vaswani2017attention,dong2024dehazedct,xu2025tam} represent a paradigm shift in deep learning by introducing self-attention mechanisms that directly model long-range dependencies within data. This framework was rapidly extended to computer vision tasks, most notably with the Vision Transformer (ViT)~\cite{dosovitskiy2020image,yuan2024lhnetv2}, which treats images as sequences of patches, thus enabling holistic scene understanding across large spatial extents.

\begin{figure}[t]
	\centering 
	\includegraphics[width=13.5cm]{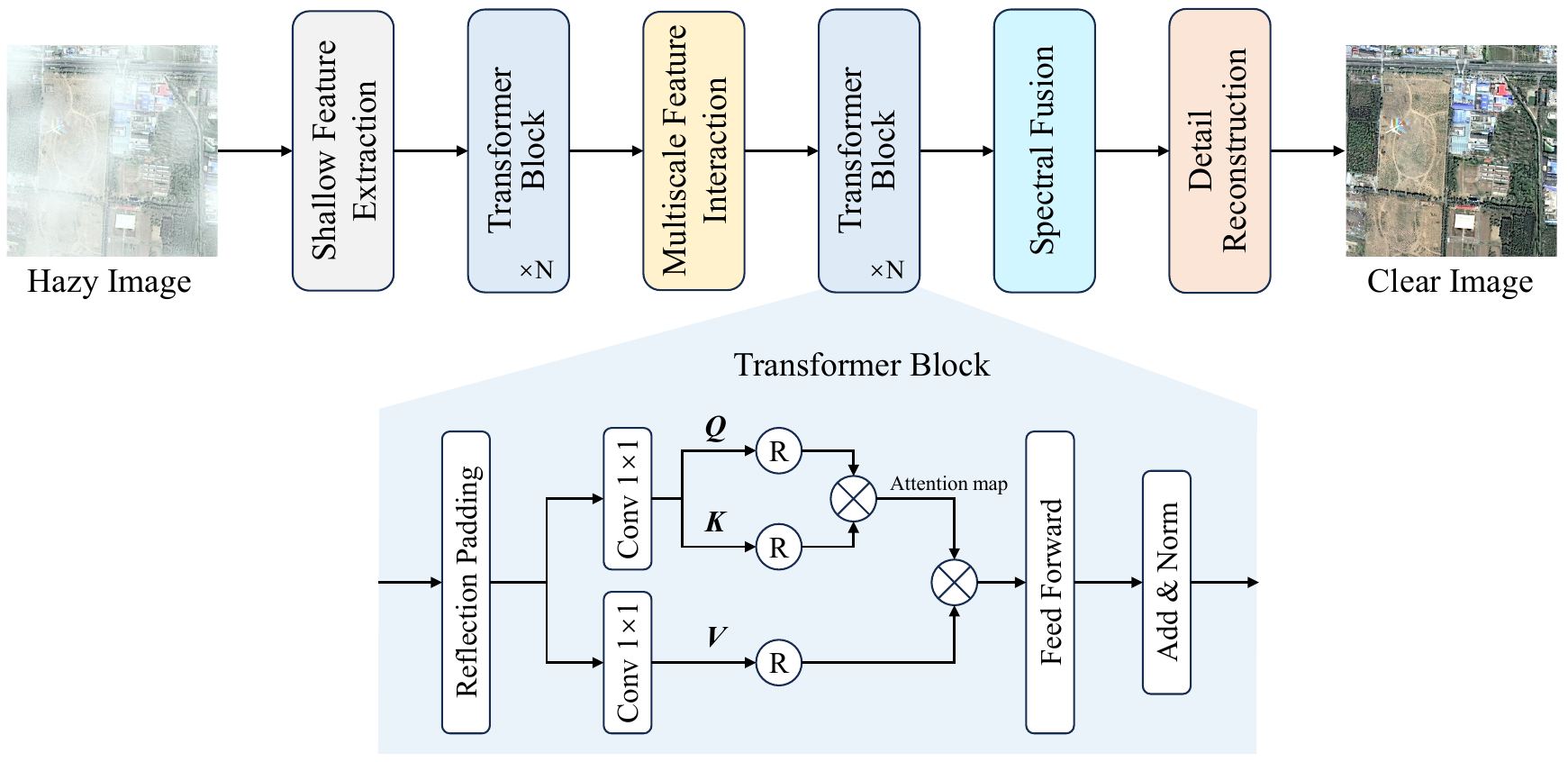}
	\caption{Diagram of Vision Transformers Methods for RSIs Dehazing.}
	\label{Fig.Transformer}
\end{figure}

Recent advances have leveraged the remarkable capacity of the Transformer to aggregate contextual information, resulting in significant progress for RSIs restoration~\cite{han2023former,song2023learning,chen2025tokenize}. 
In the context of dehazing, a hazy image $\boldsymbol{I}_{hazy}$ is first divided into $N$ non-overlapping patches, each embedded as a feature vector. The self-attention mechanism then constructs query, key, and value matrices:
\begin{equation}
    \mathbf{Q} = [\mathbf{x}_1, \ldots, \mathbf{x}_N]\mathbf{W}^Q, \quad
    \mathbf{K} = [\mathbf{x}_1, \ldots, \mathbf{x}_N]\mathbf{W}^K, \quad
    \mathbf{V} = [\mathbf{x}_1, \ldots, \mathbf{x}_N]\mathbf{W}^V,
\end{equation}
where $\mathbf{W}^Q, \mathbf{W}^K, \mathbf{W}^V$ are learnable projections and $[\mathbf{x}_1, \ldots, \mathbf{x}_N]$ are the embedded patch features. The self-attention operation is then defined as:
\begin{equation}
    \text{Attention}(\mathbf{Q}, \mathbf{K}, \mathbf{V}) = \text{softmax}\left(\frac{\mathbf{Q}\mathbf{K}^\top}{\sqrt{d}}\right)\mathbf{V},
    \label{eq:self_attention_rsdhaze}
\end{equation}
where $d$ denotes the feature dimension.
This self-attention mechanism allows the model to selectively aggregate information from spatially distant yet semantically correlated regions, significantly enhancing its capacity to handle the complex and diverse haze distributions and land-cover variations commonly encountered in remote sensing scenarios.
By leveraging local texture and global scene structure, Transformer-based methods~\cite{dong2022transra,song2023vision} demonstrate distinct advantages over CNN approaches, facilitating the restoration of fine-grained surface details while preserving scene-scale consistency.

{
Recent transformer-based methods~\cite{song2023learning,li2025decloudformer} have been motivated by the limitations of earlier CNN and GAN models in handling the highly irregular shapes and non-uniform distributions of haze in RSIs, and thus increasingly emphasize long-range dependency modeling and adaptive feature interaction.
Building upon the need to better preserve fine textures and sharp boundaries that are often over-smoothed by global attention, Kulkarni et al.~\cite{kulkarni2023aerial} proposed deformable multi-head attention with a spatially attentive offset extractor and edge-boosting skip connections to retain texture and edge details. 
To improve physical plausibility beyond purely data-driven attention, Trinity-Net~\cite{chi2023trinity} injects DCP-derived priors into a Swin Transformer backbone, leveraging global modeling to achieve more reliable estimation of haze-related parameters. 
Motivated by the strong data dependency of existing restoration models, Quan et al.~\cite{quan2024density} further proposed DCR-GLFT, introducing cloud density classification labels as explicit guidance to control the restoration process under different coverage levels. 
Moreover, to address the practical constraint of unpaired training data, Zheng et al.~\cite{zheng2024dehaze} developed Dehaze-TGGAN based on Cycle-GAN~\cite{zhu2017unpaired}, integrating spatial-spectrum attention and semi-transparent mask pre-training, and adopting total variation loss to enhance generation quality for unsupervised RSIs dehazing. 
Extending multimodal restoration toward high-dimensional inputs, Du et al.~\cite{du2024ssgt} proposed SSGT with a spatiospectral-guided transformer to jointly tackle hyperspectral–multispectral fusion and cloud removal.}
To further address the memory limitation issue in processing large-scale RSIs, 
Chen et al.~\cite{chen2025tokenize} employed patch-wise encoding and global attention fusion mechanisms to effectively integrate long-range contextual information while preserving local details and supporting end-to-end processing of $ 8192\times8192 $ images.

% ----------- Diffusion ---------

\begin{figure}[t]
	\centering 
	\includegraphics[width=13cm]{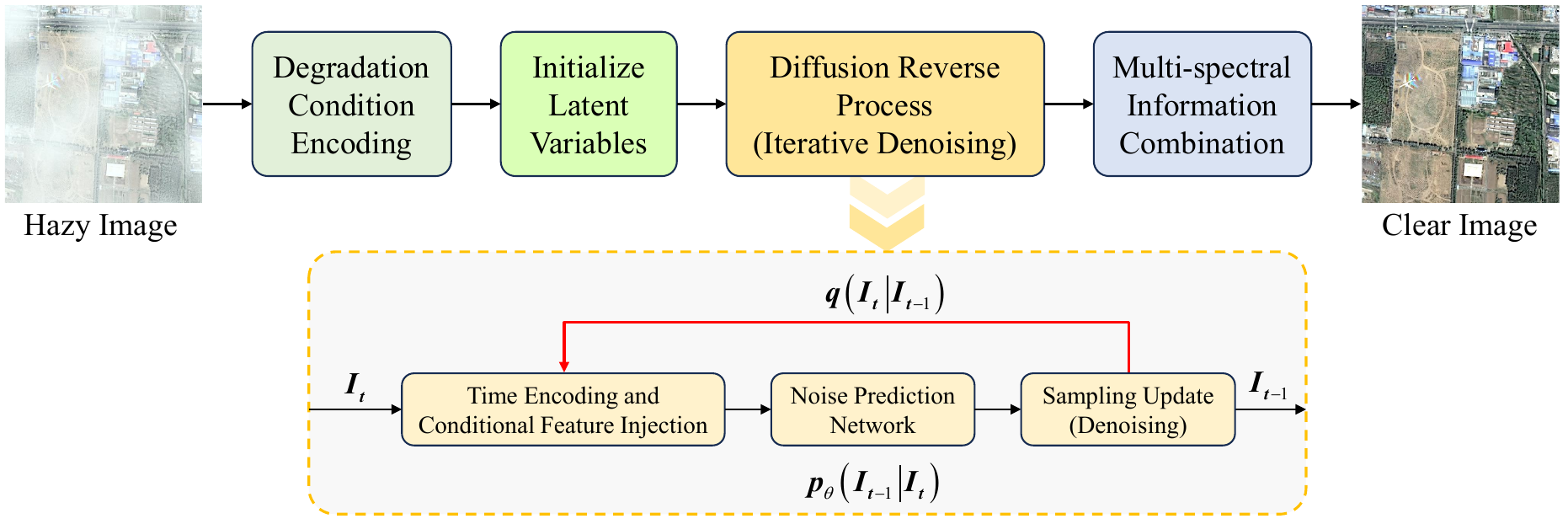}
	\caption{Diagram of Diffusion Generation Methods for RSIs Dehazing.}
	\label{Fig.Diffusion}
\end{figure}

\subsubsection{Diffusion Generation for Remote Sensing Image Dehazing}
Recently, diffusion models have achieved remarkable progress in content generation quality and complex scene handling~\cite{xing2024survey,chen2025comprehensive}.
Diffusion generative frameworks operate through Markov chain processes involving progressive forward noising and reverse denoising operations~\cite{jiangzhou2024dgrm,liu2024diffusion,sui2024diffusion}.
As illustrated in \figurename~\ref{Fig.Diffusion}, denoising diffusion probabilistic models (DDPMs)~\cite{ho2020denoising,ozdenizci2023restoring,yin2025aware} have recently emerged as powerful generative models for image dehazing tasks. The iterative denoising process for dehazing can be described as:
\begin{equation}
	\boldsymbol{I}_{t-1} = \frac{1}{\sqrt{\alpha_t}}\left(\boldsymbol{I}_t - \frac{1-\alpha_t}{\sqrt{1-\bar{\alpha}_t}} \boldsymbol{\epsilon}_\theta(\boldsymbol{I}_t, t) \right) + \sigma_t \mathbf{z},
	\label{eq:ddpm_restore}
\end{equation}
where $\boldsymbol{I}_t$ represents the intermediate noisy prediction at timestep $t$, $\boldsymbol{\epsilon}_\theta$ is the learned noise estimator, $\alpha_t$ and $\bar{\alpha}_t$ are the noise schedule and its cumulative product, $\sigma_t$ is the standard deviation of the noise, and $\mathbf{z}$ is standard Gaussian noise.

As an early exploration, ARDD-Net~\cite{huang2023remote} {introduced an adaptive region-based diffusion framework to enable arbitrary-sized inputs through region segmentation.
Motivated by the need to further exploit complementary cues beyond single-modality restoration, Zhao et al.~\cite{zhao2023cloud} extended diffusion dehazing to a multi-modal setting, integrating auxiliary modal information to strengthen optical distribution learning.
Nevertheless, these early diffusion-based attempts still suffer from two fundamental limitations~\cite{guo2022alleviating,tang2023emergent}: 
local distortion due to insufficient high-frequency detail representation and difficulty in maintaining global semantic consistency across reconstructed scenes. 
Building upon this foundation, Huang et al.~\cite{huang2024diffusion} proposed ADND-Net, which formulates atmospheric transmission as a range–null-space decomposition and explicitly enforces null-space consistency during reverse diffusion.
Along a complementary direction, RSHazeDiff~\cite{xiong2024rshazediff} enhanced diffusion sampling by acquiring texture structures and color priors in early steps via a Fourier-aware iterative refinement module, further alleviating both local distortion and global inconsistency.
}

\begin{table}[t]
	\caption{Selected publicly available codes for RSIs dehazing methods.}
	\label{tab:Code}
	\centering
	\resizebox{\textwidth}{!}{%
	\begin{tabular}{lll}
		\toprule
		\textbf{Method} & \textbf{Publication} & \textbf{Code Link} \\
		\midrule
		CR-GAN-PM~\cite{li2020thin} & ISPRS P\&RS 2020 & \url{https://github.com/Neooolee/CR-GAN-PM} \\
		% FCTF-Net~\cite{li2020coarse} & GRSL 2020 & \url{https://github.com/Neooolee/FCTF-Net} \\
		RSDehazeNet~\cite{guo2020rsdehazenet} & TGRS 2020 & \url{https://github.com/tianqiong123/RSDehazeNet} \\
		DCIL~\cite{zhang2022dense} & TGRS 2022 & \url{https://github.com/Shan-rs/DCI-Net} \\
		GLF-CR~\cite{xu2022glf} & ISPRS P\&RS 2022 & \url{https://github.com/xufangchn/GLF-CR} \\
		DehazeFormer~\cite{song2023vision} & TIP 2023 & \url{https://github.com/IDKiro/DehazeFormer} \\
		RSDformer~\cite{song2023learning} & GRSL 2023 & \url{https://github.com/MingTian99/RSDformer} \\
		PSMB-Net~\cite{sun2023partial} & TGRS 2023 & \url{https://github.com/thislzm/PSMB-Net} \\
		Trinity-Net~\cite{chi2023trinity} & TGRS 2023 & \url{https://github.com/chi-kaichen/Trinity-Net} \\
		LFD-Net~\cite{jin2023lfd} & JSTARS 2023 & \url{https://github.com/RacerK/LFD-Net} \\
		EMPF-Net~\cite{wen2023encoder} & TGRS 2023 & \url{https://github.com/chdwyb/EMPF-Net} \\
        HS2P~\cite{li2023hs2p} & Information Fusion 2023 & \url{https://github.com/weifanyi515/HS2P} \\
        AIDTransformer~\cite{kulkarni2023aerial} & WACV 2023 & \url{https://github.com/AshutoshKulkarni4998/AIDTransformer} \\
		SFAN~\cite{shen2024spatial} & TGRS 2024 & \url{https://github.com/it-hao/SFAN} \\
		IDF-CR~\cite{wang2024idf} & TGRS 2024 & \url{https://github.com/SongYxing/IDF-CR} \\
		DiffCR~\cite{zou2024diffcr} & TGRS 2024 & \url{https://github.com/XavierJiezou/DiffCR} \\
		ASTA~\cite{cai2024additional} & GRSL 2024 & \url{https://github.com/Eric3200C/ASTA} \\
		TCBC~\cite{xu2024thin} & ISPRS P\&RS 2024 & \url{https://github.com/Liying-Xu/TCBC} \\
		RSHazeDiff~\cite{xiong2024rshazediff} & TITS 2024 & \url{https://github.com/jm-xiong/RSHazeDiff} \\
        PhDnet~\cite{lihe2024phdnet} & Information Fusion 2024 & \url{https://github.com/colacomo/PhDnet} \\
        C2AIR~\cite{kulkarni2024c2air} & WACV 2024 & \url{https://github.com/AshutoshKulkarni4998/C2AIR} \\
		HPN-CR~\cite{gu2025hpn} & TGRS 2025 & \url{https://github.com/G-pz/HPN-CR} \\
		MT\_GAN~\cite{wang2025mt_gan} & ISPRS P\&RS 2025 & \url{https://github.com/NUAA-RS/MT_GAN} \\
        DecloudFormer~\cite{li2025decloudformer} & PR 2025 & \url{https://github.com/N1rv4n4/DecloudFormer} \\
		DehazeXL~\cite{chen2025tokenize} & CVPR 2025 & \url{https://github.com/CastleChen339/DehazeXL} \\
		EMRDM~\cite{liu2025effective} & CVPR 2025 & \url{https://github.com/Ly403/EMRDM} \\
        % CLIP-HNet~\cite{wang2025clip} & ACM MM & \url{https://github.com/CastleChen339/DehazeXL} \\
		\bottomrule
	\end{tabular}
}
\end{table}

To address the inherent blocking artifacts and detail blurring in diffusion models, Wang et al.~\cite{wang2024idf} proposed a two-stage optimization strategy: initially performing coarse cloud removal through integrating the Swin Transformer and cloudy attention components, followed by detail refinement via a latent space iterative noise diffusion network.
Liu et al.~\cite{liu2025effective} introduced an improved cloud removal network EMRDM based on mean-reverting diffusion models, which significantly enhances cloud removal performance in complex scenarios by reconstructing the mathematical representation and framework design of the diffusion process.

\subsection{Stage III: Hybrid Physical–Intelligent Generation}
The third stage reflects the convergence of physics-based reasoning and intelligent learning within unified frameworks. 
Physically constrained learning, multimodal data fusion, and novel network architectures aim to combine the interpretability of physical modeling with the adaptability of deep generation. 
By embedding physical priors into neural representations and integrating multisource information, this stage seeks to balance scientific fidelity, semantic generalization, and computational scalability.

Large language models (LLMs) and vision–language models increasingly enable dehazing systems to move beyond purely pixel-driven learning by providing text prompts for atmospheric reasoning, cross-modal guidance, semantic priors, and adaptive defogging instructions. 
{
Motivated by the limited supervision and weak detail recovery of conventional self-supervised pipelines, Wang et al.~\cite{wang2025clip} proposed a cross-modal guided self-supervised dehazing network that leverages CLIP~\cite{radford2021learning} to assess haze concentration, thereby enhancing fine-detail restoration. 
Building on the need for more reliable region localization and stronger generalization to real scenes, Wang et al.~\cite{wang2025hazeclip} further introduced a CLIP language-guided adaptive model that accurately identifies haze regions and injects human-like prior knowledge, effectively alleviating domain shift in real-world haze images. 
Beyond language-guided priors, Zhou et al.~\cite{zhou2025scalevim} addressed the persistent coupling between haze degradation and intrinsic scene structures by proposing a multi-scale physical-decoupling strategy.
More recently, motivated by the spatiotemporal inconsistency of real-world video defogging, Ren et al.~\cite{ren2025triplane} exploited CLIP to obtain a richer high-level understanding of fog effects, and further extending language-guided dehazing from single images to dynamic scenarios.
}

Multimodal fusion and state-space models (SSMs)~\cite{zhang2025dehazemamba,kong2025efficient} have recently provided new directions for RSIs dehazing.
{Early multimodal attempts, Meraner et al.~\cite{meraner2020cloud}, improved dehazing under cloud cover by jointly designing a residual network with SAR–optical fusion and a cloud-adaptive loss.
It demonstrates that SAR cues can compensate for the severe attenuation of optical observations. 
Building on the need for more effective global–local interaction beyond conventional convolutions, Han et al.~\cite{han2023former} introduced the Lewin-Transform block 
to better extract and fuse complementary information from SAR and optical images. 
More recently, motivated by the high computational cost of attention and the difficulty of capturing spatially varying haze over wide areas, Zhao et al.~\cite{zhao2025dehazemamba} proposed DehazeMamba, a SAR–optical fusion framework that leverages an adaptive SSM and progressive haze-decoupling fusion to enhance large-scale RSIs dehazing. 
In parallel, to improve efficiency while retaining long-range modeling capacity, Zhou et al.~\cite{zhou2024rsdehamba} developed RSDehamba, a lightweight Vision Mamba-based architecture that embeds an SSM into a U-Net backbone to adaptively aggregate spatially varying haze features. 
Along the same line of addressing non-uniform haze without heavy attention modules, Sui et al.~\cite{sui2025u} further presented a Mamba-based dehazing network tailored for remote sensing imagery, strengthening robustness to heterogeneous haze distributions.
}

For researchers engaged in long-term RSIs dehazing studies, the sharing of open-source code significantly lowers the barrier to algorithm reproduction and technological iteration. 
\tablename~\ref{tab:Code} systematically catalogs representative open-source projects in this domain on GitHub, covering classic methodologies based on CNN, GAN, Transformer, and diffusion models. 
Given the predominant use of Python in existing open-source repositories, programming language categorization is not presented separately in this survey.

\subsection{Evolutionary Coupling and Trajectory}
The evolution of remote sensing dehazing reveals a continuous coupling between physical modeling, representation learning, and intelligent integration rather than a sequence of isolated paradigms. 
Physics-based methods provided analytical priors describing radiative transfer and scattering, which later served as structural or loss constraints within deep architectures. 
Data-driven models, in turn, extended these physical formulations by learning nonlinear mappings that captured spatial heterogeneity, temporal variability, and multi-sensor discrepancies often neglected in traditional models. 
The emerging hybrid frameworks further close this loop by embedding physical consistency into generative networks and leveraging multimodal information from optical, thermal, and radar sources to achieve more robust and physically credible dehazing. 
This trajectory demonstrates a clear methodological maturation: from purely radiometric inversion toward physics-aware deep reasoning, gradually transforming remote sensing dehazing into a unified framework capable of balancing interpretability, generalization, and operational scalability.

\begin{table}[t]
	\caption{
    Summary of representative remote-sensing haze datasets, including data scale, real/synthetic type, sensor source, spectral counts, resolution, and applications.
    }
	\label{tab:Datasets}
	\centering
	\resizebox{\textwidth}{!}{%
    \begin{tabular}{llllcccc}
		\toprule
		\textbf{Datasets} & \textbf{Number} & \textbf{Real/Syn.} & \textbf{Sensor Type} & \textbf{$\boldsymbol{N}_{spectrum}$} & \textbf{Image Size} & \textbf{Application} &  \textbf{Link} \\
		\midrule
        RICE1$_{19}$~\cite{lin2019remote} & 500 & Real & Google Earth & 3 & $512\times512$ & Hazy, Thin cloud & \href{https://github.com/BUPTLdy/RICE_DATASET}{GitHub} \\
        RICE2$_{19}$~\cite{lin2019remote} & 736 & Real & Landsat 8 OLI/TIRS & 3 & $512\times512$ & Hazy, Thin cloud & \href{https://github.com/BUPTLdy/RICE_DATASET}{GitHub} \\
        SateHaze1k$_{20}$~\cite{huang2020single} & 1200 & Syn. & GF-2 & 3 & $512\times512$ & Hazy & \href{https://www.dropbox.com/s/k2i3p7puuwl2g59/Haze1k.zip}{GitHub} \\
        LHID$_{22}$~\cite{zhang2022dense} & 31017 & Syn. & Google Earth & 3 & $512\times512$ & Hazy & \href{https://github.com/Shan-rs/DCI-Net?tab=readme-ov-file}{GitHub} \\
        DHID$_{22}$~\cite{zhang2022dense} & 14990 & Syn. & Aerial Camera & 3 & $512\times512$ & Hazy & \href{https://github.com/Shan-rs/DCI-Net?tab=readme-ov-file}{GitHub} \\
        RS-HAZE$_{23}$~\cite{song2023vision} & 51300 & Syn. & Landsat 8 OLI/TIRS & 11 & $512\times512$ & Hazy & \href{https://github.com/IDKiro/DehazeFormer}{GitHub} \\
        RSID$_{23}$~\cite{chi2023trinity} & 1000 & Syn. & Google Earth & 3 & $256\times256$ & Hazy & \href{https://github.com/chi-kaichen/Trinity-Net}{GitHub} \\
        HN-Snowy$_{23}$~\cite{guo2023blind} & 1237 & Real & Landsat 8 OLI/TIRS & 11 & $256\times256$ & Hazy & \href{https://github.com/Merryguoguo/CP-FFCN}{GitHub} \\
        CUHK-CR$_{24}$~\cite{sui2024diffusion} & 1227 & Real & Jilin-1 & 4 & $512\times512$ & Thin cloud & \href{https://github.com/littlebeen/DDPM-Enhancement-for-Cloud-Removal?tab=readme-ov-file}{GitHub} \\
        {HyperDehazing$_{24}$~\cite{fu2024hyperdehazing}} 
        & 2140 & Real, Syn. & GF-5 AHSI & 305 & $512\times512$ & Hazy & \href{https://github.com/RsAI-lab/HyperDehazing}{GitHub} \\
        RRSHID$_{25}$~\cite{zhu2025real} & 3053 & Real & GF-PMS & 3 & $256\times256$ & Hazy & \href{https://github.com/AeroVILab-AHU/RRSHID?tab=readme-ov-file}{GitHub} \\
		\bottomrule
	\end{tabular}
    }
\end{table}

\section{Datasets and Evaluation Metrics}
\label{sec.Datasets}

\subsection{Remote Sensing Image Dehazing Datasets}

As reported in \tablename~\ref{tab:Datasets}, SateHaze1k~\cite{huang2020single} and RICE~\cite{lin2019remote} are the most frequently used in recent dehazing literature. 
The SateHaze1k includes 1,200 synthetic haze pairs based on physics-driven models, with corresponding clear images and SAR references. 
SateHaze1k is divided into three subsets according to haze density: 
thin (SH-TN), 
moderate (SH-M), and 
thick (SH-TK), 
each containing 400 image pairs. 
This structured degradation hierarchy enables fine-grained evaluation of model robustness across haze levels.

RICE datasets, a real-world benchmark, comprises two subsets: RICE1 contains 500 cloudy/clear image pairs; RICE2 includes 736 triplets of the cloudy image, cloud-free reference, and pixel-level mask. Its uniform resolution ($512\times512$) and diversity in cloud density make it ideal for learning-based training and quantitative comparison~\cite{zhang2020thick}.
LHID and DHID~\cite{zhang2022dense} are synthetic datasets with $512\times512$ pixels per image. 
LHID contains 30,517 image pairs for training and 1,000 image pairs for testing (LHID-A and LHID-B). 
RSID~\cite{chi2023trinity}, with 1,000 pairs of haze-simulated RSIs, focuses on military and tactical scenes such as shipyards and airports. 
HRSI~\cite{liu2024oriented} is a purely real-world hazy dataset, including 796 unpaired samples collected under diverse weather and viewing conditions. 
The images range from $512\times512$ to $4000\times4000$ in size and cover categories such as ships, airports, and large vehicles, presenting challenges in both spatial detail and structural consistency.
RRSHID~\cite{zhu2025real} is a pioneering large-scale real-world benchmark that is also divided into three subsets according to haze density: thin (RRSHID-TN), moderate (RRSHID-M), and thick (RRSHID-TK), comprising a total of 3,053 image pairs. 
{RRSHID provides hazy/haze-free image pairs obtained by multi-temporal acquisitions of the same geographical area. 
The haze-free reference images are selected from satellite observations with clearer atmospheric conditions within 1-3 months and are precisely aligned using geospatial metadata~\cite{zhu2025real}.}
The spectral characteristics exhibit complex color distortions and spectral shifts in the real world.

\subsection{Evaluation Metrics}

\subsubsection{Image Quality Evaluation Metrics}

\textbf{Peak Signal-to-Noise Ratio (PSNR).}  
PSNR~\cite{hore2010image} quantifies the pixel-level reconstruction fidelity by measuring the logarithmic ratio between the maximum signal intensity and the mean squared error (MSE):
\begin{equation}
\operatorname{PSNR}
= 10 \log_{10}
\frac{\operatorname{MAX}_{I}^{2}}{\operatorname{MSE}},
\end{equation}
where $\operatorname{MAX}_{I}$ denotes the maximum possible pixel value (255 for 8-bit images).  
The mean squared error (MSE) is computed as:
\begin{equation}
\operatorname{MSE}
= \frac{1}{N}\sum_{i=1}^{N}(x_i - y_i)^2,
\end{equation}
where $x_i$ and $y_i$ denote the intensity (or spectral vector) of pixel $i$ in the dehazed and reference images, and $N$ is the total number of pixels.  
Higher PSNR indicates smaller reconstruction error and finer pixel-level restoration.

\textbf{Structural Similarity Index Measure (SSIM).}  
SSIM~\cite{wang2004image} assesses perceptual similarity by jointly comparing local luminance, contrast, and structural information:
\begin{equation}
\operatorname{SSIM}(\boldsymbol{I}_{dehazed}, \boldsymbol{I}_{clear})
=
\frac{(2\mu_{d}\mu_{c}+C_1)(2\sigma_{dc}+C_2)}
{(\mu_{d}^{2}+\mu_{c}^{2}+C_1)(\sigma_{d}^{2}+\sigma_{c}^{2}+C_2)},
\end{equation}
where $\mu_{d}$ and $\mu_{c}$ are local means, $\sigma_{d}^{2}$ and $\sigma_{c}^{2}$ are local variances,  
and $\sigma_{dc}$ is the covariance between the two images.  
Constants $C_1$ and $C_2$ ensure numerical stability.  
Compared with PSNR, SSIM is more aligned with human perception by emphasizing structural continuity.

\textbf{CIEDE2000 Color-difference (CIEDE).}  
CIEDE2000~\cite{sharma2005ciede2000} evaluates color differences in the perceptually uniform CIELAB space:
\begin{equation}
\operatorname{CIEDE}
= \sqrt{
\left(\frac{\Delta L'}{k_L S_L}\right)^2
+
\left(\frac{\Delta C'_{ab}}{k_C S_C}\right)^2
+
\left(\frac{\Delta H'_{ab}}{k_H S_H}\right)^2
+
R_T
\left(\frac{\Delta C'_{ab}}{k_C S_C}\right)
\left(\frac{\Delta H'_{ab}}{k_H S_H}\right)
},
\end{equation}
where $\Delta L'$, $\Delta C'_{ab}$, and $\Delta H'_{ab}$ represent differences in lightness, chroma, and hue.  
$S_L$, $S_C$, and $S_H$ are perceptual weighting functions,  
$R_T$ is a rotation term reflecting hue–chroma interaction,  
and $k_L$, $k_C$, and $k_H$ are empirical coefficients (typically set to 1).  
Lower CIEDE reflects more faithful color restoration, complementing the luminance–contrast evaluation of SSIM.

\textbf{Learned Perceptual Image Patch Similarity (LPIPS).}  
LPIPS~\cite{zhang2018unreasonable} measures perceptual similarity by comparing deep features extracted from pretrained neural networks:
\begin{equation}
\operatorname{LPIPS}(\boldsymbol{I}_{dehazed}, \boldsymbol{I}_{clear})
=
\sum_{l}
\frac{1}{H_l W_l}
\sum_{h,w}
\left\|
w_l \odot \left( \hat{y}_{hw}^{l} - \hat{y}_{0hw}^{l} \right)
\right\|_2^{2},
\end{equation}
where $l$ indexes feature layers,  
$H_l \times W_l$ is the spatial resolution of layer $l$,  
$\hat{y}_{hw}^{l}$ and $\hat{y}_{0hw}^{l}$ denote normalized feature activations,  
and $w_l$ is a learned channel-wise weighting vector.  
Lower LPIPS indicates higher perceptual fidelity, capturing visual realism not reflected by PSNR or SSIM.

\textbf{Fréchet Inception Distance (FID).}  
FID~\cite{heusel2017gans} evaluates realism by comparing the feature distributions of dehazed and reference images in an embedding space:
\begin{equation}
\operatorname{FID}(\boldsymbol{I}_{dehazed}, \boldsymbol{I}_{clear})
=
\left\|\mu_{d}-\mu_{c}\right\|^{2}
+ \operatorname{Tr}\left(
\Sigma_{d} + \Sigma_{c}
- 2\sqrt{\Sigma_{d}\Sigma_{c}}
\right),
\end{equation}
where $\mu_d$ and $\mu_c$ are mean feature vectors of the two sets,  
$\Sigma_d$ and $\Sigma_c$ denote the corresponding covariance matrices,  
and $\operatorname{Tr}(\cdot)$ is the matrix trace.  
FID captures distribution-level consistency, providing a complementary measure to pixel- and structure-based metrics.

\subsubsection{Consistency and No-reference Metrics}

% ---------------- SAM -----------------
\textbf{Spectral Angle Mapper (SAM).}  
SAM~\cite{dennison2004comparison} measures the angular difference between spectral vectors, reflecting pixel-wise spectral fidelity. A smaller SAM indicates better preservation of spectral signatures:
\begin{equation}
\operatorname{SAM}(\boldsymbol{I}_{dehazed}, \boldsymbol{I}_{clear})
= \frac{180}{\pi} \frac{1}{N} \sum_{i=1}^{N} 
\arccos \left(
\frac{x_i \cdot y_i}{\|x_i\| \, \|y_i\| + \epsilon}
\right),
\end{equation}
where $x_i$ and $y_i$ are the spectral vectors of pixel $i$ in $\boldsymbol{I}_{dehazed}$ and $\boldsymbol{I}_{clear}$, $N$ is the number of pixels, and $\epsilon$ avoids division ambiguity.  
SAM focuses on \emph{spectral consistency} and is insensitive to absolute radiance magnitude.

% ---------------- ERGAS -----------------

\textbf{Error Relative Global Adimensional Synthesis (ERGAS).}  
ERGAS~\cite{irmak2018map} evaluates global radiometric distortion by comparing per-band RMSE values normalized by the band means. It complements SAM by assessing magnitude-related radiometric deviations:
\begin{equation}
\operatorname{ERGAS}(\boldsymbol{I}_{dehazed}, \boldsymbol{I}_{clear})
= \frac{100}{r} 
\sqrt{
\frac{1}{K} \sum_{k=1}^{K}
\left(
\frac{RMSE_k}{\mu_k + \epsilon}
\right)^2 },
\end{equation}
where $r$ is the resolution ratio between reference and target images,  
$K$ is the number of spectral bands,  
$RMSE_k$ is the root-mean-square error of band $k$,  
and $\mu_k$ is its mean radiance value.
Lower ERGAS implies better \emph{radiometric accuracy}.

% ---------------- UIQI -----------------
\textbf{Universal Image Quality Index (UIQI).}  
UIQI~\cite{liu2023uiqi} characterizes local luminance, contrast, and structure consistency—a complement to SAM and ERGAS by highlighting spatial structural fidelity:
\begin{equation}
\operatorname{UIQI}(\boldsymbol{I}_{dehazed}, \boldsymbol{I}_{clear})
= \frac{1}{N}\sum_{i=1}^{N}
\frac{
4 \mu_{x_i}\mu_{y_i}\sigma_{x_i y_i}
}{
(\mu_{x_i}^2 + \mu_{y_i}^2)(\sigma_{x_i}^2 + \sigma_{y_i}^2) + \epsilon },
\end{equation}
where $\mu_{x_i}$ and $\mu_{y_i}$ are local means,  
$\sigma_{x_i}^2$ and $\sigma_{y_i}^2$ are local variances,  
and $\sigma_{x_i y_i}$ is local covariance.
Higher UIQI indicates better preservation of fine spatial structures.

% ---------------- QNR -----------------
\textbf{Quality with No Reference (QNR).}  
QNR~\cite{wang2011reduced} jointly integrates spectral similarity (from SAM) and radiometric precision (from ERGAS) into a unified measure:
\begin{equation}
\operatorname{QNR}(\boldsymbol{I}_{dehazed}, \boldsymbol{I}_{clear})
= \exp\!\left( -\frac{\operatorname{SAM}}{10} \right) \cdot 
  \exp\!\left( -\frac{\operatorname{ERGAS}}{100} \right).
\end{equation}
Higher QNR reflects more consistent spectral–spatial characteristics.  
This metric theoretically links two physical criteria—spectral fidelity and radiometric accuracy—into a single score.

\begin{table*}[htbp]
	\centering
	\scriptsize
	\caption{
    Summary of representative remote sensing image dehazing methods, including publication venues, core designs, technical advantages, and existing limitations.
    }
	\label{tab:Comparison}
	\resizebox{\textwidth}{!}{%
		\begin{tabular}{p{2.5cm}p{1.6cm}p{5cm}p{4.8cm}p{4.8cm}}
			\toprule
			\textbf{Methods} & \textbf{Pub.} & \textbf{Overview} & \textbf{Advantages} & \textbf{Limitations} \\
			\midrule
SMIDCP~\cite{li2018haze} & GRSL 2018 & A sphere model to improve the DCP and combining it with HF for image enhancement. & Strong capability in handling inhomogeneous conditions, dense haze, and thin clouds. & Limited adaptability to extreme scenarios. \\
MDCP~\cite{shi2021novel} & GRSL 2021 & Integrate DCP into the multiscale transform and process images by selecting appropriate fusion rules and an optimized Laplacian model. & Effectively removes thin clouds and supports multi-temporal reference images. & Long processing time and highly dependent on the quality of reference images. \\ \midrule
RSC-Net~\cite{li2019thin} & ISPRS P\&RS 2019 & Employs the encoding-decoding framework to enable end-to-end thin cloud removal. & Symmetrical concatenation effectively preserves details of cloud-free regions. & Relies on specific sensor band configurations and lacks validation of adaptability to complex atmospheric conditions. \\
RSDehazeNet~\cite{guo2020rsdehazenet} & TGRS 2020 & An end-to-end network combining channel attention mechanism and global residual learning. & Fast processing speed, preserves image details, and high realism of synthetic data. & The inadequate adaptability to extreme conditions or complex terrain. \\
FCTF~\cite{li2020coarse} & GRSL 2020 & Adopting a coarse-to-fine two-stage architecture to refine dehazing results and optimize image details. & Improvement in dehazing accuracy under complex scenes. & Insufficient generalization capability to complex haze distributions and diverse ground surface characteristics. \\
CNNIM~\cite{zi2021thin} & JSTARS 2021 & Utilizing U-Net and Slope-Net to subtract the thin cloud thickness map and obtain a clear image. & Capable of acquiring thin cloud information at different altitudes and for each band. & Exhibits significant cumulative estimation errors for real-world images. \\
MSDA-CR~\cite{yu2022cloud} & GRSL 2022 & Simulating the effects of cloud reflection through cloud distortion control functions (CDCFs). & Enhancing adaptability to different cloud thicknesses and distributions. & Relies on paired dataset training, weak generalization. \\
GLF-CR~\cite{xu2022glf} & ISPRS P\&RS 2023 & Achieves cloud removal via a global-local fusion framework. & Maintains geometric consistency with cloud-free regions while enhancing SAR data utilization efficiency. & Relies on the registration accuracy between SAR and optical images. \\
SMDCNet~\cite{tu2025cloud} & ISPRS P\&RS 2025 & Integrating complementary SAR information via multimodal similarity attention. & Significantly enhances detail recovery under heterogeneous cloud. & Sensitive to sensor discrepancies. \\
HPN-CR~\cite{gu2025hpn} & TGRS 2025 & Adopting a heterogeneous encoder, cloud removal is achieved through feature fusion and a pixelshuffle-based decoder. & Adopting multi-module collaboration to enhance feature extraction capabilities. & Structural clarity in extreme scenarios remains improvable. \\ \midrule
CR-GAN-PM~\cite{li2020thin} & ISPSR P\&RS 2020 & The separation of cloud-contaminated images through a decomposition-reconstruction framework. & Fully preserves spectral information without requiring paired data. & Degraded restoration performance in complex scenarios. \\
UCR~\cite{zheng2020single} & TGRS 2020 & Distinguishing between thin and thick cloud regions through staged processing. & Computational efficiency and restoration quality are improved. & High thick cloud coverage ratio challenges generator in accurately recovering complex textural details. \\
SAR2Opt-GAN-CR~\cite{darbaghshahi2021cloud} & TGRS 2021 & Adopting a two-stage GAN framework. & Expanded receptive field and improved output image quality. & Strong dependence on SAR data. \\ 
Dehaze-AGGAN~\cite{zheng2022dehaze} & TGRS 2022 & Adopting the Cycle-GAN~\cite{zhu2017unpaired} architecture with an enhanced attention-guided generator and discriminator. & Improved dehazing details and structural preservation capabilities without requiring paired training data. & Complex model architecture incurs high computational costs. \\
MT\_GAN~\cite{wang2025mt_gan} & ISPSR P\&RS 2025 & Employs a multilayer translation GAN integrated with a despeckling module to eliminate speckle noise in SAR. & Unsupervised multi-scale feature fusion with prominent capability. & Suboptimal SAR geometric distortion mitigation and computational efficiency. \\ \midrule
RSDformer~\cite{song2023learning} & GRSL 2023 & Improving image content restoration by capturing dependency relationships in global and local regions. & Reduced computational cost while enhancing detail and structure recovery capability. & High dependency on specific loss functions and unexplored multimodal data fusion. \\
Trinity-Net~\cite{chi2023trinity} & TGRS 2023 & Performing image restoration by integrating the advantages of prior-based and deep learning strategies. & High scalability for multi-task learning. & High computational complexity and limited capability in processing low-light scenarios. \\
DCR-GLFT~\cite{quan2024density} & TGRS 2024 & Leveraging density labels to guide the cloud removal process. & Adaptive processing of different cloud densities while preserving surface information. & Dependence on the accuracy of density labels and high algorithmic complexity. \\
SSGT~\cite{du2024ssgt} & JSTARS 2024 & Proposing a dual-branch fusion framework to restore the details of areas covered by clouds. & Recovering details in thick cloud-covered regions via non-local similarity relationships. & Limited performance due to cloud thickness and reliance on multi-temporal, multi-source data. \\
PGSformer~\cite{dong2024prompt} & GRSL 2024 & Employing an inverse cognitive learning network for multi-scale image feature extraction. & Enhanced detail recovery capability and attention precision. & High parameter tuning complexity and heavy reliance on computational resources. \\
DehazeXL~\cite{chen2025tokenize} & CVPR 2025 & A dehazing method capable of effectively balancing global context and local feature extraction. & Efficient processing of large, high-resolution images. & Heavy dependence on high-quality training data and performance limitations in extreme haze scenarios. \\ \midrule
ARDD~\cite{huang2023remote} & GRSL 2023 & Primarily achieving progressive restoration of clear images through multiple denoising steps. & Restored images exhibit stable quality and support multi-scale inputs. & Slow inference speed, and overlapping regions may lead to redundant computations. \\
ADND~\cite{huang2024diffusion} & GRSL 2024 & Optimizing the reverse diffusion process via a range-null-space decomposition strategy. & Supporting multi-scale inputs with enhanced image consistency. & Slow inference speed and high computational resource requirements. \\
RSHazeDiff~\cite{xiong2024rshazediff} & TITS 2024 & Improve image perceptual quality in dense hazy scenarios by leveraging the generative capability of conditional DDPM. & The outstanding detail recovery capability in dense hazy scenarios. & Slow inference speed and potential generalization limitations due to insufficient real-world scene data. \\
IDF-CR~\cite{wang2024idf} & TGRS 2024 & An iterative diffusion process for cloud removal to achieve component divide-and-conquer  cloud removal. & Enhanced noise prediction accuracy and detail recovery capability. & Performance degradation in the absence of ground information guidance. \\ 
EMRDM~\cite{liu2025effective} & CVPR 2025 & Develop MRDMs to establish a direct diffusion process between cloudy and cloud-free images. & Enhanced SAR-optical data fusion capability. & High training and inference costs, and unstable restoration performance in low-texture regions. \\

\bottomrule
\end{tabular}
}
\end{table*}

% ---------------- NIQE -----------------
\textbf{Natural Image Quality Evaluator (NIQE).}  
NIQE~\cite{zhang2015feature} is a no-reference metric capturing naturalness by modeling the Mean Subtracted Contrast Normalized (MSCN) coefficient distribution~\cite{liu2019unsupervised}:
\begin{equation}
\operatorname{NIQE}(\boldsymbol{I}_{dehazed}) = |\hat{\alpha}| + |\hat{\beta}|,
\end{equation}
where $\hat{\alpha}$ and $\hat{\beta}$ are the fitted shape parameters of MSCN statistics.  
Lower NIQE signifies that the dehazed image exhibits more natural visual statistics, complementing reference-based indicators like SAM/ERGAS.

% ---------------- HIST -----------------
\textbf{Histogram Similarity (HIST).}  
HIST~\cite{min2019quality} measures grayscale consistency by comparing normalized histogram distributions:
\begin{equation}
\operatorname{HIST}(\boldsymbol{I}_{dehazed}, \boldsymbol{I}_{clear})
= 1 - \frac{1}{2} \sum_{i=0}^{255}
\left| h_i - g_i \right|,
\end{equation}
where $h_i$ and $g_i$ are histogram bins of the two images.  
A higher HIST value indicates better global tone preservation, serving as an intuitive complement to SAM, ERGAS and UIQI.

\section{Methods Analysis and Benchmark Comparison}
\label{sec.benchmark}
\subsection{Summary and Analysis of Methods: Strengths and Weaknesses}
\tablename~\ref{tab:Comparison} provides a structured summary of representative RSIs dehazing methods published in recent years. 

\textbf{1) Traditional and Deep Convolution Methods:}
As shown in \tablename~\ref{tab:Comparison}, early approaches~\cite{shi2021novel,li2020coarse} rely on handcrafted priors or multi-stage feature learning. 
While these methods offer interpretable designs and decent performance in controlled scenarios, their effectiveness is often constrained by poor generalizability, limited structural recovery, and heavy reliance on paired data. 
CNN-based models~\cite{guo2020rsdehazenet,zi2021thin} improve spatial fidelity and haze removal precision by learning hierarchical features, but remain sensitive to training distribution and require high computational resources.

\begin{table}[t]
	\centering
	\caption{
		Quantitative performance at PSNR (dB) and SSIM of RSIs dehazing algorithms evaluated on the SateHaze1k and RICE datasets. 
		\textcolor{red}{Red} and \textcolor{blue}{blue} values indicate the best and second-best performance for each metric, respectively.
	}
	\label{tab:Performance}
	\resizebox{\textwidth}{!}{%
		\begin{tabular}{llcccccccc}
			\toprule
			\multirow{2}{*}{\textbf{Methods}} & \multirow{2}{*}{\textbf{Category}} & \multicolumn{2}{c}{\textbf{SH-TN}} & \multicolumn{2}{c}{\textbf{SH-M}} & \multicolumn{2}{c}{\textbf{SH-TK}} & \multicolumn{2}{c}{\textbf{RICE}} \\ \cmidrule(l){3-4}  \cmidrule(l){5-6}  \cmidrule(l){7-8}  \cmidrule(l){9-10}  
			&   & {PSNR $\uparrow$}        & {SSIM $\uparrow$}   & {PSNR $\uparrow$}    & {SSIM $\uparrow$}   & {PSNR$ \uparrow $}       & {SSIM $\uparrow$}     & {PSNR $\uparrow$}    & {SSIM $\uparrow$}     \\ \midrule
            DHIM$_{15}$~\cite{pan2015haze} & Traditional & 19.445 & 0.891 & 19.916 & 0.917 & 16.595 & 0.810 & 19.240 & 0.882 \\
            GRS-HTM$_{17}$~\cite{liu2017haze} & Traditional & 15.489 & 0.762 & 15.071 & 0.784 & 10.473 & 0.462 & 18.278 & 0.825 \\
			SMIDCP$_{18}$~\cite{li2018haze} & Traditional & 13.639 & 0.833 & 15.990 & 0.863 & 14.956 & 0.757 & 16.573 & 0.712 \\
            IDeRs$_{19}$~\cite{xu2019iders} & Traditional & 15.048 & 0.772 & 14.763 & 0.785 & 11.754 & 0.702 & 15.750 & 0.611 \\
			EVPM$_{22}$~\cite{han2022local} & Traditional & 20.426 & 0.891 & 20.656 & 0.918 & 16.647 & 0.787 & 15.217 & 0.742 \\
			SRD$_{23}$~\cite{he2023remote} & Traditional & 21.327 & 0.896 & 20.774 & 0.930 & 17.265 & 0.814 & 20.550 & 0.926 \\
			FCTF-Net$_{20}$~\cite{li2020coarse} & CNN & 23.590 & 0.913 & 22.880 & 0.927 & 20.030 & 0.816 & 25.535 & 0.870 \\
            DCIL$_{22}$~\cite{zhang2022dense}& CNN & 22.170 & 0.902 & 24.972 & 0.942 & 20.707 & 0.845 & 27.720 & 0.876 \\
            PSMB-Net$_{23}$~\cite{sun2023partial} & CNN & 23.533 & 0.909 & 27.037 & 0.942 & 21.087 & 0.839 & 28.057 & 0.893 \\
			EMPF-Net$_{23}$~\cite{wen2023encoder} & CNN & \textcolor{red}{27.400} & 0.960 & \textcolor{red}{31.450} & \textcolor{blue}{0.975} & 26.330 & 0.928 & 35.845 & \textcolor{red}{0.979} \\
            LFD-Net$_{23}$~\cite{jin2023lfd} & CNN & 18.202 & 0.858 & 17.537 & 0.598 & 17.346 & 0.773 & - & - \\
            MAP-Net$_{23}$~\cite{xu2023video}	& CNN	&13.265	&0.424	&14.296	&0.422	&12.470	&0.268	&14.344	&0.649 \\
			SFAN$_{24}$~\cite{shen2024spatial} & CNN & 23.688 & \textcolor{red}{0.963} & \textcolor{blue}{28.191} & \textcolor{red}{0.977} & 23.006 & \textcolor{red}{0.942} & 35.374 & 0.941 \\
			ICL-Net$_{24}$~\cite{dong2024icl} & CNN & 24.590 & 0.923 & 25.670 & 0.937 & 21.780 & 0.859 & \textcolor{red}{36.940} & \textcolor{blue}{0.960} \\
			EDED-Net$_{24}$~\cite{dong2024end} & CNN & 24.605 & 0.893 & 25.369 & 0.913 & 22.418 & 0.846 & 31.907 & 0.945 \\
            DVD$_{24}$~\cite{fan2024driving}		& CNN &13.134	&0.649	&12.983	&0.673	&8.960	&0.356	&23.295	&0.826 \\
            SpA-GAN$_{20}$~\cite{pan2020cloud} & GAN & 23.131 & 0.902 & 18.629 & 0.555 & 17.964 & 0.739 & - & - \\
            DUVD$_{24}$~\cite{yang2024depth}	& GAN	&15.698 &0.838	&15.896	&0.849	&10.879	&0.635	&19.042	&0.549 \\
			TransRA$_{22}$~\cite{dong2022transra} & Transformer & 25.200 & 0.930 & 26.500 & 0.947 & 22.730 & 0.875 & 31.130 & 0.955 \\
            DehazeFormer$_{23}$~\cite{song2023vision} & Transformer & 21.970 & 0.953 & 25.550 & 0.970 & 20.370 & 0.916 & - & - \\
            Trinity-Net$_{23}$~\cite{chi2023trinity} & Transformer & 24.716 & \textcolor{blue}{0.961} & 25.866 & 0.959 & 21.076 & 0.923 & 29.248 & 0.908 \\
            RSDformer$_{23}$~\cite{song2023learning} & Transformer & 24.210 & 0.912 & 26.241 & 0.934 & 23.011 & 0.853 & 33.013 & 0.953 \\
			PGSformer$_{24}$~\cite{dong2024prompt} & Transformer & 25.534 & 0.918 & 26.622 & 0.933 & 23.596 & 0.863 & 34.404 & 0.948 \\
            ASTA$_{24}$~\cite{cai2024additional} & Transformer & 23.730 & \textcolor{red}{0.963} & 25.060 & 0.968 & 21.390 & 0.926 & - & - \\
            DehazeXL$_{25}$~\cite{chen2025tokenize} & Transformer & 17.233 & 0.759 & 16.470 & 0.428 & 14.299 & 0.544 & - & - \\
			ARDD-Net$_{23}$~\cite{huang2023remote} & Diffusion  & 26.840 & 0.926 & 26.470 & 0.932 & \textcolor{blue}{26.830} & 0.932 & -  & -  \\
			ADND-Net$_{24}$~\cite{huang2024diffusion} & Diffusion  & \textcolor{blue}{26.910} & 0.927 & 26.670 & 0.936 & \textcolor{red}{26.940} & \textcolor{blue}{0.936} & - & - \\
			RSHazeDiff$_{24}$~\cite{xiong2024rshazediff} & Diffusion  & - & - & - & - & - & - & \textcolor{blue}{36.560} & 0.958 \\
			\bottomrule
		\end{tabular}
	}
\end{table}

\textbf{2) Adversarial Generation Methods:}
As shown in the third block of  \tablename~\ref{tab:Comparison}, adversarial generation methods have been extensively explored for remote sensing haze removal, such as
Dehaze-AGGAN~\cite{zheng2022dehaze}, CR-GAN-PM~\cite{li2020thin}, SAR2Opt-GAN-CR~\cite{darbaghshahi2021cloud}. 
Adversarial generation models exploit adversarial loss and style transfer to produce perceptually convincing outputs, and CR-GAN-PM explicitly incorporates physical priors or multi-modal constraints. 
While GANs demonstrate superior visual quality and training flexibility, their instability during optimization and mode collapse under complex distributions remain ongoing challenges.

\textbf{3) Vision Transformer Methods:}
The introduction of ViT structures (\textit{e.g.}, Trinity-Net~\cite{chi2023trinity}, DCR-GLFT~\cite{quan2024density}, PGSformer~\cite{dong2024prompt}) provides enhanced global context modeling and long-range dependency capture. 
These models exhibit superior performance in challenging environments such as wide-area haze and non-uniform clouds. 
Hybrid designs like Former-CR~\cite{han2023former} and RSDformer~\cite{song2023learning} integrate convolutional backbones with attention modules, achieving a balance between efficiency and expressiveness. 
However, transformer-based approaches still face limitations in handling very high-resolution images due to memory constraints.

\textbf{4) Diffusion Generation Methods:}
As listed in \tablename~\ref{tab:Comparison},
diffusion methods (\textit{e.g.}, ARDD~\cite{huang2023remote}, ADND~\cite{huang2024diffusion}, RSHazeDiff~\cite{xiong2024rshazediff}, EMRDM~\cite{liu2025effective}) represent a promising frontier. 
Diffusion dehazing models progressively denoise hazy images using probabilistic sampling and Markov-chain modeling. 
Diffusion-based approaches show exceptional robustness in preserving structural consistency and suppressing over-saturation. 
Despite their superior dehazing quality, diffusion models often suffer from high inference latency and high computational cost, making real-time deployment challenging.

% SH数据集上新增指标
\begin{table*}[t]
\centering
\caption{
Quantitative performance at SAM, ERGAS, UIQI, QNR. NIQE and HIST on the SateHaze1k datasets. 
\textcolor{red}{Red} and \textcolor{blue}{blue} values indicate the best and second-best performance for each metric, respectively.
}
\label{tab:SateHaze1k_SAM}
\resizebox{\textwidth}{!}{%
\setlength{\tabcolsep}{2.2pt}
\renewcommand{\arraystretch}{1}
\begin{tabular}{ccccccccc}
\toprule
\textbf{Metrics} &
\textbf{SpA-GAN~\cite{pan2020cloud}} &
\textbf{DCIL~\cite{zhang2022dense}} &
\textbf{DehazeFormer~\cite{song2023vision}} &
\textbf{PSMB-Net~\cite{sun2023partial}} &
\textbf{Trinity-Net~\cite{chi2023trinity}} &
\textbf{LFD-Net~\cite{jin2023lfd}} &
\textbf{ASTA~\cite{cai2024additional}} &
\textbf{DehazeXL~\cite{chen2025tokenize}} \\
\midrule
%--------------- ERGAS ------------------------
\multirow{3}{*}{ERGAS $\downarrow$}
& 13.386 & 14.473 & 18.127 & \textcolor{blue}{13.075} & 22.765 & 23.813 & \textcolor{red}{12.188} & 26.444 \\
& 32.128 & 30.591 & 29.493 & \textcolor{blue}{28.706} & 34.624 & 34.873 & \textcolor{red}{11.727} & 37.214 \\
& 24.088 & 17.574 & \textcolor{blue}{16.013} & 16.808 & 22.293 & 25.929 & \textcolor{red}{15.343} & 36.951 \\
\midrule

% ------------------------ UIQI ------------------------
\multirow{3}{*}{UIQI $\uparrow$} 
& 0.902 & 0.904 & 0.889 & \textcolor{blue}{0.911} & 0.650 & 0.858 & \textcolor{red}{0.913} & 0.759 \\
& 0.555 & 0.621 & \textcolor{blue}{0.624} & \textcolor{blue}{0.624} & 0.425 & 0.598 & \textcolor{red}{0.944} & 0.428 \\
& 0.739 & \textcolor{blue}{0.843} & 0.837 & 0.841 & 0.650 & 0.773 & 0.845 & 0.544 \\
\midrule
% ------------------------ QNR ------------------------
\multirow{3}{*}{QNR $\uparrow$} 
& 0.670 & \textcolor{blue}{0.677} & 0.641 & 0.652 & 0.601 & 0.594 & \textcolor{red}{0.696} & 0.384 \\
& 0.501 & 0.519 & 0.543 & \textcolor{blue}{0.571} & 0.525 & 0.433 & \textcolor{red}{0.672} & 0.497 \\
& 0.542 & \textcolor{blue}{0.633} & 0.630 & 0.600 & 0.565 & 0.476 & \textcolor{red}{0.635} & 0.496 \\
\midrule
% ------------------------ NIQE ------------------------
\multirow{3}{*}{NIQE $\downarrow$}
& 1.691 & \textcolor{blue}{1.689} & 1.771 & 1.766 & \textcolor{red}{1.576} & 1.786 & 1.724 &  1.903 \\
& 2.135 & 1.924 & 1.986 & 1.920 & \textcolor{blue}{1.606} & 2.087 & 1.924 & \textcolor{red}{1.380} \\
& 1.891 & 1.611 & 1.805 & 1.644 & \textcolor{blue}{1.561} & 1.914 & 1.665 & \textcolor{red}{1.146} \\
\midrule
% ------------------------ HIST ------------------------
\multirow{3}{*}{HIST $\uparrow$} 
& \textcolor{blue}{0.908} & 0.875 & 0.842 & 0.898 & 0.806 & 0.783 & \textcolor{red}{0.925} & 0.800 \\
& \textcolor{blue}{0.915} & 0.871 & 0.885 & 0.905 & 0.821 & 0.856 & \textcolor{blue}{0.914} & 0.819 \\
& 0.845 & 0.874 & \textcolor{blue}{0.925} & 0.884 & 0.876 & 0.887 & \textcolor{red}{0.930} & 0.663 \\
\midrule
% ------------------------ Rank------------------------
{${R} \downarrow$} 
& 5.121 & 3.606 & 3.697 & \textcolor{blue}{3.242} & 4.909 & 6.636 & \textcolor{red}{1.636} & 7.091 \\
% \midrule
% ------------------------ Ranking gaps ------------------------
{$ \Delta {R} \downarrow$} 
& 3.485 & 1.970 & 2.061 & \textcolor{blue}{1.606} & 3.273 & 5.000 & \textcolor{red}{0.000} & 5.455 \\
% \midrule
% ------------------------ statistical significance -----------------------
{${p} \downarrow$}
& 0.113 & 0.852 & 0.114 & 0.053 & 0.010$^*$ & 0.109 & - & 0.208 \\
\bottomrule
\multicolumn{9}{l}{\small $\bullet$ Each row corresponds to results under thin haze, moderate haze, and thick haze, respectively.}\\
\end{tabular}
} % end resizebox
\end{table*}

\subsection{Large-scale Remote Sensing Dehazing Benchmark}

\subsubsection{Quantitative Comparison and Analysis}

\tablename~\ref{tab:Performance} shows clear performance differences across haze levels. Under thin haze, CNN models achieve the highest fidelity, with EMPF-Net reaching 27.400 dB PSNR and 0.960 SSIM, while traditional priors such as SRD and DHIM still perform reasonably well. 
Light haze preserves most spatial structures, allowing simple contrast constraints to remain effective.
As haze becomes moderate, traditional methods decline rapidly, typically falling below 20 dB PSNR. 
Deep networks maintain strong reconstruction, with EMPF-Net reaching 31.450 dB PSNR and SFAN exceeding 0.977 SSIM. 
Deep learning methods advantage arises from multi-scale feature extraction and stronger global–local context modeling, which better capture spatially varying haze.
Under thick haze, the gap widens further.
Traditional approaches lose structural detail and color stability, whereas deep models such as DCIL and PSMB-Net retain higher fidelity with SSIM values above 0.88. 
CNN-based methods recover severely attenuated features through richer representation learning.
As shown in \tablename~\ref{tab:Performance}, Diffusion models show the strongest generalization to real scenes. 
RSHazeDiff reaches 36.560 dB PSNR on RICE, outperforming all other methods. Its iterative refinement better handles irregular aerosol patterns and complex illumination.

\begin{table}[t]
	\centering
	\caption{
    Quantitative performance at CIEDE ($ \downarrow $), LPIPS ($ \downarrow $), FID ($ \downarrow $) PSNR ($ \uparrow $) and SSIM ($ \uparrow $) of SOTA methods evaluated on the LHID and DHID datasets.
    \textcolor{red}{Red} and \textcolor{blue}{blue} values indicate the best and second-best performance for each metric, respectively.
	}
	\label{tab:LHID_res}
	\resizebox{\textwidth}{!}{%
		\begin{tabular}{llcccccccccc}
			\toprule
			\multirow{2}{*}{\textbf{Methods}} & \multirow{2}{*}{\textbf{Category}} & \multicolumn{5}{c}{\textbf{LHID}} & \multicolumn{5}{c}{\textbf{DHID}} \\ \cmidrule(l){3-7}  \cmidrule(l){8-12}  
			&   & {CIEDE $\downarrow $}        & {LPIPS $\downarrow $}   & {FID $\downarrow$}  & {PSNR $\uparrow $} & {SSIM $\uparrow $} & {CIEDE $\downarrow$}        & {LPIPS $\downarrow$}   & {FID $\downarrow$} & {PSNR $\uparrow$} & {SSIM $\uparrow$}    \\ \midrule
		  DCP$_{10}$~\cite{he2010single} & Traditional & 9.161 & 0.481 & 21.490 & 21.120 & 0.818 & 9.867 & 0.163 & 79.920 & 18.920 & 0.824 \\
            IDeRs$_{19}$~\cite{xu2019iders} & Traditional & 19.563 & 0.633 & 44.650 & 14.680 & 0.627 & 19.942 & 0.344 & 120.17 & 13.330 & 0.575   \\
            FCTF-Net$_{20}$~\cite{li2020coarse} & CNN & 5.423 & 0.459 & 21.210 & 25.850 & 0.876 & 12.794 & 0.221 & 128.490 & 17.210 & 0.739   \\
            DCIL$_{22}$~\cite{zhang2022dense} & CNN & 5.180 & \textcolor{blue}{0.447} & \textcolor{blue}{14.930} & 28.120 & 0.898 & 4.877 & 0.131 & 64.730 & 26.950 & \textcolor{blue}{0.896}    \\
            PSMB-Net$_{23}$~\cite{sun2023partial} & CNN & 5.864 & 0.454 & 16.130 & \textcolor{blue}{28.890} & \textcolor{blue}{0.900} & 4.965 & \textcolor{blue}{0.113} & \textcolor{blue}{53.830} & \textcolor{blue}{27.070} & 0.887   \\
            AU-Net$_{24}$~\cite{du2024dehazing} & CNN & \textcolor{blue}{4.045} & 0.451 & 16.960 & 28.440 & 0.893 & \textcolor{blue}{4.489} & 0.138 & 69.28 & 26.420 & 0.891  \\
            Trinity-Net$_{23}$~\cite{chi2023trinity} & Transformer & 6.737 & 0.460 & 17.760 & 24.780 & 0.867 & 5.479 & 0.139 & 75.460 &25.860 & 0.878   \\
            RSHazeDiff$_{24}$~\cite{xiong2024rshazediff} & Diffusion & \textcolor{red}{3.875} & \textcolor{red}{0.436} & \textcolor{red}{12.850} & \textcolor{red}{29.650} & \textcolor{red}{0.905} & \textcolor{red}{3.945} & \textcolor{red}{0.102} & \textcolor{red}{49.520} & \textcolor{red}{27.910} & \textcolor{red}{0.900}   \\
			\bottomrule
            \multicolumn{9}{l}{\small $\bullet$ $\uparrow$ and $\downarrow$ indicate that higher values are better and lower values are better, respectively.}\\
		\end{tabular}
	}
\end{table}

Furthermore, we compared the adaptability of video dehazing methods on remote sensing data. 
On the SH-M datasets, TransRA achieved a PSNR exceeding 26.500 dB and an SSIM score of 0.947. 
DUVD achieved a PSNR exceeding 15.896 dB; DVD achieved a PSNR exceeding 12.983 dB and 0.849 SSIM score. 
MAP-Net had the lowest overall score, with an SSIM below 0.422. 
On the RICE dataset, ICL-Net maintained competitive performance with the best PSNR of 36.940 dB and the second-best SSIM score of 0.960, while most other methods had PSNR below 36 dB.
\tablename~\ref{tab:SateHaze1k_SAM} demonstrates clear performance differences among model families across thin, moderate and thick haze. Physically guided and attention-based methods achieve the most stable radiometric behavior. 
Under thin haze, DCIL and ASTA reach SAM values of 2.462 and 2.416, which are improvements of approximately 10\% compared with most CNN methods. 
Under thick haze, ASTA obtains the lowest ERGAS of 15.343, which reduces global radiometric error by more than 25\% relative to conventional convolutional models whose ERGAS is typically above 16.808. 
ASTA ranks within the top two for UIQI, QNR, and HIST, indicating strong preservation of structural and spectral consistency. 
As reported in \tablename~\ref{tab:SateHaze1k_SAM}, DehazeXL and lightweight CNN models exhibit the most pronounced degradation with SAM rising to values between 3.345 and 4.869 and ERGAS reaching values between 25.929 and 36.951 under thick haze.
On SateHaze1k, ASTA attains the most stable behaviour with the lowest overall ranking score of 1.636, PSMB-Net attains the second–best performance with $\Delta \textit{R}$ of 1.606, while Trinity-Net is the only method that exhibits a statistically significant difference relative to ASTA. 
These quantitative trends of \tablename~\ref{tab:SateHaze1k_SAM} confirm that methods integrating global attention and physical constraints provide significantly higher consistency and robustness, while architectures relying mainly on local convolutions or adversarial sharpening struggle as haze density increases.

\begin{figure}[t]
	\centering 
	\includegraphics[width=13.5cm]{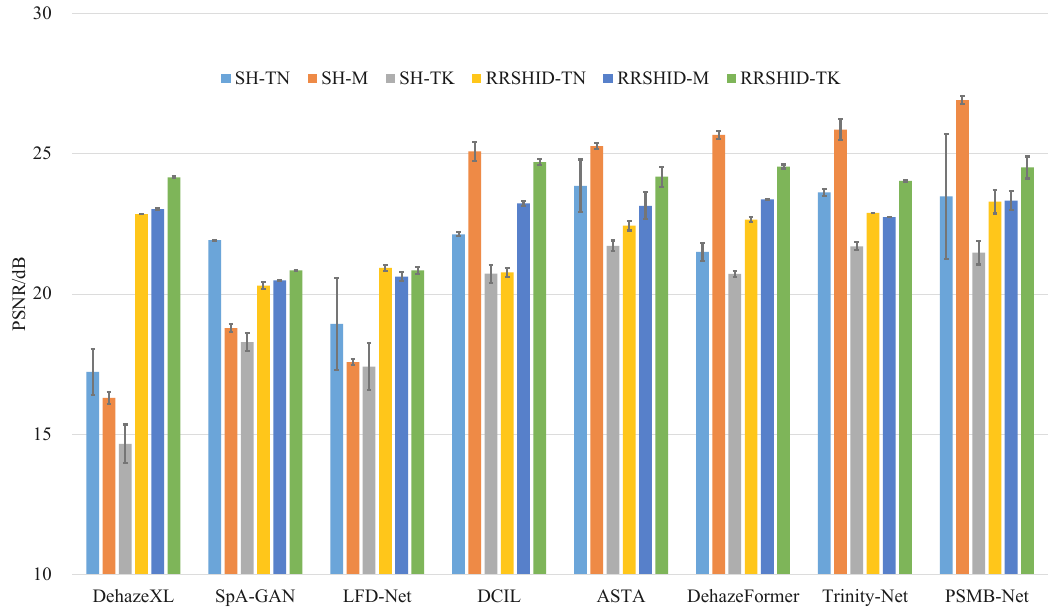}
	\caption{
    Quantitative comparison of PSNR (dB) across eight SOTA dehazing models on SateHaze1k (SH) and RRSIHID. The error bars indicate performance variance.
    }
	\label{Fig.PSNR}
\end{figure}

% --------- LHID and DHID结果分析
As shown in \tablename~\ref{tab:LHID_res}, the diffusion-based method RSHazeDiff consistently outperforms all other approaches across both LHID and DHID datasets. 
It achieves the lowest values in CIEDE, LPIPS, and FID, indicating the most accurate color dehazing, the highest perceptual quality, and the most natural image distribution. 
RSHazeDiff leads with a CIEDE of 3.875 and an FID of 12.850 on LHID, while also obtaining the best LPIPS of 0.102 on DHID. 
Among CNN-based methods, PSMB-Net shows strong performance, as reflected in \tablename~\ref{tab:LHID_res} by its second-best FID and the best LPIPS among CNNs on DHID, suggesting effective preservation of structure and texture. 
In contrast, as shown in \tablename~\ref{tab:LHID_res}, the Transformer-based Trinity-Net ranks in the middle range across all metrics, with no clear advantage over CNN counterparts. 
Traditional methods such as DCP and IDeRs exhibit the highest CIEDE and FID values, revealing poor generalization under complex degradation. 
The performance variations across LHID and DHID in \tablename~\ref{tab:LHID_res} further emphasize the importance of robustness to different scene conditions. 
These results underline the potential of diffusion models for RSIs dehazing and highlight the value of integrating their generative capabilities with the structural efficiency of CNNs.

To evaluate the robustness of different methods under varying haze conditions, we analyze PSNR and SSIM trends across SH-TN (thin haze), SH-M (moderate haze), and SH-TK (thick haze), as visualized in \figurename~\ref{Fig.PSNR}.
% and \figurename~\ref{Fig.SSIM}. 
Traditional methods exhibit significant degradation with increasing haze density. 
As shown in \figurename~\ref{Fig.PSNR}, GRS-HTM drops from 15.489 dB (SH-TN) to 10.473 dB (SH-TK), a decline of 5.016 dB, while IeRs decreases by 3.294 dB (from 15.048 to 11.754 dB), indicating their poor adaptability to nonlinear scattering effects in thick haze.
CNN-based EMPF-Net and PSMB-Net methods maintain higher stability. EMPF-Net drops only 1.07 dB (from 27.400 to 26.330 dB), while FCTF-Net declines by 1.673 dB, both showing less than 8\% relative degradation, highlighting their strong generalization and resilience to heavy haze.

\begin{table}[t]
\centering
\scriptsize     % -------- 调小表格字号（推荐）
\setlength{\tabcolsep}{10pt}
\renewcommand{\arraystretch}{1.15}
\caption{Model efficiency and performance comparison on SateHaze1k datasets.}
\label{tab:model_efficiency}
\begin{tabular}{lcccc}
\toprule
    \textbf{Methods} & \textbf{PSNR} & \textbf{FLOPs(G)} & \textbf{FPS} & \textbf{\#Param(M)} \\
    \midrule
    SpA-GAN$_{20}$~\cite{pan2020cloud}      & 19.664 &  67.931 &  29.84  &   0.210 \\
    DCIL$_{22}$~\cite{zhang2022dense}         & 22.641 & 107.445 &  52.35  &  26.509 \\
    DehazeFormer$_{23}$~\cite{song2023vision} & 22.624 &  24.817 &  41.48  &   0.686 \\
    PSMB-Net$_{23}$~\cite{sun2023partial}     & 23.953 & 392.754 &  18.54  &  12.488 \\
    Trinity-Net$_{23}$~\cite{chi2023trinity}  & 23.723 & 123.889 &  14.40  &  20.139 \\
    LFD-Net$_{23}$~\cite{jin2023lfd}      & 17.975 &  23.352 & \textbf{269.07}  &   \textbf{0.090} \\
    EMPF-Net$_{23}$~\cite{wen2023encoder}     & 28.393 &  \textbf{11.590} &   -     &   0.300 \\
    RSDformer$_{23}$~\cite{song2023learning}    & 24.487 &  50.120 &   -     &   4.270 \\
    SFAN$_{24}$~\cite{shen2024spatial}        & 24.962 &  16.410 &   -     &   3.950 \\
    ICL-Net$_{24}$~\cite{dong2024icl}      & 24.013 & 321.660 &   -     &   9.810 \\
    PGSformer$_{24}$~\cite{dong2024prompt}    & \textbf{25.251} &  63.620 &   -     &   7.040 \\
    ASTA$_{24}$~\cite{cai2024additional}         & 23.613 & 155.229 &  19.57  &   3.322 \\
    DehazeXL$_{25}$~\cite{chen2025tokenize}     & 16.060 &  90.919 &  91.86  & 118.983 \\
    \bottomrule
\end{tabular}
\end{table}

\subsubsection{Model Efficiency Comparison and Analysis}

\begin{figure}[t]
	\centering 
	\includegraphics[width=12cm]{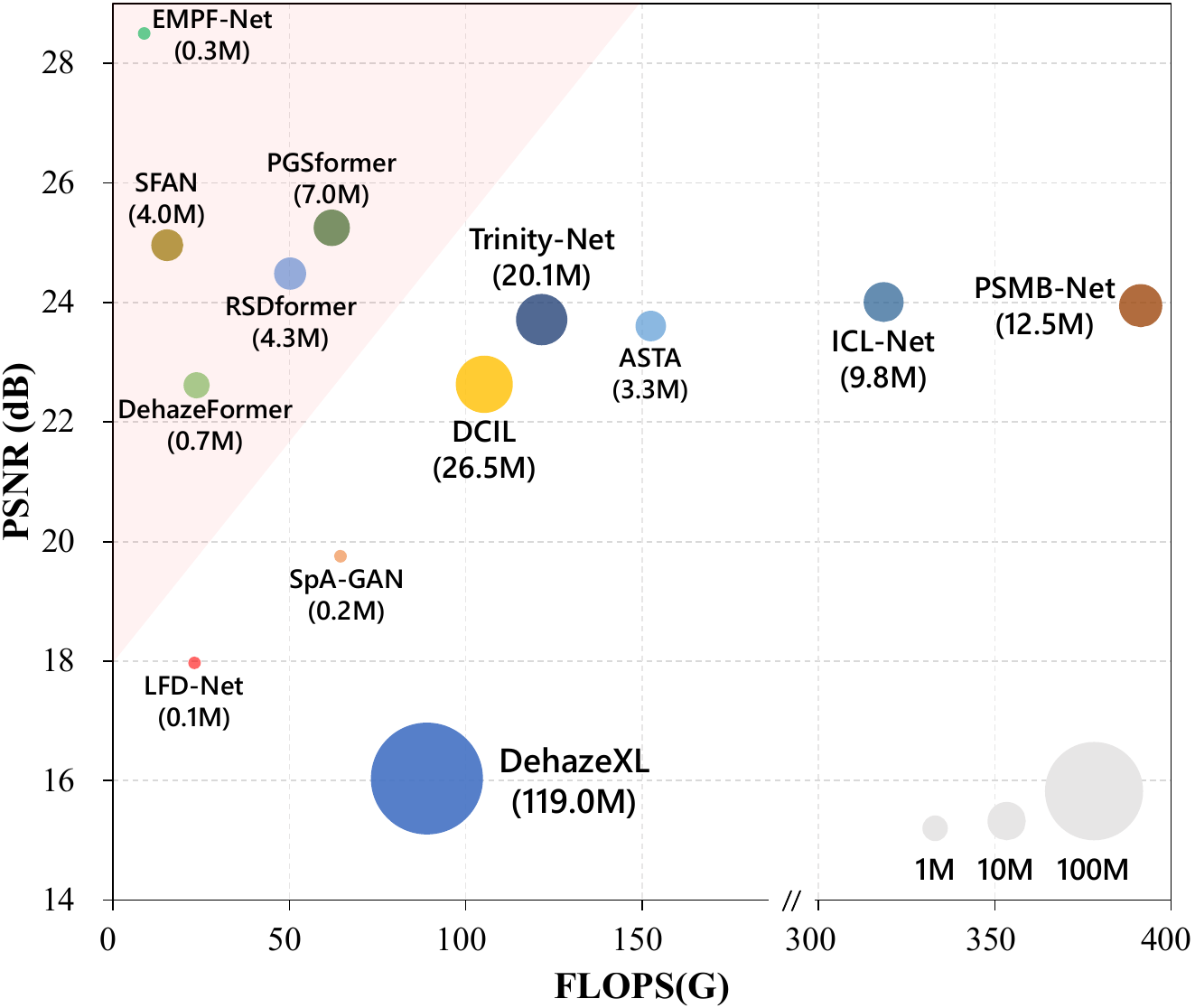}
	\caption{{Comparison of representative RSIs dehazing methods in terms of dehazing accuracy (PSNR) and computational complexity (FLOPs and Parameter).}}
	\label{Fig.FLOPs}
\end{figure}

\tablename~\ref{tab:model_efficiency} and \figurename~\ref{Fig.FLOPs} shows model efficiency to balance model performance and cost.
Latency and throughput were measured on an RTX 3090 GPU and a dual-socket Intel Xeon Gold 6148 CPU.
Lightweight LFD-Net achieves the fastest throughput at 269.07 FPS with only 0.09M parameters, but this extreme compactness corresponds to the weakest radiometric and geometric performance, confirming that aggressive parameter reduction limits the ability to model complex atmospheric scattering.
DehazeFormer provides a more balanced profile. 
With 24.82G FLOPs and 41.48 FPS, it maintains strong perceptual and structural stability under cross-domain evaluation, indicating that attention mechanisms improve robustness without high computational cost.
As listed in \tablename~\ref{tab:model_efficiency}, DCIL incurs a higher cost at 107.45G FLOPs and 26.51M parameters, yet delivers more stable radiometric behaviour and achieves a moderate speed of 52.35 FPS. 
This reflects the stabilizing effect of physical regularization.
Multi-branch PSMB-Net is the least efficient, requiring 392.75G FLOPs and reaching only 18.54 FPS, while offering no proportional gain on real-scene datasets. 
Mid-scale architectures adopt different strategies. 
ASTA emphasises representational depth and runs at 19.57 FPS. As reported in \tablename~\ref{tab:model_efficiency}, DehazeXL, despite having 118.98M parameters, maintains relatively high throughput at 91.86 FPS due to efficient parallel processing.
Overall, Transformer methods currently provide the most favorable balance, whereas lightweight CNNs sacrifice fidelity, and adversarial multi-branch networks impose heavy computational burdens. 

% ----------- 一致性评价指标 -----------
\begin{figure}[t]
	\centering 
	\includegraphics[width=13.5cm]{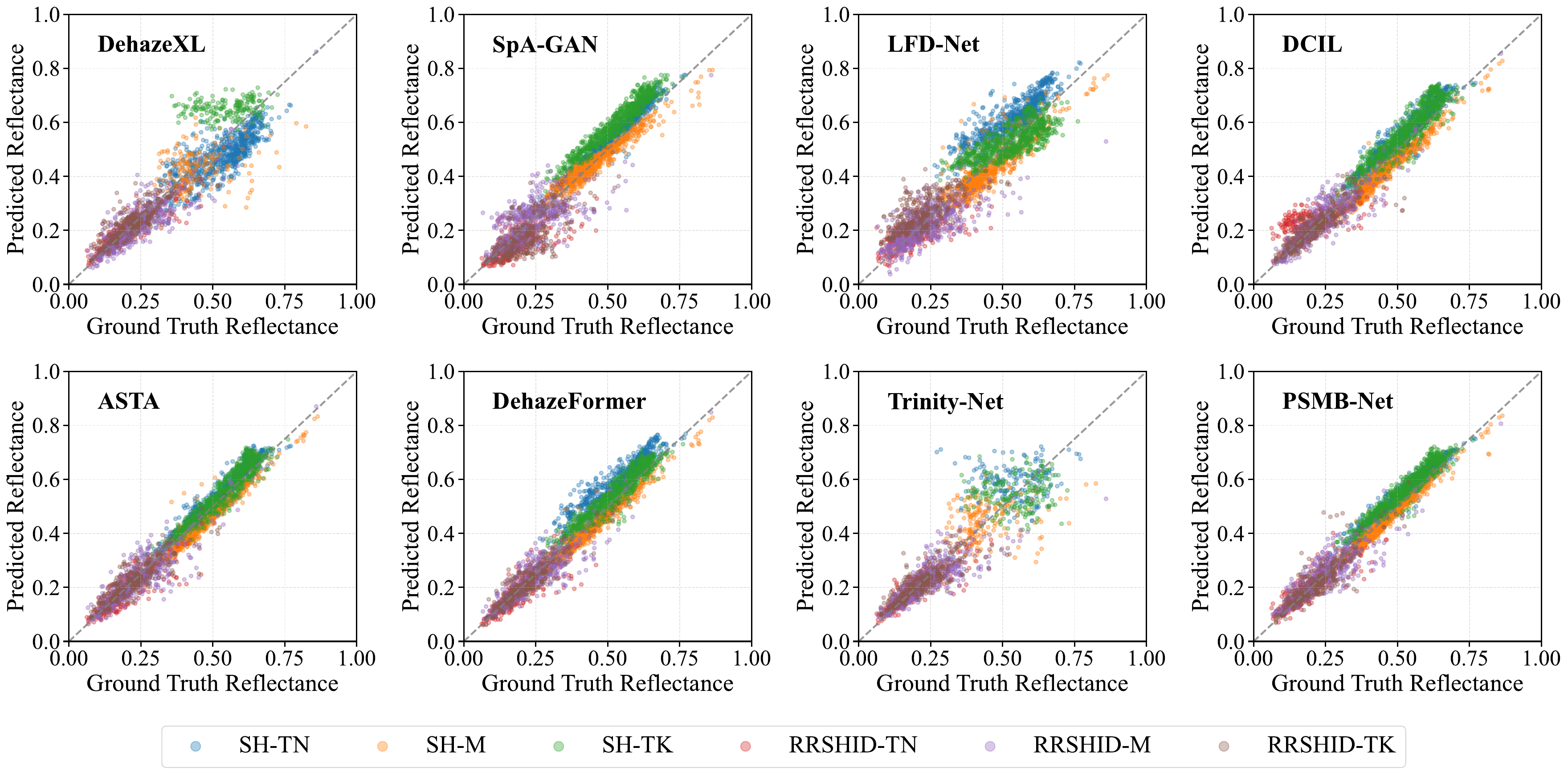}
	\caption{Radiometric consistency analysis of RSIs before and after dehazing.}
	\label{Fig.cons}
\end{figure}

{As shown in \figurename~\ref{Fig.FLOPs}, the methods highlighted in the upper left corner achieve a better trade-off between performance and efficiency. 
SFAN, PGSformer, and DehazeFormer focus on aggregating salient structures and local details, suppressing redundant and interfering features, thus maintaining good dehazing performance with lower computational overhead. 
EMPF-Net and RSDformer improve the effectiveness of feature learning by introducing a lightweight physical perception fusion module and global and local feature extraction modules, thereby achieving robust dehazing results with moderate parameter sizes.}

\begin{table}[t]
\centering
\scriptsize     % -------- 调小表格字号（推荐）
\caption{Physical consistency comparison in terms of SAM and ERGAS on SateHaze1k.}
\label{tab:sam_ergas_sh}
\setlength{\tabcolsep}{10pt}
\renewcommand{\arraystretch}{1.1}
\begin{tabular}{lcccccc}
\toprule
    \multirow{2}{*}{\textbf{Methods}} &
    \multicolumn{2}{c}{\textbf{SH-TN}} &
    \multicolumn{2}{c}{\textbf{SH-M}}   &
    \multicolumn{2}{c}{\textbf{SH-TK}} \\ \cmidrule(l){2-3} \cmidrule(l){4-5}  \cmidrule(l){6-7}
    & SAM$\downarrow$ & ERGAS$\downarrow$ &
      SAM$\downarrow$ & ERGAS$\downarrow$ &
      SAM$\downarrow$ & ERGAS$\downarrow$ \\
    \midrule
    SpA-GAN~\cite{pan2020cloud}      & 2.670 & 13.386 & 4.246 & 32.128 & 3.725 & 24.088 \\
    DCIL~\cite{zhang2022dense}         & {\color{blue}2.462} & 14.473 &
                    4.304 & 30.591 &
                    {\color{red}2.816} & 17.574 \\
    DehazeFormer~\cite{song2023vision} & 2.646 & 18.127 &
                    3.923 & 29.493 &
                    3.032 & {\color{blue}16.013} \\
    PSMB-Net~\cite{sun2023partial}     & 2.980 & {\color{blue}13.075} &
                    3.565 & {\color{blue}28.706} &
                    3.440 & 16.808 \\
    Trinity-Net~\cite{chi2023trinity}  & 2.835 & 22.765 &
                    {\color{blue}3.457} & 34.624 &
                    3.494 & 22.293 \\
    LFD-Net~\cite{jin2023lfd}      & 2.857 & 23.813 &
                    5.507 & 34.873 &
                    4.869 & 25.929 \\
    ASTA~\cite{cai2024additional}         & {\color{red}2.416} & {\color{red}12.188} &
                    {\color{red}2.284} & {\color{red}11.727} &
                    {\color{blue}3.014} & {\color{red}15.343} \\
    DehazeXL~\cite{chen2025tokenize}     & 6.988 & 26.444 &
                    3.751 & 37.214 &
                    3.345 & 36.951 \\
    \bottomrule
\end{tabular}
% }
\end{table}

\begin{figure}[t]
	\centering 
	\includegraphics[width=13.5cm]{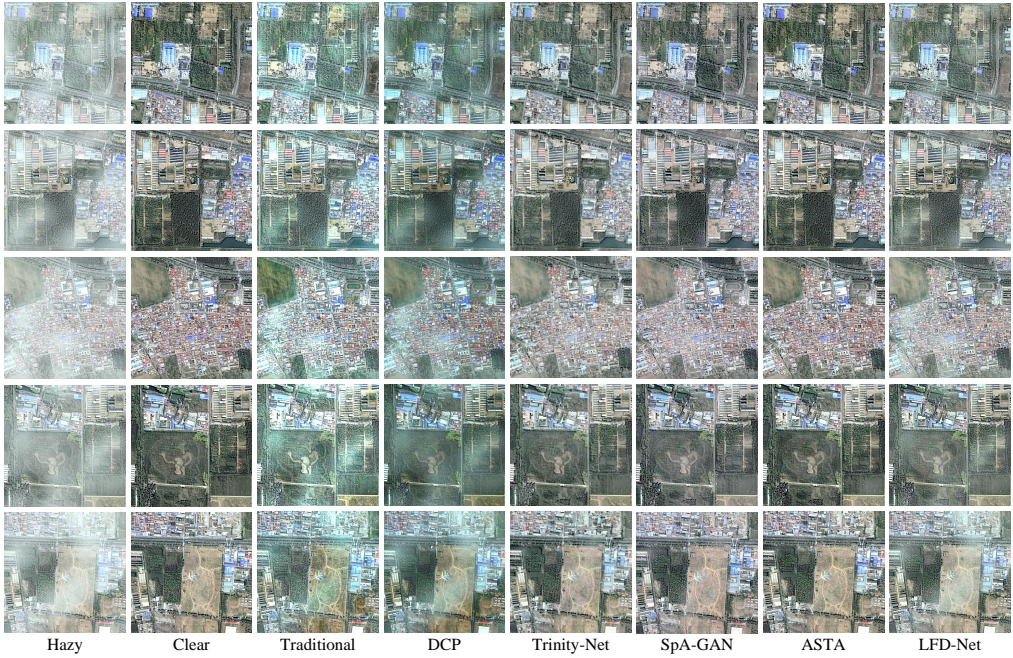}
	\caption{Visual comparison of dehazing results on the SateHaze1k datasets.}
	\label{Fig.SateHaze1k}
\end{figure}

\subsubsection{Physical Radiometric Consistency Evaluation}

\begin{figure}[t]
	\centering 
	\includegraphics[width=13.5cm]{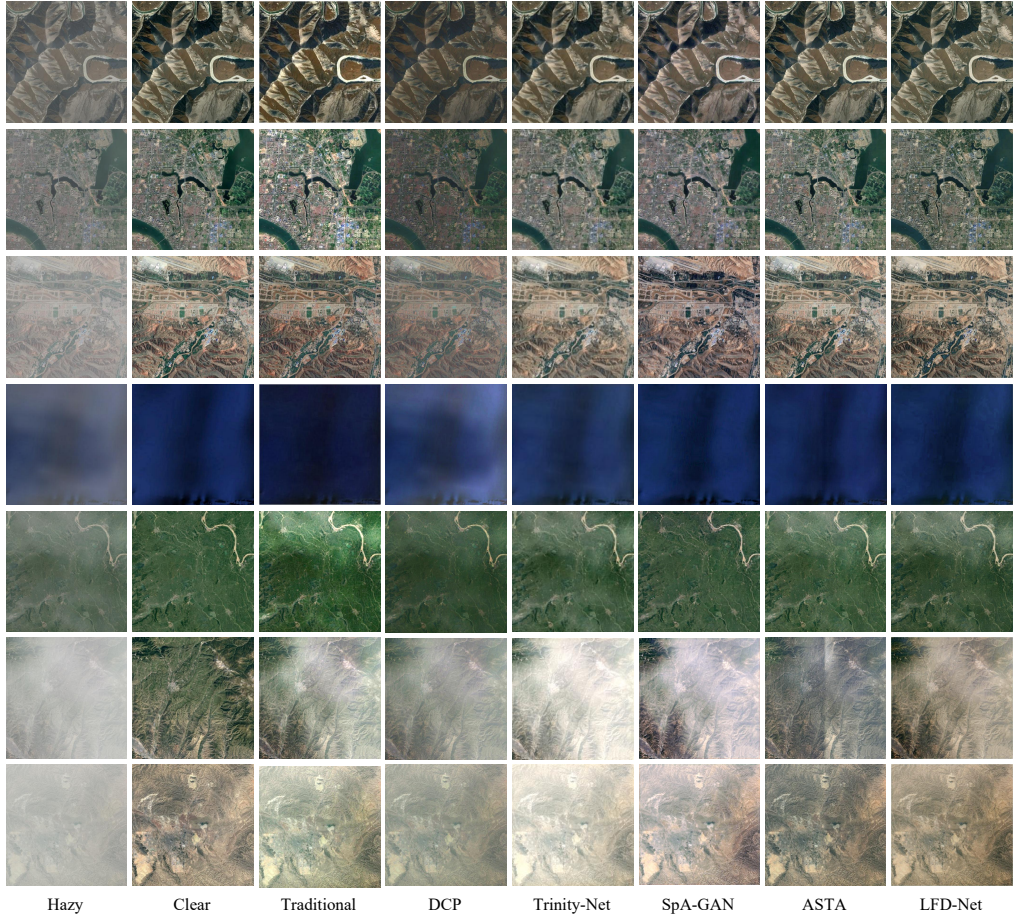}
	\caption{Visual comparison of dehazing results on the RICE datasets.}
	\label{Fig.RICE}
\end{figure}

Figure \ref{Fig.cons} and Table \ref{tab:sam_ergas_sh} demonstrate the radiometric reliability~\cite{syariz2019spectral} of each method across the SH-TN, SH-M and SH-TK domains. The scatter distributions in Figure \ref{Fig.cons} show that DehazeFormer, ASTA, DCIL and DehazeXL remain tightly aligned with the one-to-one reference line, preserving the proportionality between predicted and true surface reflectance. 
Their point clusters stay compact across all reflectance ranges, including low-reflectance zones near 0.2 and mid-reflectance regions near 0.5. 
SpA-GAN, Trinity-Net, LFD-Net and PSMB-Net exhibit wide spreads and a noticeable downward shift, reflecting systematic reflectance underestimation and inconsistent brightness recovery in both synthetic and real domains.The quantitative evidence in \tablename~\ref{tab:sam_ergas_sh} supports these visual observations. Under SH-TN, ASTA achieves the lowest SAM of 2.416 and ERGAS of 12.188, reducing SAM by 1.9\% compared to the next best DCIL and ERGAS by 6.8\% relative to PSMB-Net, demonstrating superior preservation of spectral direction and radiometric consistency in weak haze. 
Under SH-M, ASTA maintains a clear advantage with a SAM of 2.284 and an ERGAS of 11.727, achieving a 1.173 reduction in SAM compared to Trinity-Net. 
DehazeFormer and PSMB-Net form a strong second tier with ERGAS values of 29.493 and 28.706, respectively, but still fall short of ASTA in radiometric fidelity. Under the densest haze, ASTA attains the lowest ERGAS of 15.343, improving upon DehazeFormer by 0.670, while DCIL achieves the minimum SAM of 2.816, 0.198 lower than the second-best ASTA. 
LFD-Net, SpA-GAN, and DehazeXL, lacking explicit physical priors or global attention, exhibit larger SAM values, indicating reduced radiometric reliability and increased sensitivity to haze-induced reflectance distortions.

Across the three RRSHID domains shown in \figurename~\ref{Fig.cons}, DehazeFormer and ASTA maintain narrow and domain-invariant distributions, while the scatter patterns of PSMB-Net, Trinity-Net, SpA-GAN and LFD-Net expand significantly along both axes. 
This reflects weakened radiometric proportionality under real aerosol structures and heterogeneous atmospheric scattering. The combined results confirm that models incorporating global attention, adaptive structural modulation, or physics-guided constraints provide the most stable and physically coherent reflectance recovery, whereas purely convolutional and adversarial architectures suffer from domain-dependent deviations and reduced physical credibility.

\begin{figure}[t]
	\centering 
	\includegraphics[width=13.5cm]{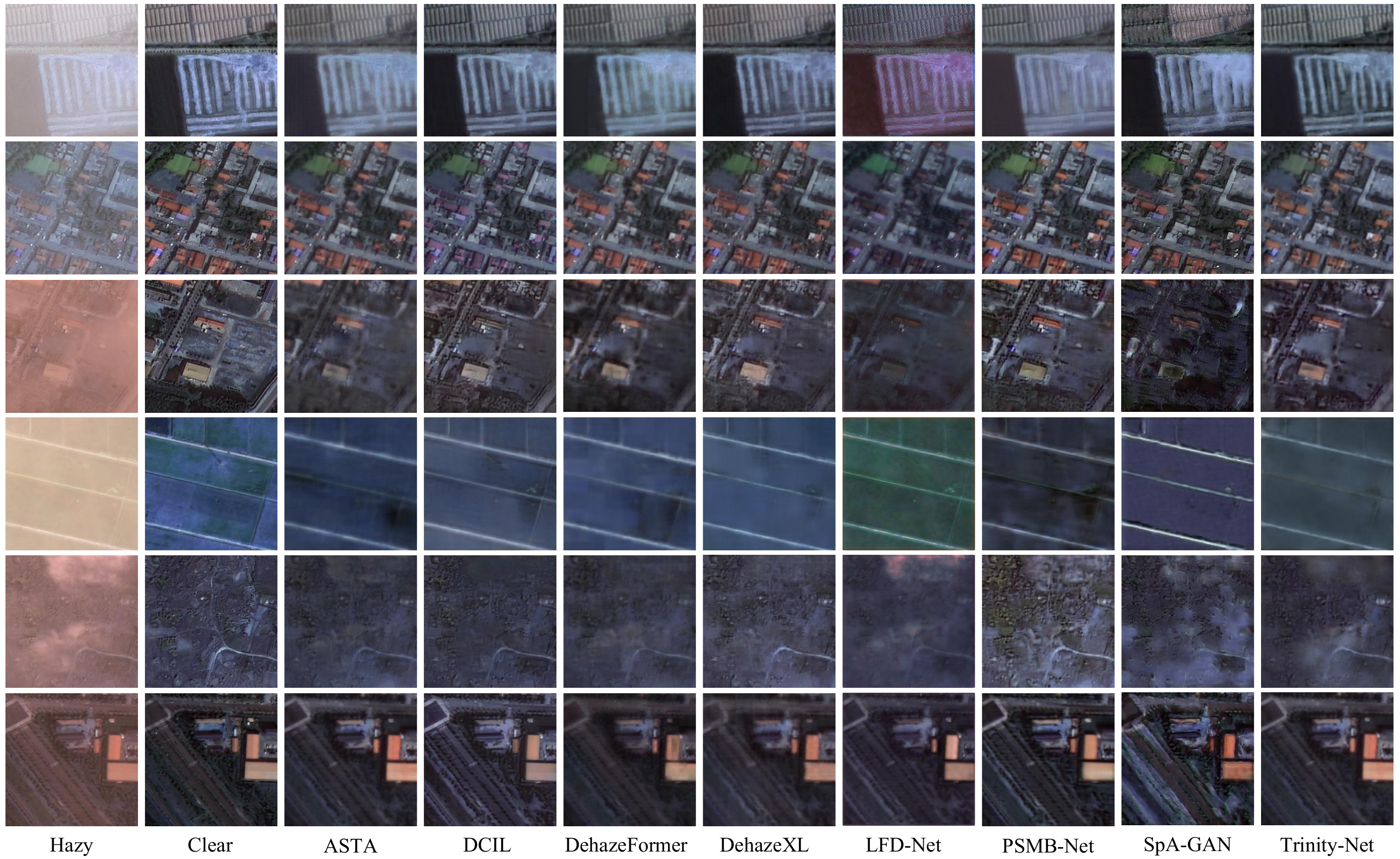}
	\caption{Visual comparison of dehazing results on the RRSHID datasets.}
	\label{Fig.RRSHID}
\end{figure}

\begin{figure}[t]
	\centering 
	\includegraphics[width=13.5cm]{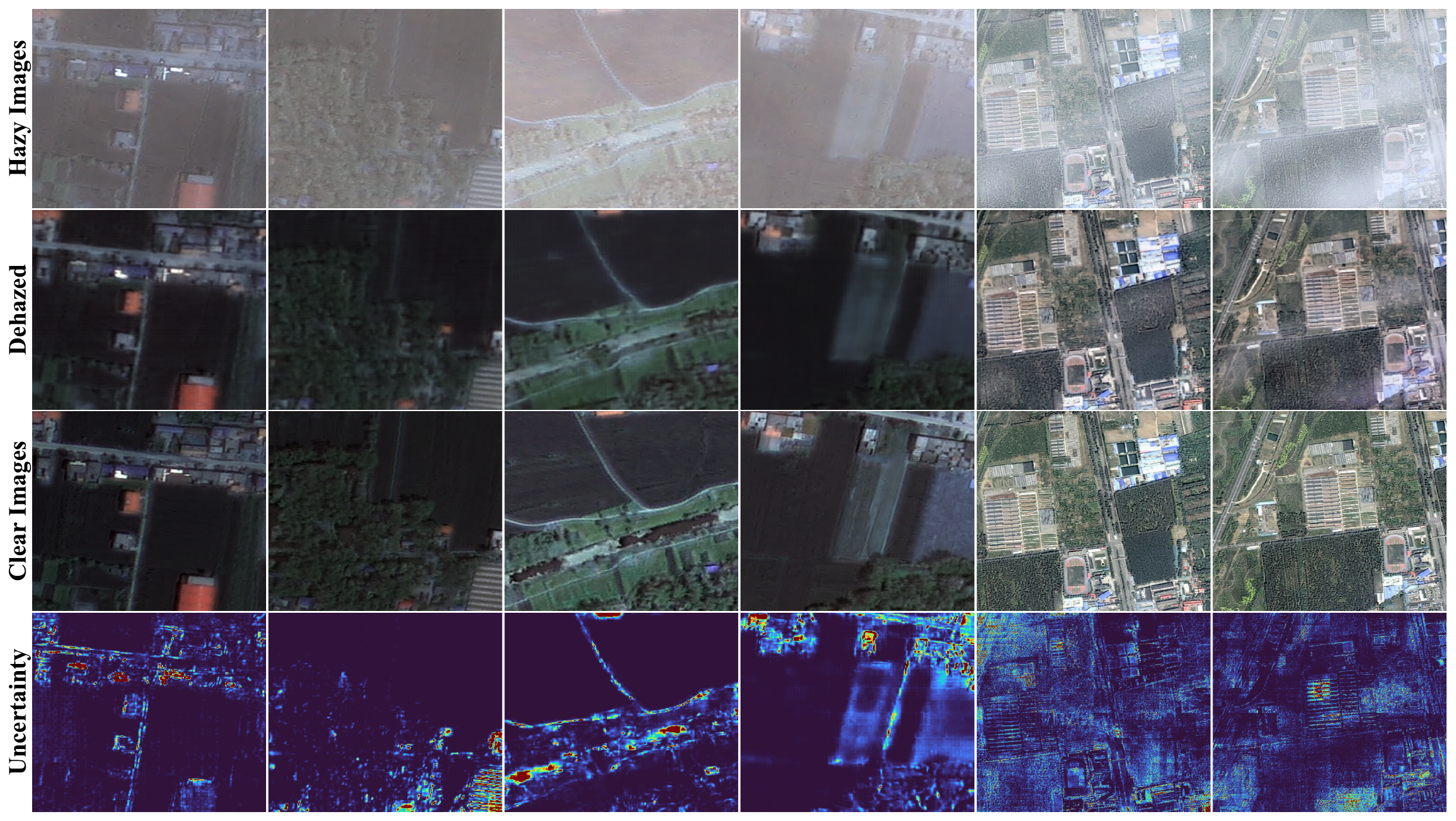}
	\caption{Uncertainty maps of DehazeXL on the RRSHID-TN, RRSHID-M and SateHaze1k-M datasets.}
	\label{Fig.MC}
\end{figure}

% ------------ 定性分析 -------------
\subsubsection{Visual Comparison}
As shown in \figurename~\ref{Fig.SateHaze1k} on the SateHaze1k datasets, where haze is synthetically controlled, we observe clearer differentiation across methods under diverse scene textures. 
In dense haze cases (\textit{e.g.}, Rows 1–3), traditional methods struggle to recover fine details, while learning-based models like Trinity-Net and ASTA demonstrate better structural recovery and contrast enhancement. 
These visual results of \figurename~\ref{Fig.SateHaze1k} emphasize the importance of balancing haze removal strength and detail preservation. 
While GAN-based models may achieve high subjective clarity, they risk introducing artificial textures.
Transformer-inspired Trinity-Net and ASTA benefit from progressive refinement, achieving both numerical and perceptual quality.

As shown in \figurename~\ref{Fig.RICE} and \figurename~\ref{Fig.RRSHID},
traditional image enhancement and prior-based methods show limited haze removal, with residual haze artifacts and color shifts, especially in blue or dark regions (\textit{e.g.}, water bodies in Row 4) on the RICE datasets. 
In contrast, deep learning methods like Trinity-Net and ASTA offer more natural tone dehazing but sometimes exhibit issues such as excessive local color saturation, subtle edge artifacts, and noise amplification.
Under thin haze conditions, SpA-GAN generates visually clean results with well-preserved edges and minimal color distortion.
ASTA and LFD-Net exhibit comparatively favorable performance under thick haze scenarios, demonstrating their robustness in handling dense haze conditions.

We further evaluate the reliability of haze removal by estimating pixel-wise predictive uncertainty using MC Dropout. 
As shown in \figurename~\ref{Fig.MC}, DehazeXL produces low uncertainty on homogeneous regions in SateHaze1k-M, indicating stable restoration when haze distribution and surface reflectance are regular. 
Higher uncertainty appears along edges, thin-haze transitions, and high-frequency textures, which correspond to regions where residual haze or color shifts are more likely.
On the real RRSHID datasets, uncertainty significantly increases, especially around vegetation–soil boundaries, industrial structures, and shadowed areas. This reflects the model’s reduced confidence under real atmospheric variability and sensor-induced radiometric deviations. Overall, the uncertainty maps reveal spatial patterns of potential over-dehazing and artifact risk, offering a diagnostic indicator of model trustworthiness under synthetic and real-world conditions.

\subsubsection{Real-world Dehazing Comparison and Analysis}

\begin{table*}[t]
\centering
\small
\caption{Quantitative evaluation on the authentic RRSHID datasets.
}
\label{tab:rrshid_real}
\resizebox{\textwidth}{!}{%
\setlength{\tabcolsep}{2.0pt}
\renewcommand{\arraystretch}{1}
\begin{tabular}{ccccccccc}
\toprule
    \textbf{Metrics} &
    \textbf{SpA-GAN~\cite{pan2020cloud}} &
    \textbf{DCIL~\cite{zhang2022dense}} &
    \textbf{DehazeFormer~\cite{song2023vision}} &
    \textbf{PSMB-Net~\cite{sun2023partial}} &
    \textbf{Trinity-Net~\cite{chi2023trinity}} &
    \textbf{LFD-Net~\cite{jin2023lfd}} &
    \textbf{ASTA~\cite{cai2024additional}} &
    \textbf{DehazeXL~\cite{chen2025tokenize}} \\
    \midrule
    \multirow{3}{*}{PSNR$\uparrow$} &
    20.640 & 20.846 & 22.640 & \textcolor{red}{23.190} & \textcolor{blue}{23.003} & 21.002 & 22.420 & 22.980 \\
    & 20.525 & \textcolor{blue}{23.254} & \textcolor{red}{23.360} & 23.120 & 22.767 & 20.838 & 20.843 & 23.023 \\
    & 20.947 & \textcolor{blue}{24.640} & 24.550 & \textcolor{red}{25.180} & 24.055 & 20.433 & 24.160 & 24.171 \\
    \midrule
    \multirow{3}{*}{SSIM$\uparrow$} &
    0.493 & 0.527 & 0.594 & \textcolor{red}{0.643} & \textcolor{blue}{0.631} & 0.517 & 0.591 & 0.592 \\
    & 0.506 & 0.651 & \textcolor{blue}{0.660} & 0.657 & \textcolor{red}{0.665} & 0.544 & 0.651 & 0.640 \\
    & 0.573 & 0.707 & 0.705 & \textcolor{red}{0.726} & \textcolor{blue}{0.725} & 0.527 & 0.699 & 0.681 \\
    \midrule
    \multirow{3}{*}{CIEDE$\downarrow$} &
    8.142 & 8.536 & \textcolor{blue}{6.374} & 6.466 & 6.616 & 8.542 & 6.635 & \textcolor{red}{6.286} \\
    & 8.767 & 6.278 & \textcolor{red}{6.096} & 6.551 & 7.152 & 10.071  & 6.339 & \textcolor{blue}{6.208} \\
    & 8.274 & \textcolor{red}{5.511} & \textcolor{blue}{5.530} & 6.173 & 6.153 & 10.267 & 5.785 & 5.859 \\
    \midrule
    \multirow{3}{*}{LPIPS$\downarrow$} &
    \textcolor{red}{0.337} & \textcolor{blue}{0.352} & 0.456 & 0.367 & 0.443 & 0.454 & 0.447 & 0.419 \\
    & \textcolor{blue}{0.374} & \textcolor{red}{0.359} & 0.430 & 0.384 & 0.432 & 0.515 & 0.416 & 0.416 \\
    & 0.388 & \textcolor{red}{0.361} & 0.434 & \textcolor{blue}{0.375} & 0.428 & 0.510 & 0.426 & 0.419 \\
    \midrule
    \multirow{3}{*}{FID$\downarrow$} &
    201.035 & \textcolor{red}{136.350} & 196.099 & \textcolor{blue}{163.432} & 197.717 & 218.376 & 191.525 & 190.446 \\
    & 161.389 & \textcolor{red}{130.122} & 164.628 & \textcolor{blue}{146.939} & 164.744 & 191.525 & 159.036 & 162.509 \\
    & 201.962 & \textcolor{red}{162.604} & \textcolor{blue}{171.623} & 171.623 & 198.392 & 227.452 & 214.422 & 208.458 \\
    \midrule
    \multirow{3}{*}{SAM$\downarrow$} &
    6.300 & 5.744 & \textcolor{red}{4.426} & 4.834 & 4.815 & 6.388 & 4.804 & \textcolor{blue}{4.552} \\
    & 3.732 & 3.188 & \textcolor{blue}{2.970} & 3.600 & 3.618 & 3.101 & 3.119 & \textcolor{red}{2.915} \\
    & 3.932 & \textcolor{red}{2.618} & 2.695 & 3.267 & 2.713 & 5.646 & \textcolor{blue}{2.624} & 2.673 \\
    \midrule
    \multirow{3}{*}{ERGAS$\downarrow$} &
    52.773 & 58.788 & 42.373 & \textcolor{red}{39.871} & \textcolor{blue}{40.377} & 51.053 & 43.386 & 40.684 \\
    & 42.733 & 31.654 & \textcolor{blue}{31.464} & \textcolor{red}{31.340} & 32.754 & 43.386 & 38.506 & 32.883 \\
    & 41.469 & \textcolor{red}{27.429} & \textcolor{blue}{27.907} & 28.612 & 29.544 & 47.459 & 29.345 & 30.163 \\
    \midrule
    \multirow{3}{*}{UIQI$\uparrow$} &
    0.493 & 0.527 & 0.572 & \textcolor{red}{0.644} & \textcolor{blue}{0.611} & 0.517 & 0.566 & 0.592 \\
    & 0.506 & \textcolor{blue}{0.650} & 0.637 & \textcolor{red}{0.661} & 0.647 & 0.544 & 0.521 & 0.629 \\
    & 0.573 & \textcolor{blue}{0.708} & 0.685 & \textcolor{red}{0.715} & 0.691 & 0.660 & 0.678 & 0.640 \\
    \midrule
    \multirow{3}{*}{QNR$\uparrow$} &
    0.335 & 0.368 & \textcolor{blue}{0.444} & 0.434 & 0.433 & 0.339 & 0.426 & \textcolor{red}{0.445} \\
    & 0.448 & 0.545 & \textcolor{red}{0.559} & 0.532 & 0.518 & 0.383 & 0.547 & \textcolor{blue}{0.554} \\
    & 0.455 & \textcolor{red}{0.593} & \textcolor{blue}{0.586} & 0.552 & 0.577 & 0.373 & 0.582 & 0.576 \\
    \midrule
    \multirow{3}{*}{NIQE$\downarrow$} &
    1.070 & \textcolor{blue}{1.033} & 1.171 & 1.118 & \textcolor{red}{1.016} & 1.130 & 1.302 & 1.346 \\
    & 1.145 & 1.007 & 1.050 & \textcolor{blue}{0.981} & \textcolor{red}{0.860} & 1.222 & 1.192 & 1.021 \\
    & \textcolor{blue}{0.881} & 1.070 & 1.135 & 0.990 & \textcolor{red}{0.877} & 1.599 & 1.250 & 1.537 \\
    \midrule
    \multirow{3}{*}{HIST$\uparrow$} &
    0.737 & 0.654 & 0.631 & 0.744 & \textcolor{blue}{0.779} & \textcolor{red}{0.781} & 0.618 & 0.721 \\
    & 0.607 & \textcolor{red}{0.658} & 0.658 & 0.631 & \textcolor{blue}{0.645} & 0.551 & 0.637 & 0.622 \\
    & 0.494 & \textcolor{blue}{0.588} & 0.562 & 0.587 & \textcolor{red}{0.593} & 0.415 & 0.520 & 0.529 \\
    \midrule
    % ------------------------ Rank------------------------
    {$R \downarrow$} 
    & 6.182 & \textcolor{blue}{3.212} & 3.667 & \textcolor{red}{2.848} & 3.727 & 7.545 & 4.515 & 4.303 \\
    % \midrule
    % ------------------------ Ranking gaps ------------------------
    {$\Delta R \downarrow$} 
    & 3.333 & \textcolor{blue}{0.364} & 0.818 & \textcolor{red}{0.000} & 0.879 & 4.697 & 1.667 & 1.455 \\
    % \midrule
    % ------------------------ statistical significance -----------------------
    {$p \downarrow$}
    & 0.357 & 0.061 & 0.930 & - & 0.464 & 0.086 & 0.548 & 0.986 \\
    \bottomrule
    \multicolumn{9}{l}{
    \small $\bullet$ Each row corresponds to results under thin haze, moderate haze, and thick haze, respectively.
    }
    \\
\end{tabular}
} % end resizebox
\end{table*}

\begin{figure}[t]
	\centering 
	\includegraphics[width=13cm]{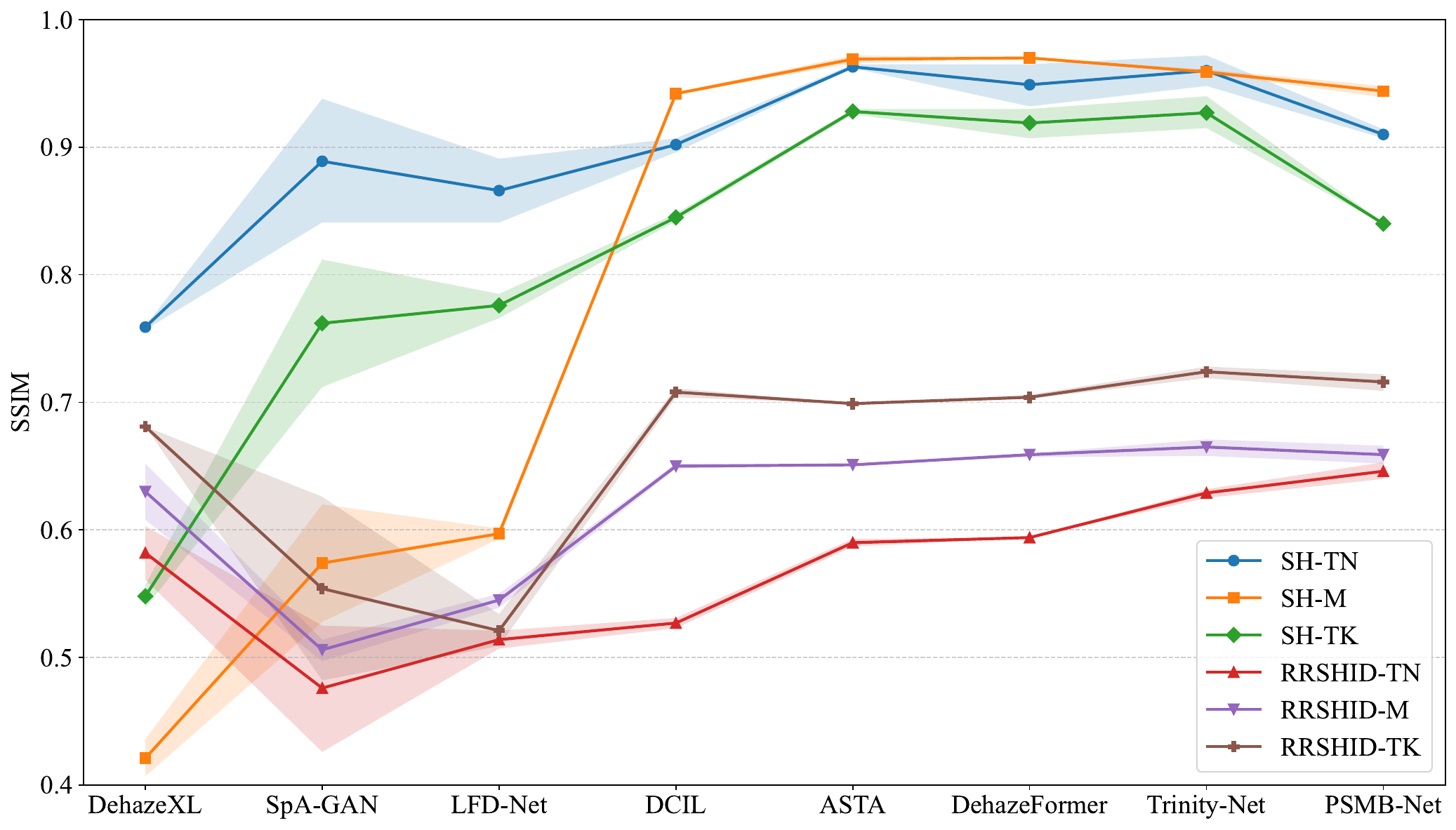}
	\caption{
    SSIM curves of different methods on SateHaze1k and RRSHID (thin to thick haze), reflecting structural fidelity under varying haze conditions.}
	\label{Fig.SSIM}
\end{figure}

\tablename~\ref{tab:rrshid_real} provides quantitative evaluation results of the real-world haze RRSHID datasets.
{The haze-free images of RRSHID are obtained by performing multi-temporal acquisitions with temporal-spatial alignment for the same geographical area under clearer atmospheric conditions~\cite{zhu2025real}.}
The authentic haze causes a clear and systematic degradation for all methods. 
PSNR on RRSHID falls within 20.433 to 25.180 dB, well below the synthetic SateHaze1k results, while CIEDE increases by roughly 20 to 40 percent, showing stronger color distortion. 
As reported in \tablename~\ref{tab:rrshid_real}, ERGAS frequently exceeds 40, and UIQI drops to 0.493 to 0.715, about 15\% to 25\% lower than synthetic-domain performance, confirming the radiometric and structural complexity of real atmospheric scattering.
The model differences become more pronounced under real haze. 
Transformer-based approaches, especially DehazeFormer and ASTA, maintain comparatively stable perceptual and radiometric fidelity, achieving CIEDE near 6.300 and LPIPS around 0.430, and showing narrow uncertainty regions in their SSIM curves in \figurename~\ref{Fig.SSIM}. 
CNN and GAN models exhibit amplified degradation, with PSMB-Net reaching CIEDE around 6.466 and SAM near 4.834, and SpA-GAN producing SAM between 3.732 and 6.300, revealing unstable color reproduction and weaker geometric consistency. 
Their SSIM curves show larger variance, highlighting stronger sensitivity to haze non-uniformity and domain shift.

As shown in \tablename~\ref{tab:rrshid_real}, RRSHID presents irregular aerosol density, illumination imbalance, and haze–shadow interactions, causing noticeable residual haze, texture flattening, and color drift in CNN and GAN outputs. 
Trinity-Net and PSMB-Net, with stronger global context modeling, retain more stable SSIM and reduced variance under thick haze in \figurename~\ref{Fig.SSIM}. 
Finally, the combined quantitative indicators and shaded confidence bands show that synthetic performance cannot reliably predict real-scene behavior and that models integrating physical priors and long-range dependency modeling offer superior resilience under authentic atmospheric conditions.

\subsubsection{Cross-domain Dehazing Generalization Analysis}

\begin{table*}[t]
\centering
\caption{Cross-domain evaluation: SateHaze1k $\rightarrow $ RICE.}
\label{tab:cross_satehaze_rice}
\resizebox{\textwidth}{!}{%
\begin{tabular}{lccccccccccc}
\toprule
    Models & PSNR$\uparrow$ & SSIM$\uparrow$ & CIEDE$\downarrow$ & LPIPS$\downarrow$ &
    FID$\downarrow$ & SAM$\downarrow$ & ERGAS$\downarrow$ &
    UIQI$\uparrow$ & QNR$\uparrow$ & NIQE$\downarrow$ & HIST$\uparrow$ \\
    \midrule
    SpA-GAN~\cite{pan2020cloud}     
    & \textcolor{blue}{20.956} & \textcolor{blue}{0.762} & \textcolor{red}{9.141} 
    & 0.336 & 88.895 & 5.526 & \textcolor{blue}{42.712} 
    & \textcolor{red}{0.762} & 0.453 & 2.342 & 0.311 \\
    DCIL~\cite{zhang2022dense}        
    & 18.762 & 0.725 & 11.124 
    & 0.351 & 87.667 & 6.287 & 63.187 
    & 0.725 & 0.418 & 1.749 & 0.297 \\
    DehazeFormer~\cite{song2023vision}
    & 20.842 & 0.736 & 9.336 
    & \textcolor{blue}{0.325} & \textcolor{blue}{83.375} & \textcolor{blue}{5.166} & 43.439 
    & 0.736 & \textcolor{blue}{0.467} & \textcolor{blue}{1.056} & \textcolor{blue}{0.314} \\
    PSMB-Net~\cite{sun2023partial}    
    & 13.161 & 0.569 & 23.019 
    & 0.569 & 132.442 & 10.735 & 121.745 
    & 0.569 & 0.229 & 1.172 & 0.235 \\
    Trinity-Net~\cite{chi2023trinity} 
    & 12.884 & 0.410 & 26.921 
    & 0.739 & 83.749 & 7.414 & 44.230 
    & 0.693 & 0.413 & \textcolor{red}{0.977} & 0.170 \\
    LFD-Net~\cite{jin2023lfd}     
    & 13.695 & 0.618 & 27.638 
    & 0.441 & 117.951 & 7.848 & 183.025 
    & 0.618 & 0.322 & 1.529 & 0.257 \\
    ASTA~\cite{cai2024additional}        
    & \textcolor{red}{21.318} & \textcolor{red}{0.771} & \textcolor{blue}{9.220} 
    & \textcolor{red}{0.324} & \textcolor{red}{79.937} & \textcolor{red}{5.032} & \textcolor{red}{40.481} 
    & \textcolor{blue}{0.740} & \textcolor{red}{0.477} & 2.374 & \textcolor{red}{0.327} \\
    DehazeXL~\cite{chen2025tokenize}    
    & 14.855 & 0.384 & 17.387 
    & 0.725 & 126.511 & 8.917 & 46.055 
    & 0.620 & 0.394 & 1.062 & 0.293 \\
    \bottomrule
    \end{tabular}
}
\end{table*}

\begin{table*}[t]
\centering
\caption{Cross-domain evaluation: SateHaze1k  $\to$ RRSHID.}
\label{tab:cross_m_rrshid}
\resizebox{\textwidth}{!}{%
\begin{tabular}{lccccccccccc}
\toprule
    Models & PSNR$\uparrow$ & SSIM$\uparrow$ & CIEDE$\downarrow$ & LPIPS$\downarrow$ &
    FID$\downarrow$ & SAM$\downarrow$ & ERGAS$\downarrow$ &
    UIQI$\uparrow$ & QNR$\uparrow$ & NIQE$\downarrow$ & HIST$\uparrow$ \\
    \midrule
    SpA-GAN~\cite{pan2020cloud}  & 11.971 & 0.337 & 24.913 & 0.526 & 201.851 & 10.506 & 153.132 & 0.337 & 0.101 & 1.210 & 0.144 \\
    DCIL~\cite{zhang2022dense}   & 10.847 & 0.344 & 27.813 & 0.564 & 210.276 & \textcolor{blue}{9.816} & 174.042 & 0.343 & 0.096 & 1.174 & 0.094 \\
    DehazeFormer~\cite{song2023vision}  & 12.490 & 0.352 & 23.755 & 0.537 & 207.182 & 10.259 & 144.739  & 0.337 & 0.113 & 1.957 & 0.142 \\
    PSMB-Net~\cite{sun2023partial}    & 11.690 & 0.324 & 26.372 & \textcolor{blue}{0.460} & \textcolor{red}{151.500} & \textcolor{red}{8.545} & 159.266 & 0.324 & 0.118 & \textcolor{blue}{1.032} & \textcolor{blue}{0.227} \\
    Trinity-Net~\cite{chi2023trinity} & \textcolor{red}{15.620} & \textcolor{red}{0.446} & \textcolor{red}{17.703} & \textcolor{red}{0.448} & \textcolor{blue}{169.603} & 10.564 & \textcolor{red}{101.522} & \textcolor{red}{0.427} & \textcolor{red}{0.159} & 1.148 & \textcolor{red}{0.319} \\
    LFD-Net~\cite{jin2023lfd}   & 10.761 & 0.279 & 33.477 & 0.811 & 274.743 & 29.765 & 172.186 & 0.279 & 0.014 & \textcolor{red}{0.961} & 0.197 \\
    ASTA~\cite{cai2024additional}   & \textcolor{blue}{13.740} & 0.361 & \textcolor{blue}{21.105} & 0.524 & 195.031 & 10.742 & \textcolor{blue}{124.775} & 0.348 & \textcolor{blue}{0.128} & 3.014 & 0.189 \\
    DehazeXL~\cite{chen2025tokenize}   & 13.274 & \textcolor{blue}{0.381} & 22.970 & 0.518 & 179.213 & 11.961 & 133.398 & \textcolor{blue}{0.381} & 0.110 & 1.818 & 0.198 \\
    \bottomrule
\end{tabular}
}
\end{table*}

\begin{table*}[t]
\centering
\caption{{Cross-domain evaluation: SateHaze1k $\rightarrow $ UAV-HAZE.}}
{
\label{tab:cross_satehaze_uav}
\resizebox{\textwidth}{!}{%
\begin{tabular}{lccccccccccc}
\toprule
    Models & PSNR$\uparrow$ & SSIM$\uparrow$ & CIEDE$\downarrow$ & LPIPS$\downarrow$ &
    FID$\downarrow$ & SAM$\downarrow$ & ERGAS$\downarrow$ &
    UIQI$\uparrow$ & QNR$\uparrow$ & NIQE$\downarrow$ & HIST$\uparrow$ \\
    \midrule
    SpA-GAN~\cite{pan2020cloud}       
    & 17.088 & 0.806 & 15.290 
    & 0.286 & 62.415 & 8.687 & 28.157 
    & 0.806 & 0.329 & 1.128 & 0.760 \\
    DCIL~\cite{zhang2022dense}          
    & 19.244 & 0.804 & 9.119 
    & 0.204 & 22.702 & 3.227 & 22.186 
    & 0.804 & 0.589 & \textcolor{red}{0.765} & 0.736 \\
    DehazeFormer~\cite{song2023vision}
    & \textcolor{blue}{19.661} & 0.873 & \textcolor{blue}{8.322} 
    & 0.122 & \textcolor{blue}{21.810} & \textcolor{blue}{3.099} & \textcolor{red}{21.407} 
    & 0.873 & \textcolor{blue}{0.601} & 1.187 & 0.754 \\
    PSMB-Net~\cite{sun2023partial}     
    & 18.566 & 0.831 & 10.745 
    & 0.176 & 22.051 & 4.401 & 23.809 
    & 0.831 & 0.516 & \textcolor{blue}{0.917} & 0.748 \\
    Trinity-Net~\cite{chi2023trinity} 
    & 18.767 & 0.693 & 9.117 
    & 0.487 & 25.253 & 3.872 & 23.059 
    & 0.693 & 0.549 & 0.921 & \textcolor{blue}{0.762} \\
    LFD-Net~\cite{jin2023lfd}       
    & 16.032 & 0.765 & 14.063 
    & 0.261 & 51.687 & 7.119 & 31.573 
    & 0.765 & 0.365 & 1.018 & 0.709 \\
    ASTA~\cite{cai2024additional}         
    & 19.480 & \textcolor{red}{0.889} & \textcolor{red}{8.226} 
    & \textcolor{blue}{0.117} & 23.795 & \textcolor{red}{2.998} & \textcolor{blue}{21.629} 
    & \textcolor{red}{0.889} & \textcolor{red}{0.603} & 1.115 & 0.777 \\
    DehazeXL~\cite{chen2025tokenize}    
    & \textcolor{red}{19.693} & \textcolor{blue}{0.887} & 8.328 
    & \textcolor{red}{0.116} & \textcolor{red}{15.458} & 3.546 & 21.773 
    & \textcolor{blue}{0.887} & 0.574 & 1.056 & \textcolor{red}{0.792} \\
    \bottomrule
    \end{tabular}
}
}
\end{table*}

\begin{figure}[t]
	\centering 
	\includegraphics[width=13cm]{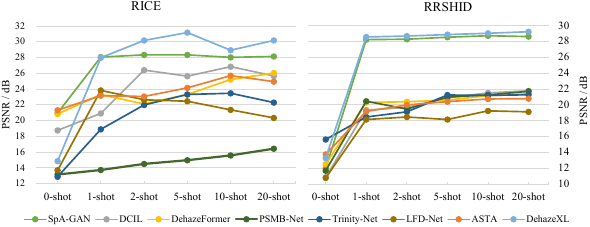}
	\caption{
    Cross-domain few-shot analysis of synthetic-real remote sensing dehazing datasets}
	\label{Fig.few-shot}
\end{figure}

To assess robustness under realistic distribution shifts, all models are trained on simulation SateHaze1k and evaluated on the real RICE and RRSHID datasets, as reported in \tablename~\ref{tab:cross_satehaze_rice},  \tablename~\ref{tab:cross_m_rrshid} {and \tablename~\ref{tab:cross_satehaze_uav}.} 
We statistically compared the haze properties of RICE and RRSHID and found that RICE corresponds to thin-haze conditions, while RRSHID aligns with moderate haze in terms of contrast, dark-channel strength, and color attenuation. 
Therefore, SateHaze1k-TN was used for SateHaze1k $\rightarrow$ RICE and SateHaze1k-M for SateHaze1k $\rightarrow$ RRSHID to match haze levels and ensure fair cross-dataset generalization evaluation.

Spa-GAN delivers a higher UIQI and a relatively smaller CIEDE; yet, its perceptual metrics vary considerably across domains in \tablename~\ref{tab:cross_satehaze_rice}. 
Models integrating physical constraints such as ASTA show more consistent PSNR, SAM, and LPIPS values, illustrating the stabilizing influence of physics-based regularization.
As shown in \tablename~\ref{tab:cross_m_rrshid}, SpA-GAN, DehazeFormer, and LFD-Net yield PSNR values in the range of 10.76 to 12.490 dB, yet present severe color deviation and poor perceptual quality. 
Their CIEDE often exceeds 23.755, LPIPS is typically above 0.526, and FID frequently surpasses 201.851. 
These results indicate that perceptual enhancement trained on synthetic haze does not transfer effectively to real conditions.
Transformer-based approaches provide comparatively stronger stability. 
DehazeFormer achieves the second lowest LPIPS on RICE and maintains competitive SAM on both datasets. 
However, its color accuracy remains limited, with CIEDE values of 9.336 on RICE and 23.755 on RRSHID. 
The two real datasets reveal different weaknesses. 
RICE primarily challenges radiometric fidelity and fine structure recovery. 
RRSHID, which contains dense and irregular haze, exposes sensitivity to variations in scattering intensity and scene complexity. 
{As shown in \tablename~\ref{tab:cross_satehaze_uav}, the model was trained on the SateHaze1k and quantitatively compared on the UAV-HAZE~\cite{wang2024depth}. 
The UAV-HAZE dataset contains approximately 35,000 synthetic haze images taken from drone aerial perspectives and approximately 400 real-world haze images.
As can be seen from \tablename~\ref{tab:cross_satehaze_uav}, DehazeXL achieved the best PSNR, LPIPS and FID scores in cross-domain scenarios, reaching 19.693 dB, 0.116 and 15.458, respectively. 
ASTA ranked first in the SSIM, CIEDE, SAM, UIQI and QNR metrics, reaching 0.889, 8.226, 2.998, 0.889 and 0.603 respectively. In contrast, although DCIL achieved the highest 0.765 in NIQE, it still lagged behind the above models in full-reference metrics.}

\begin{figure}[t]
	\centering 
	\includegraphics[width=13.5cm]{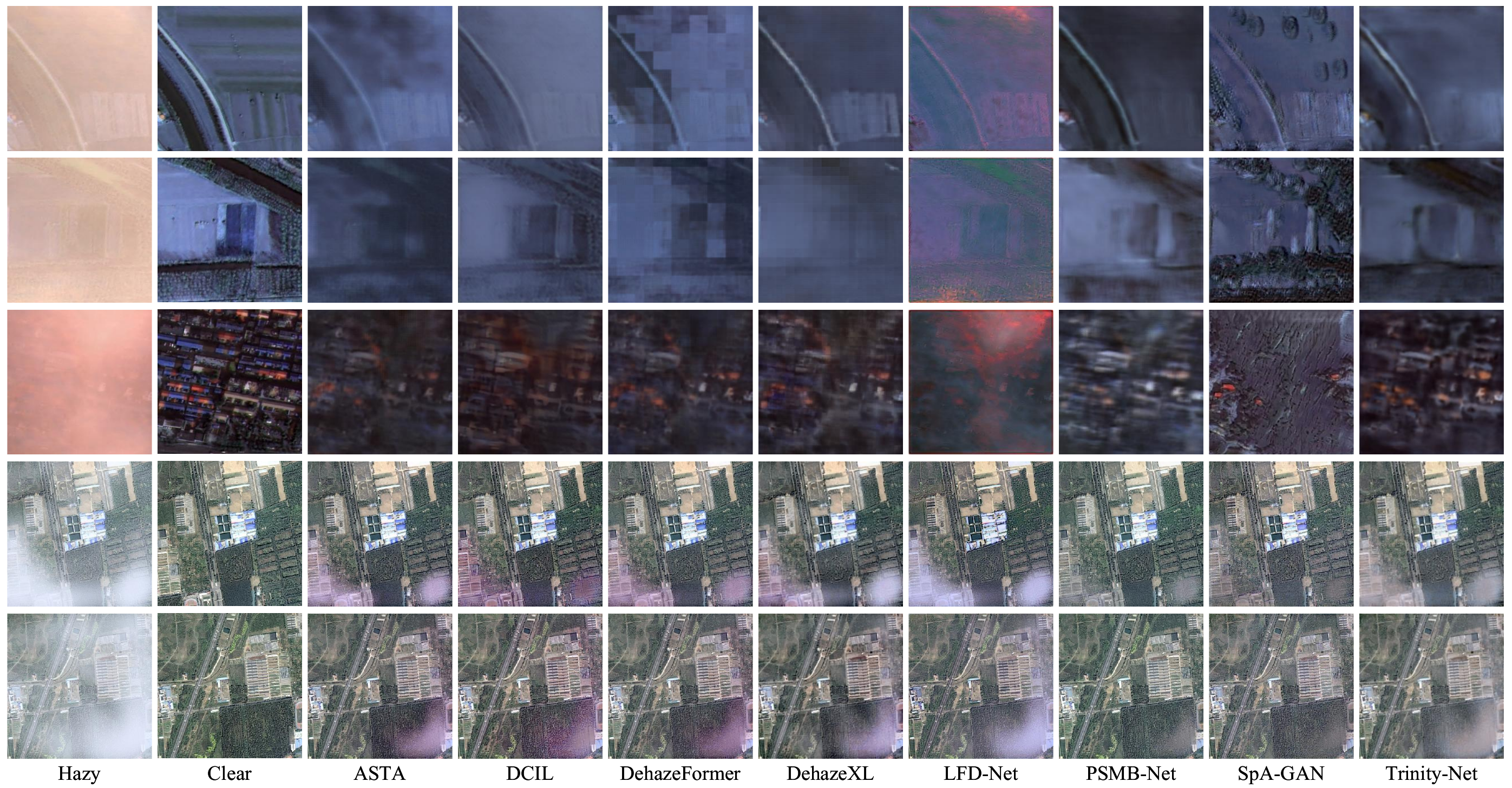}
	\caption{Some failure cases on RRSHID (1-3 rows) and SateHaze1k (4-5 rows) datasets.}
	\label{Fig.Failure}
\end{figure}

Furthermore, as shown in \figurename~\ref{Fig.few-shot}, we further explore the few-shot learning performance of remote sensing image dehazing.
Transformer-based methods achieve the strongest adaptation, rapidly rising from low 0-shot scores to around 28 dB with only 1–2 target samples, while maintaining the most stable growth across all shot settings. 
CNN models exhibit weaker generalization, starting from significantly lower 0-shot performance and improving slowly even with additional supervision. The gap between Transformer models and earlier architectures persists across both RICE and RRSHID, indicating that global-context modeling provides superior resilience to domain shift. 
Performance gains taper after 5–10 shots for all models, reflecting limited benefit from additional examples once core haze statistics are learned. 
Overall, the combined results show that existing dehazing models are unable to maintain radiometric accuracy, geometric structure and perceptual realism when facing real distribution shifts.

\begin{table*}[t]
\centering
% \small
\scriptsize  
\caption{HSI dehazing results on the HyperDehazing datasets.}
\label{tab:hyper_hsi}
\resizebox{\textwidth}{!}{%
\begin{tabular}{llcccc}
\toprule
    Methods & Publication & PSNR $\uparrow$ & SSIM $\uparrow$ & UIQI $\uparrow$ & SAM $\downarrow$ \\
    \midrule
    DCP~\cite{he2010single}  & CVPR, 2009  & 26.292 & 0.781 & 0.691 & 0.182 \\
    AOD-Net~\cite{li2017aod}  & ICCV, 2017  & 18.010 & 0.861 & 0.733 & --    \\
    FFANet~\cite{qin2020ffa}  & AAAI, 2020  & 30.891 & 0.898 & 0.944 & 0.090 \\
    NHRN~\cite{wei2021non} & TIP, 2021   & 26.140 & 0.823 & 0.886 & --    \\
    Defog~\cite{kang2021fog}  & TGRS, 2022  & 24.323 & 0.642 & 0.592 & 0.421 \\
    SGNet~\cite{ma2022spectral}  & ISPRS P\&RS, 2022 & 34.720 & 0.947 & 0.965 & 0.077 \\
    Restormer~\cite{zamir2022restormer} & CVPR, 2022  & \textbf{37.908} & 0.970 & 0.976 & 0.042 \\
    DehazeFormer~\cite{song2023vision}       & TIP, 2023   & 35.282 & 0.969 & 0.973 & 0.045 \\
    AIDTransformer~\cite{kulkarni2023aerial}     & WACV, 2023  & 35.472 & 0.971 & 0.974 & 0.036 \\
    AACNet~\cite{xu2023aacnet}             & TGRS, 2023  & 35.122 & 0.962 & 0.974 & 0.043 \\
    PSMB-Net~\cite{sun2023partial}           & TGRS, 2023  & 35.661 & 0.969 & 0.980 & 0.049 \\
    HDMba~\cite{fu2024hdmba}              & arXiv, 2024 & 33.797 & 0.953 & 0.965 & 0.038 \\
    DEA-Net~\cite{chen2024dea}            & TIP, 2024   & 24.940 & 0.789 & 0.851 & --    \\
    HyperDehazeNet~\cite{fu2024hyperdehazing}     & ISPRS P\&RS, 2024 & 37.211 & \textbf{0.976} & 0.982 & \textbf{0.031} \\
    UAVD-Net~\cite{li2025dehazing} & RS, 2025 &37.830	&0.967 &- &- \\
    PSGNet~\cite{liu2026progressive}              & ESWA, 2026  &31.760	&0.961	&\textbf{0.985}	& - \\
    \bottomrule
\end{tabular}
}
\end{table*}

\subsubsection{Failure Cases} 
As shown in \figurename~\ref{Fig.Failure}, typical failure cases are observed in both real and synthetic scenarios. CNN-based methods suffer from residual haze, block artifacts, and color shifts due to limited long-range modeling and weak radiometric constraints. GAN-based methods tend to over-enhance contrast and edges, leading to ringing artifacts, spurious textures, and hallucinated or collapsed structures, especially under dense haze. 
Transformer-based methods alleviate residual haze via global attention but still exhibit over-smoothed textures and color distortion in thick haze and cloud–haze transition regions. 
These results indicate that global context modeling must be jointly constrained by local detail preservation and radiometric priors to achieve robust restoration under challenging conditions.

\begin{figure}[t]
    \centering
    \includegraphics[width=13cm]{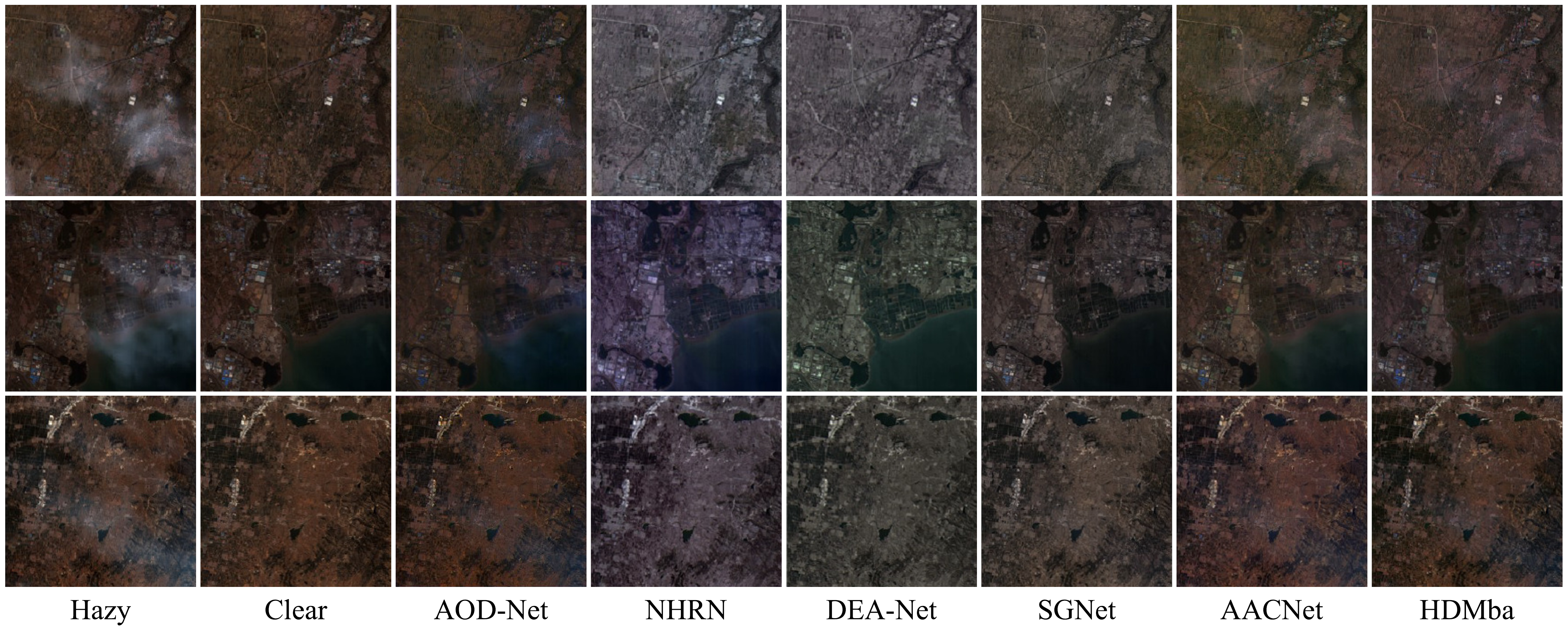}
    \caption{Visual comparison of dehazing results on the HyperDehazing datasets.}
    \label{fig:HSI}
\end{figure}

\subsubsection{Hyperspectral Image Dehazing Comparison}
The HyperDehazing dataset~\cite{fu2024hyperdehazing} represents the first large-scale benchmark dedicated to hyperspectral image (HSIs) dehazing, comprising 2,000 synthetic haze–clear pairs across 305 spectral channels and 70 additional real hazy HSIs. 
As shown in \tablename~\ref{tab:hyper_hsi}, early-stage methods and RGB-oriented approaches such as DCP, AOD-Net and FFANet yield PSNR values ranging from 18.010 to 30.891. 
The stronger DehazeFormer model reaches only 35.282dB PSNR and 0.973 UIQI, underscoring the difficulty of transferring RGB-based designs to high-dimensional hyperspectral data.
HSI-tailored methods achieve substantially higher restoration accuracy by explicitly modeling spectral correlation and wavelength-dependent degradation. 
In \tablename~\ref{tab:hyper_hsi}, Restormer, DehazeFormer and AIDTransformer deliver PSNR above 35.282 with UIQI exceeding 0.965, while HyperDehazeNet attains SSIM 0.976 and SAM 0.031, demonstrating strong spectral fidelity.

The visual comparisons in \figurename~\ref{fig:HSI} further validate these trends. 
Non-specialized methods introduce color imbalance, spectral inconsistency, and texture over-smoothing, particularly in heterogeneous urban materials and water bodies. 
HSI-tailored approaches preserve sharper boundaries, richer spatial–spectral structures, and more stable reflectance across wavelengths, with AACNet and HDMba producing the most physically plausible reconstructions.
\figurename~\ref{fig:HSI} shows that hyperspectral-specific modules such as spectral fusion, channel attention, and high-dimensional feature alignment are essential for stable HSI dehazing.

\subsubsection{Task-driven Evaluation}

\textbf{Detection-friendly Dehazing Evaluation.}
To assess whether dehazing benefits downstream perception, we adopt Faster R-CNN~\cite{ren2016faster} {training on clear images} and measure performance changes before and after applying different restoration models.
HazyDet~\cite{feng2024hazydet} is a large-scale UAV-oriented hazy-weather object detection dataset containing 383K annotated instances, designed to benchmark and improve object detection robustness under real-world haze conditions.
The results in \tablename~\ref{tab:hazydet} show that dehazing has a clear and model-dependent impact on detection accuracy. 
Methods with stronger structural and radiometric stability, such as DehazeFormer and PSMB-Net, yield notable gains of 6.2–9.6 AP at the 0.5 IoU threshold and 4.5–5.7 AP under the AP[0.5:0.95] metric, demonstrating their capability as detection-friendly dehazing solutions. 
SpA-GAN and LFD-Net with weaker haze removal or unstable textures reduce AP by 3–8 points, indicating that artifact-prone enhancement can degrade semantic cues essential for object detection. 

\begin{table}[t]
{
\centering
\scriptsize  
\caption{{Task-driven evaluation of detection-friendly dehazing on the HazyDet datasets.}}
{
\label{tab:hazydet}
\resizebox{\textwidth}{!}{%
\begin{tabular}{lcccccc}
\toprule
\multirow{2}{*}{\textbf{Methods}} 
& \multirow{2}{*}{\textbf{PSNR$ \uparrow$}} 
& \multirow{2}{*}{\textbf{SSIM$ \uparrow$}} 
& \multicolumn{2}{c}{\textbf{Trained on Clear Images}} 
& \multicolumn{2}{c}{\textbf{Trained on Dehazing Images}} \\
\cmidrule(lr){4-5} \cmidrule(lr){6-7}
& & &
\textbf{mAP@0.5$ \uparrow$} 
& \textbf{mAP@[0.5:0.95]$ \uparrow$} 
& \textbf{mAP@0.5$ \uparrow$} 
& \textbf{mAP@[0.5:0.95]$ \uparrow$} \\
\midrule
Faster RCNN~\cite{ren2016faster} & -- & -- & 55.4 & 39.5 & 68.0 & 48.1 \\
+ SpA-GAN~\cite{pan2020cloud} 
& 14.95 & 0.735 
& 47.3$_{\textcolor{blue}{-8.1}}$ 
& 33.2$_{\textcolor{blue}{-6.3}}$ 
& 63.1$_{\textcolor{blue}{-4.9}}$ 
& 43.6$_{\textcolor{blue}{-4.5}}$ \\
+ DCIL~\cite{zhang2022dense} 
& 16.03 & 0.772 
& 56.2$_{\textcolor{red}{+0.8}}$ 
& 39.8$_{\textcolor{red}{+0.3}}$ 
& 67.7$_{\textcolor{blue}{-0.3}}$ 
& 47.8$_{\textcolor{blue}{-0.3}}$ \\
+ DehazeFormer~\cite{song2023vision} 
& \textbf{23.27} & \textbf{0.886} 
& 61.6$_{\textcolor{red}{+6.2}}$ 
& 44.0$_{\textcolor{red}{+4.5}}$ 
& 68.6$_{\textcolor{red}{+0.6}}$ 
& 48.5$_{\textcolor{red}{+0.4}}$ \\
+ PSMB-Net~\cite{sun2023partial} 
& 15.37 & 0.806 
& \textbf{65.0}$_{\textcolor{red}{+9.6}}$ 
& \textbf{45.2}$_{\textcolor{red}{+5.7}}$ 
& \textbf{68.9}$_{\textcolor{red}{+0.9}}$ 
& \textbf{48.6}$_{\textcolor{red}{+0.5}}$ \\
+ LFD-Net~\cite{jin2023lfd} 
& 15.70 & 0.729 
& 51.9$_{\textcolor{blue}{-3.5}}$ 
& 36.8$_{\textcolor{blue}{-2.7}}$ 
& 63.5$_{\textcolor{blue}{-4.5}}$ 
& 44.6$_{\textcolor{blue}{-3.5}}$ \\
+ ASTA~\cite{cai2024additional} 
& 17.20 & 0.795 
& 54.0$_{\textcolor{blue}{-1.4}}$ 
& 38.7$_{\textcolor{blue}{-0.8}}$ 
& 67.7$_{\textcolor{blue}{-0.3}}$ 
& 47.7$_{\textcolor{blue}{-0.4}}$ \\
\bottomrule
\end{tabular}
}
}
}
\end{table}

{As shown in the right half of \tablename~\ref{tab:hazydet}, Under the setting of training on dehazed images and testing on dehazed images, the detector is fully domain matched to the dehazing outputs, which largely eliminates the performance degradation caused by dehazing induced appearance shifts such as contrast enhancement, texture sharpening, and color distribution changes. 
Using Faster R-CNN as the baseline, the detector achieves 68.0 mAP at 0.5 and 48.1 mAP at 0.5 to 0.95. 
With dehazing preprocessing, the mAP at 0.5 ranges from 63.1 to 68.9, corresponding to a change of minus 4.9 to plus 0.9 relative to the baseline, while the mAP at 0.5 to 0.95 ranges from 43.6 to 48.6, corresponding to a change of minus 4.5 to plus 0.5. 
Compared with the cross domain setting where the detector is trained on clear images and tested on dehazed images, the overall variation becomes much smaller, indicating that once the detector is adapted to the dehazing domain, the differences among dehazing methods contribute less to detection accuracy. This protocol therefore reflects an upper bound of detectability in the dehazing domain, rather than a strict guarantee of improved physical fidelity or radiometric consistency in the restored images.}

\begin{figure}[t]
    \centering
    \includegraphics[width=13.5cm]{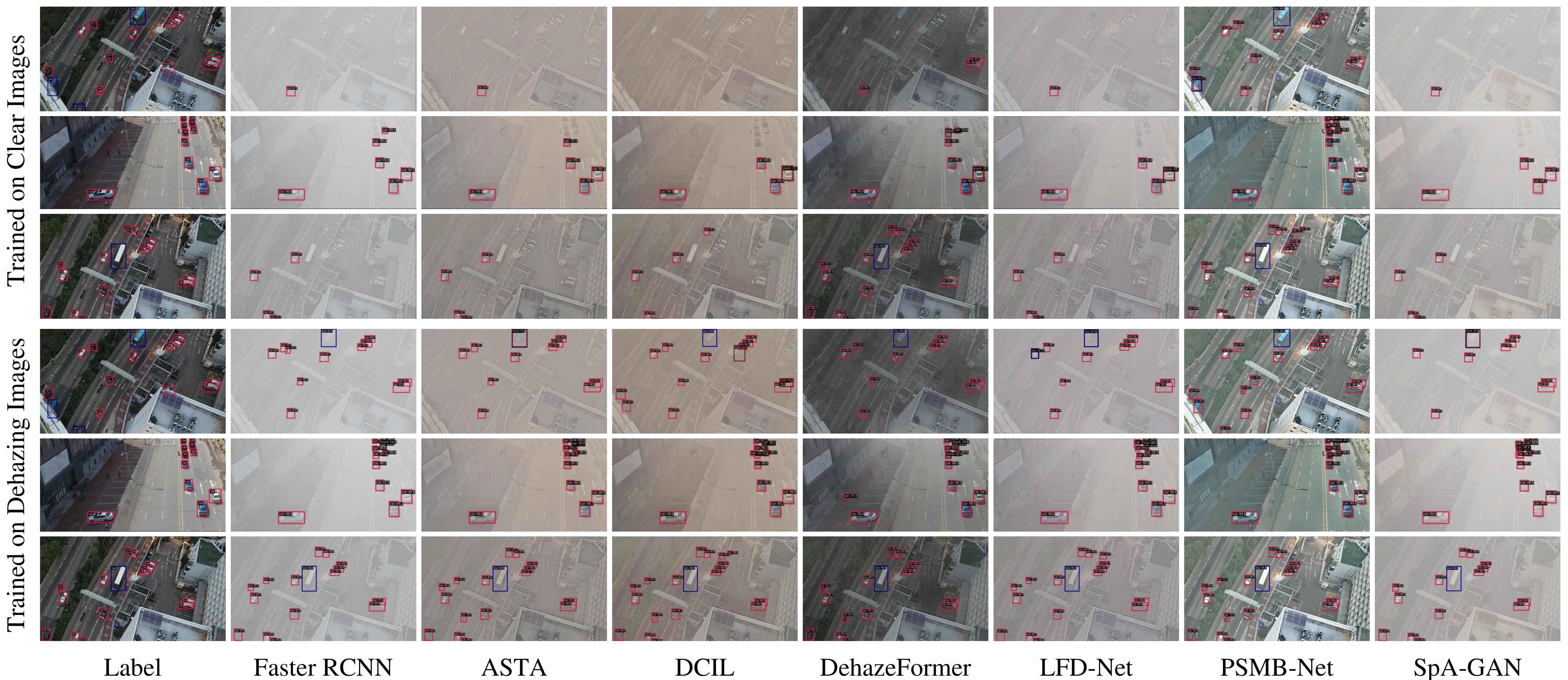}
    \caption{{Visual comparison of detection-friendly dehazing on the HazyDet datasets.}}
    \label{fig:Detection-friendly}
\end{figure}

The qualitative results in \figurename~\ref{fig:Detection-friendly} show that haze severely suppresses the detector’s ability to localize objects: Faster RCNN on hazy inputs misses small vehicles, produces imprecise bounding boxes, and fails in low-contrast road regions. 
After applying dehazing, the detection quality varies noticeably across methods. 
Transformer-based DehazeFormer and ASTA produce clearer structural boundaries and higher local contrast, enabling the detector to recover most missing vehicles and refine bounding boxes.
DCIL also enhances detectability, particularly in mid-size vehicle regions. 
SpA-GAN and LFD-Net methods either over-smooth details or introduce color shifts, resulting in incomplete or unstable detections despite improved visibility. 
{As can be seen from \figurename~\ref{fig:Detection-friendly}, the model trained on dehazed images has a better false alarm and false negative rates than the model trained on clear images.}
These findings from \tablename~\ref{tab:hazydet} and \figurename~\ref{fig:Detection-friendly} confirm that effective dehazing must not only improve visual quality but also preserve task-relevant structures to support reliable RS object detection.

\begin{table}[t]
\centering
\small
\caption{\cod{Task-driven evaluation of segmentation-friendly dehazing on the HazyUDD datasets.}}
\cod{
\label{tab:seg_metrics}
\resizebox{\textwidth}{!}{
\begin{tabular}{lccccc ccc}
\toprule
\multirow{2}{*}{\textbf{Methods}} 
& \multirow{2}{*}{\textbf{PSNR$\uparrow$}} 
& \multirow{2}{*}{\textbf{SSIM$\uparrow$}} 
& \multicolumn{3}{c}{\textbf{Trained on Clear Images}} 
& \multicolumn{3}{c}{\textbf{Trained on Dehazing Images}} \\
\cmidrule(lr){4-6}\cmidrule(lr){7-9}
& & 
& \textbf{mIoU$\uparrow$} & \textbf{Acc$\uparrow$} & \textbf{F$_1$-score$\uparrow$}
& \textbf{mIoU$\uparrow$} & \textbf{Acc$\uparrow$} & \textbf{F$_1$-score$\uparrow$} \\
\midrule
U-Net~\cite{ronneberger2015u}
& -- & --
& 14.87 & 70.03 & 24.64
& 30.95 & 80.14 & 43.98 \\
+ SpA-GAN~\cite{pan2020cloud}
& 13.45 & 0.639
& 13.65$_{\textcolor{blue}{-1.22}}$ & 68.43$_{\textcolor{blue}{-1.60}}$ & 23.06$_{\textcolor{blue}{-1.58}}$
& 13.62$_{\textcolor{blue}{-17.33}}$ & 74.17$_{\textcolor{blue}{-5.97}}$ & 21.54$_{\textcolor{blue}{-22.44}}$ \\
+ DCIL~\cite{zhang2022dense}
& 13.60 & 0.680
& 18.17$_{\textcolor{red}{+3.30}}$ & 73.05$_{\textcolor{red}{+3.02}}$ & 29.41$_{\textcolor{red}{+4.77}}$
& 31.42$_{\textcolor{red}{+0.47}}$ & 80.05$_{\textcolor{blue}{-0.09}}$ & 45.08$_{\textcolor{red}{+1.10}}$ \\
+ DehazeFormer~\cite{song2023vision}
& \textbf{21.02} & \textbf{0.872}
& \textbf{30.39}$_{\textcolor{red}{+15.52}}$ & \textbf{80.77}$_{\textcolor{red}{+10.74}}$ & \textbf{43.42}$_{\textcolor{red}{+18.78}}$
& 36.18$_{\textcolor{red}{+5.23}}$ & 82.83$_{\textcolor{red}{+2.69}}$ & 49.95$_{\textcolor{red}{+5.97}}$ \\
+ PSMB-Net~\cite{sun2023partial}
& 13.14 & 0.752
& 17.72$_{\textcolor{red}{+2.85}}$ & 73.60$_{\textcolor{red}{+3.57}}$ & 29.07$_{\textcolor{red}{+4.43}}$
& 26.76$_{\textcolor{red}{+4.19}}$ & 77.98$_{\textcolor{red}{+2.16}}$ & 40.33$_{\textcolor{red}{+3.65}}$ \\
+ LFD-Net~\cite{jin2023lfd}
& 13.11 & 0.651
& 14.61$_{\textcolor{blue}{-0.26}}$ & 71.08$_{\textcolor{red}{+1.05}}$ & 25.15$_{\textcolor{red}{+0.51}}$
& 29.71$_{\textcolor{blue}{-1.24}}$ & 78.79$_{\textcolor{blue}{-1.35}}$ & 43.46$_{\textcolor{blue}{-0.52}}$ \\
+ ASTA~\cite{chen2025tokenize}
& 12.79 & 0.641
& 15.94$_{\textcolor{red}{+1.07}}$ 
& 72.54$_{\textcolor{red}{+2.51}}$ 
& 25.27$_{\textcolor{red}{+0.63}}$
& 28.06$_{\textcolor{blue}{-2.89}}$ & 79.32$_{\textcolor{blue}{-0.82}}$ & 40.28$_{\textcolor{blue}{-3.70}}$ \\
\bottomrule
\end{tabular}
}
}
\end{table}

\begin{figure}[t]
    \centering
    \includegraphics[width=13.5cm]{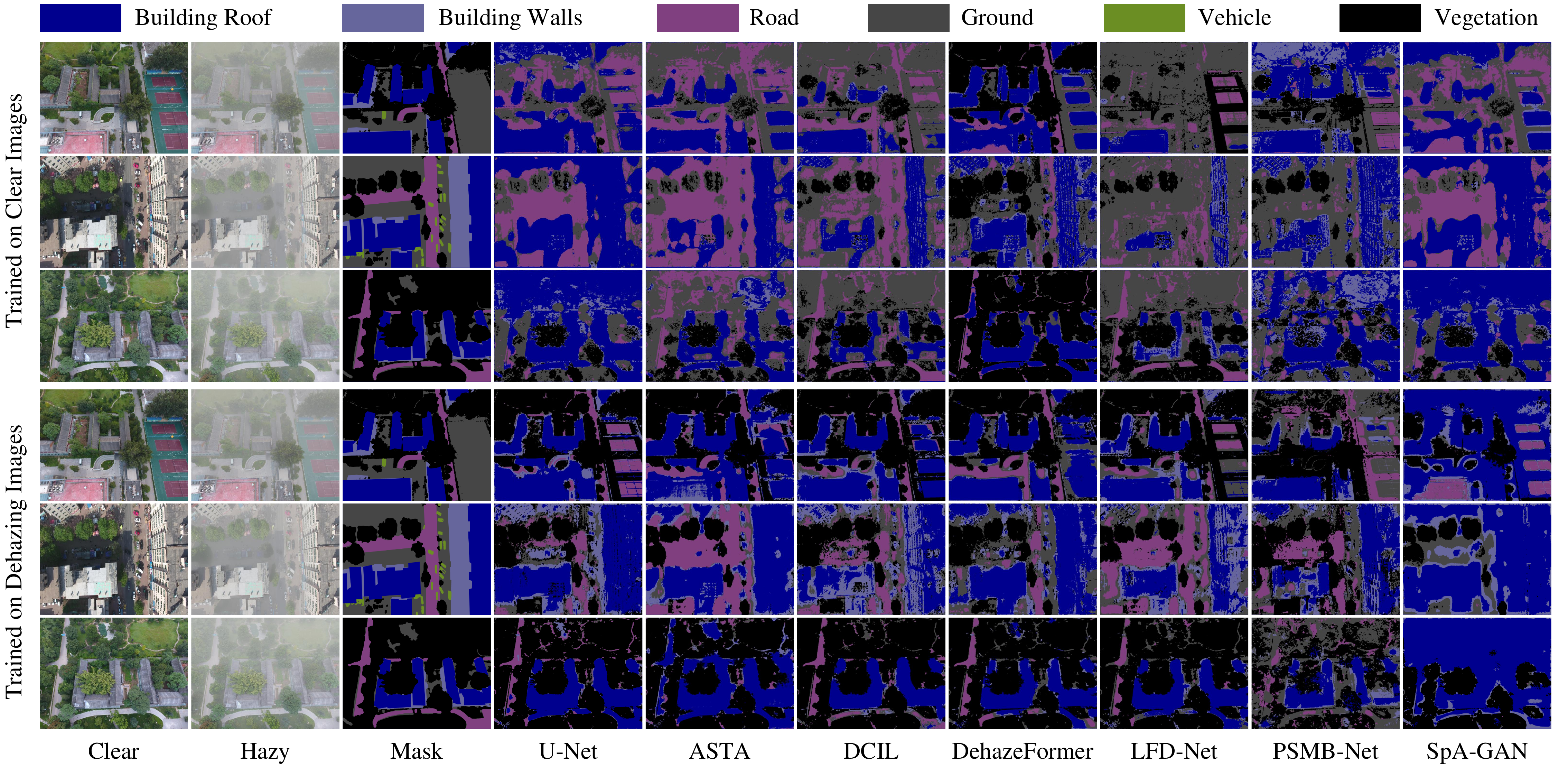}
    \caption{{Visual comparison of segmentation-friendly dehazing on the HazyUDD datasets.}}
    \label{fig:Segmentation-friendly}
\end{figure}

% 分析
\textbf{Segmentation-friendly Dehazing Evaluation.}
For task-driven evaluation, we additionally adopt U-Net~\cite{ronneberger2015u} {training on clear images} and report performance changes before and after applying different dehazing models.
The HazyUDD datasets~\cite{wang2025dehaze} are a synthetic hazy extension of the UDD6 UAV~\cite{chen2018large} , adding light and heavy fog to the original scenes via depth-based atmospheric scattering for robust UAV semantic segmentation evaluation.
As shown in \tablename~\ref{tab:seg_metrics} and \figurename~\ref{fig:Segmentation-friendly}, effective dehazing provides substantial benefits for downstream segmentation. 
DehazeFormer yields the greatest improvement, increasing mIoU from 14.87 to 30.39 and boosting the F1-score by 18.78 points, demonstrating that removing haze-induced structural ambiguity significantly enhances semantic boundary recovery. 
DCIL and PSMB-Net also consistently achieved mIoU scores of over 2.85 and F1 scores of 4.43.
{As shown in the right half of \tablename~\ref{tab:seg_metrics}, when training and testing the segmentation model on dehazed images, the domain gap is largely removed and the results reflect the segmentation upper bound in the dehazing domain. 
DehazeFormer achieves the best performance with 36.18 mIoU, 82.83 Acc, and 49.95 F1, outperforming the U-Net baseline by 5.23 mIoU, 2.69 Acc, and 5.97 F1. 
In contrast, SpA-GAN causes severe degradation, indicating that visually improved dehazing does not necessarily produce segmentation-friendly representations, and preserving region semantics and boundary cues is more critical than pixel-level fidelity.}

\figurename~\ref{fig:Segmentation-friendly} shows that haze severely reduces boundary sharpness and class separability, causing the U-Net to miss thin structures such as roads and vegetation contours. 
After dehazing, DehazeFormer and ASTA restore clearer textures and more stable illumination, enabling the segmentation network to recover finer region boundaries and correct previously confused classes. 
PSMB-Net and SpA-GAN often introduce color bias or over-smoothing, leading to fragmented masks and inaccurate region shapes. 
{As can be seen from the comparison between the top and bottom of \figurename~\ref{fig:Segmentation-friendly}, the model trained on the dehazed image can segment more object pixels.}
Overall, segmentation-friendly dehazing requires radiometrically stable restoration, as methods preserving structural details produce significantly more coherent and complete segmentation outputs.

{
Looking ahead, the recent studies of remote sensing large models~\cite{hu2025rsgpt,yu2024metaearth,liu2025text2earth} demonstrate that treating dehazing as an independent and universally applicable pre-processing step is increasingly insufficient. 
While dehazing can reshape image appearance and improve visual quality, its contribution to downstream tasks is highly task dependent and closely tied to the training domain of subsequent models. 
This observation motivates a shift away from isolated restoration pipelines toward integrated, foundation-model-driven frameworks, in which restoration, representation learning, and task reasoning are jointly optimized. 
In such unified paradigms, dehazing is no longer an explicit front-end operation, but an adaptive and implicit capability embedded within large-scale remote sensing models, enabling more robust, task-aware, and physically consistent scene understanding across diverse atmospheric conditions.
}

\begin{figure}[t]
	\centering 
	\includegraphics[width=11cm]{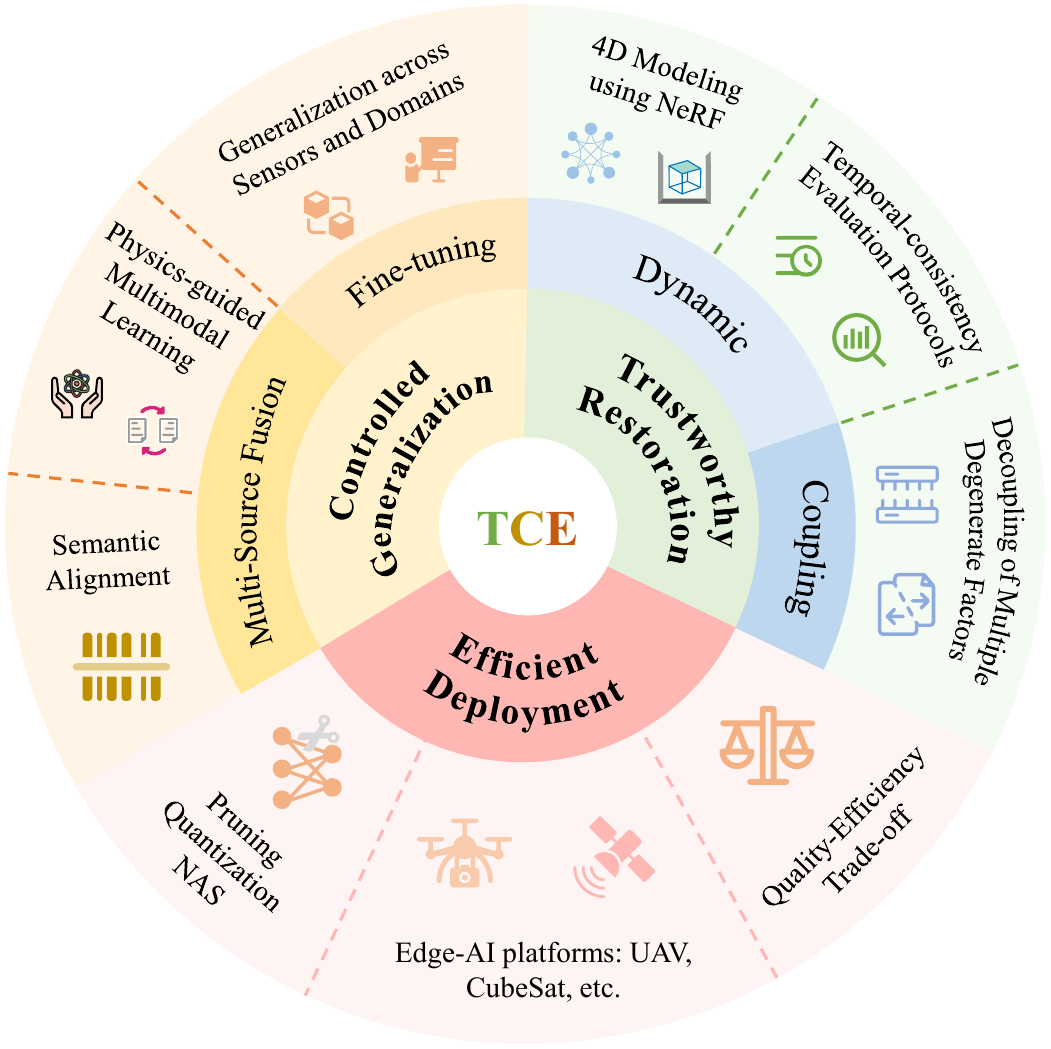}
	\caption{
    Future prospects for RSIs dehazing: Trustworthy, controllable, and efficient (TCE) remote sensing dehazing system.
    }
	\label{Fig.Directions}
\end{figure}

\section{Current Challenges and Future Prospects}
\label{sec.challenges}
Remote sensing image dehazing has progressed from heuristic contrast enhancement to physics-informed deep networks and diffusion-based generative dehazing~\cite{yin2025remote}. 
However, current approaches still face critical limitations, including oversimplified and static degradation assumptions, insufficient coupling across spectral, thermal, and temporal modalities, and a high computational cost that restricts large-scale or edge-side deployment. 
These constraints indicate that the next generation of remote sensing image dehazing must move beyond isolated algorithmic advances toward an integrated paradigm. 
As illustrated in \figurename~\ref{Fig.Directions}, we envision three convergent research trajectories: trustworthy spatiotemporal atmospheric reconstruction, controllable multimodal generalization guided by physical principles, and efficient model deployment for UAV and CubeSat platforms. 
Together, these directions outline a unified trustworthy, controllable, and efficient (TCE) system that balances interpretability, cross-domain scalability, and real-world operational efficiency.
{As shown in \tablename~\ref{tab:TCE}, we have specified concrete and executable research directions as well as clearly defined and measurable goals for the TCE system. 
}

\subsection{Trustworthy: From 2D Static to 4D Spatiotemporal Atmospheric Reconstruction}
Conventional dehazing networks assume stationary degradation governed by simplified scattering models, overlooking the temporal and volumetric dynamics of real atmospheric phenomena.
In real-world environments, degradation is often non-stationary, manifesting as overlapping thin clouds and heavy haze~\cite{li2018haze,jiang2021deep}, 
time-varying phenomena like rain–snow transitions, or extreme occlusions exceeding 90\%~\cite{ma2023cloud}. 
Additionally, RS scenes often exhibit coupled degradations, such as cloud–haze coexistence or the superposition of terrain texture and cloud shadows, which further complicate restoration~\cite{quan2024density,shu2025restore}. 
Traditional single-task frameworks cannot disentangle these interferences, leading to detail confusion and spectral distortions. 

Future research should reconceptualize RSIs dehazing as a \textbf{4D spatiotemporal inverse scattering problem}, in which haze, fog, and cloud layers evolve continuously over time and depth. 
By integrating volumetric rendering and neural radiative field modeling~\cite{mildenhall2021nerf,zuo2024nerf,chen2024dehazenerf}, atmospheric scattering can be represented as a differentiable field, enabling temporally consistent restoration and interpretable physical reconstruction~\cite{cheng2024continual,zhang2024jointly,zhang2025unified}. 
{
In concrete research actions of \tablename~\ref{tab:TCE}}.
{\textbf{First}, atmospheric degradation should be modeled through learnable volumetric or depth-aware representations, where scattering is parameterized as a continuous but low-dimensional field evolving over time and space. 
By integrating differentiable radiative transfer operators or neural radiative field formulations, atmospheric effects can be embedded as explicit physical processes instead of implicit feature correlations. 
\textbf{Second}, trustworthy restoration requires explicit disentanglement of coupled degradations. 
Multi-task learning frameworks can be designed to jointly estimate atmospheric parameters, surface reflectance, and degradation types (\textit{e.g.}, haze, thin cloud, cloud shadow), while cross-degradation knowledge distillation enables shared physical priors to be transferred across related degradation modes. 
This prevents semantic confusion and spectral distortion caused by overlapping atmospheric and surface effects. \textbf{Third}, instead of assuming continuous observations, models should enforce temporal coherence of physical variables across sporadic satellite passes, enabling physically plausible interpolation and stabilization of restoration results without relying on unrealistic temporal sampling density.
}

{
Measurable objectives include: limited performance degradation across datasets and across sensors; semantic consistency when adding, removing, or replacing auxiliary modes; and invariance in estimating physical variables over regions with similar atmospheric conditions. 
A controllable system should exhibit stable and predictable behavior under modal and regional perturbations, demonstrating true generalization ability rather than optimization for a specific dataset.}

\begin{table}[t]
\centering
\caption{{TCE framework with concrete research actions and measurable goals.}}
{
\label{tab:TCE}
\scriptsize
\setlength{\tabcolsep}{2.5pt}
\renewcommand{\arraystretch}{1.15}
\resizebox{\linewidth}{!}{
\begin{tabular}{>{\centering\arraybackslash}m{0.14\linewidth}
                >{\raggedright\arraybackslash}m{0.43\linewidth}
                >{\raggedright\arraybackslash}m{0.43\linewidth}}
\hline
\textbf{Dimension} & \textbf{Concrete Research Actions} & \textbf{Clear and Measurable Goals} \\
\hline
\textbf{Trustworthy} &
\begin{itemize}[leftmargin=*, nosep]
  \item Reliable parameter estimation based on differential equations, including transmittance, atmospheric light, and scattering intensity;
  \item Radiation-consistency regularization via knowledge distillation;
  \item Uncertainty modeling for extreme haze regions using depth estimation.
\end{itemize}
&
\begin{itemize}[leftmargin=*, nosep]
  \item Accuracy and stability of estimated physical variables across datasets and haze levels;
  \item Radiometric metrics: SAM$\downarrow$, ERGAS$\downarrow$, QNR$\uparrow$, HIST$\uparrow$;
  \item Reduced color bias and spectral distortion under real-world conditions.
\end{itemize}
\\ \hline
\textbf{Controllable} &
\begin{itemize}[leftmargin=*, nosep]
  \item Physics-aware latent disentanglement separating shared physical factors and modality-private appearance;
  \item Cross-modal semantic anchoring using high-level prior knowledge from large multimodal models such as CLIP;
  \item Hierarchical and region-adaptive multimodal alignment based on segmenting anything model;
  \item Self-supervised learning and test-time adaptation guided by SAR geometric location cues.
\end{itemize}
&
\begin{itemize}[leftmargin=*, nosep]
  \item Performance decay rate under cross-dataset and cross-sensor transfer;
  \item Semantic consistency when modalities are added, removed, or substituted;
  \item Cross-domain invariance of estimated physical variables under similar atmospheric conditions.
\end{itemize}
\\ \hline
\textbf{Efficient} &
\begin{itemize}[leftmargin=*, nosep]
  \item Lightweight Mamba architectures reducing redundant feature learning;
  \item Parameter-efficient multimodal fusion via reparameterization;
  \item Efficient diffusion-generative models with reduced inference steps;
  \item Hardware-aware operator optimization for edge deployment.
\end{itemize}
&
\begin{itemize}[leftmargin=*, nosep]
  \item Model size, FLOPs, and inference latency;
  \item Scalability to large-scale or wide-swath RS imagery;
  \item Performance--efficiency trade-off (accuracy/physical consistency per computation cost).
\end{itemize}
\\ \hline
\end{tabular}
}}
\end{table}

\subsection{Controllable: Physics-Guided Multimodal Generalization}

While multi-source RSIs modalities such as SAR~\cite{wang2025mt_gan,gu2025hpn,tu2025cloud,liu2025enhancing}, thermal~\cite{fang2025guided,xu2025efficient}, and hyperspectral imaging~\cite{sun2025bidirectional,fu2024hyperdehazing,liao2024joint,luo2025deep} offer valuable complementary information, current methods often rely on shallow concatenation without physically meaningful alignment. 
This leads to poor synergy between modalities and restricts generalization in complex or unseen scenarios. 
Furthermore, the lack of large-scale paired datasets under diverse atmospheric conditions~\cite{ge2025rsteller}, 
especially for rare weather events (\textit{e.g.}, polar snowfall) or emerging sensor combinations (\textit{e.g.}, hyperspectral–LiDAR), 
limits the applicability of supervised learning approaches~\cite{li2024multi}. 

The next generation of RSIs restoration should adopt a \textbf{physics-guided controllable multimodal generation} paradigm, integrating physical priors (\textit{e.g.}, Maxwell’s equations, radiative transfer constraints) into unified large-model architectures. 
Such models can embed optical–SAR~\cite{meraner2020cloud,huang2020single,han2023former}, 
RGB–Thermal~\cite{wang2015infrared,erlenbusch2023thermal}, and 
RGB–hyperspectral~\cite{mehta2020hidegan,xu2023aacnet,chen2025bidirectional} modalities into a shared latent space via CLIP-style contrastive alignment, augmented by physics-constrained adapters that preserve spectral fidelity and structural coherence. 
To enhance scalability, self-supervised~\cite{xu2025sdcluster} and meta-learning~\cite{hu2023cross,qin2023bi,tang2025meta} strategies can facilitate few-shot adaptation under data-scarce or rare weather conditions~\cite{zhao2021domain,hu2022switch,zheng2023multiple,wu2023domain}. 
{As shown in \figurename~\ref{Fig.Directions} and \tablename~\ref{tab:TCE}, the concrete research actions involve: \textbf{First}, physical-aware latent disentanglement technology should be adopted to separate shared physical factors from modality-specific appearance components, thereby preventing semantic drift during multimodal fusion. 
\textbf{Second}, cross-modal semantic anchoring mechanisms based on large multimodal models such as CLIP should utilize sensor-invariant physical quantities or high-level semantic priors to constrain latent representations, avoiding uncontrolled semantic transfer. 
\textbf{Then}, hierarchical and region-adaptive alignment strategies based on segmenting anything model can be employed, where low-level fusion is guided by physical consistency, and high-level fusion is restricted to semantically reliable regions. 
\textbf{Finally}, sensor-aware normalization and physics-guided self-supervised and test-time adaptation techniques should be combined to decouple atmospheric effects from acquisition-specific features and recalibrate the model in unseen domains.
}

{Clear and measurable goals include: limited performance degradation under cross-dataset and cross-sensor transfer; semantic consistency when auxiliary modalities are added, removed, or replaced; and the invariance of estimated physical variables within regions with similar atmospheric conditions. A controllable system should exhibit stable and predictable behavior under modality and regional perturbations, demonstrating true generalization ability rather than optimization for specific datasets.
}
% -------------- 

\subsection{Efficient: Lightweight Dehazing for Edge AI}
Although recent models based on
Transformers~\cite{chi2023trinity,li2025decloudformer,song2023learning} and 
diffusion architectures~\cite{liu2025effective,huang2024diffusion,xiong2024rshazediff} have shown impressive dehazing performance, 
they are often computationally intensive and impractical for deployment on edge devices or real-time satellite platforms. 
These limitations are particularly critical in time-sensitive applications, such as onboard UAV sensing. 

A future direction lies in constructing \textbf{lightweight practical dehazing models} that jointly optimize physical fidelity and computational efficiency in \figurename~\ref{Fig.Directions}. 
This can be achieved by embedding physical priors into neural architecture search (NAS)~\cite{li2025towards,popov2024remote,luo2025large}, pruning~\cite{fu2025iterative,lv2023pruning,shinde2025model}, and quantization objectives, thereby ensuring that learned architectures preserve radiometric consistency while minimizing latency and power consumption~\cite{zhou2025lightweight,hu2025lightweight}. 
{
As listed in \tablename~\ref{tab:TCE}, concrete research actions include:
Parameter-efficient multimodal fusion mechanisms based on reparameterization should replace high-dimensional feature concatenation or excessive attention mechanisms. 
It is necessary for generative methods to adopt designs with high inference efficiency, such as simplifying stepwise diffusion or using a hybrid deterministic-probabilistic framework, to balance restoration quality and computational cost. 
In addition, hardware-aware convolution operator optimization should be considered to facilitate edge and on-board deployment.
}

{In clear and measurable goals, efficiency can be quantitatively evaluated through model size, the number of FLOPs, and inference latency under standardized input resolutions. Other goals include scalability to high-resolution remote sensing images, as well as a good performance-efficiency trade-off, measured by restoration quality or physical properties.}

In summary, RSIs dehazing is transitioning from static correction to trustworthy dynamic reconstruction, from single-modality to physics-guided multimodal controllable generation, 
and from high-cost offline inference to efficient edge intelligence. 
These trajectories redefine the methodological frontier, establishing a unified roadmap toward physically grounded, data-efficient, and operationally deployable RSIs restoration systems.

\section{Conclusions}
\label{sec.conclusion}  
This review presents a systematic review of recent developments in RSIs dehazing, covering a wide spectrum of approaches ranging from traditional enhancement and physical model-based techniques to deep learning paradigms, including CNNs, GANs, Transformers, and diffusion models. 
Through extensive benchmarking on datasets, we highlight the evolution and performance of representative algorithms, providing insight into their strengths and limitations.
CNN-based methods offer strong local feature representation and efficiency, Transformer-based models introduce global context modeling, and diffusion-based approaches demonstrate superior stability across diverse haze levels through iterative refinement.

Despite notable progress, several challenges remain unresolved, including robust modeling of complex and dynamic degradation, effective utilization of multimodal and temporal data, deployment of lightweight models for real-time applications, and ensuring restoration reliability in operational scenarios. 
Future research is encouraged to explore emerging directions such as neural radiance fields for dynamic reconstruction, physics-guided cross-modal fusion strategies, and few-shot adaptive restoration frameworks. 
Furthermore, bridging the gap between academic advances and large-scale engineering deployment remains an essential frontier.
In conclusion, this review aims to provide a clear roadmap for researchers and practitioners by consolidating existing knowledge, identifying key bottlenecks, and proposing forward-looking directions for advancing RSIs restoration technologies.

\section*{CRediT authorship contribution statement}
Methodology, H.Z., X.L., Z.Z., and J.Y.; 
Supervision, Z.Z., X.W. and C.T.; 
Writing—Original Draft, H.Z., X.L., C.L., and Z.Z.; 
Writing—Review and Editing, H.Z., X.L., J.Y., C.L., Y.Y. and D.X;
Visualization, X.L., J.Y., Z.Z., D.X., C.L., and X.W.;
All authors have read and agreed to the published version of the manuscript.

\section*{Declaration of competing interest}
The authors declare that they have no known competing financial interests or personal relationships that could have appeared to influence the work reported in this paper.

\section*{Acknowledgments}
This research was supported in part by Funded by Basic Research Program of Jiangsu (BK20251623);
the Jiangsu Funding Program for Excellent Postdoctoral Talent (2025ZB126);
the China Postdoctoral Science Foundation (2024M761178, 2025M771739); 
the Postdoctoral Fellowship Program of CPSF (GZC20241332, GZC20251207);
the Wuxi Science and Technology Development Fund Project (K20252022);
the National Natural Science Foundation of China (62401429, 62502538, U25A20527); 
the Oil \& Gas Major Project (2025ZD1404600);
the Fundamental Research Funds for the Central Universities (JUSRP202501074);
the Science Foundation of China University of Petroleum (2462025YJRC006).
We sincerely thank the editors and the anonymous reviewers for their insightful comments on and valuable improvements to our manuscript.

{
\small
\bibliographystyle{elsarticle-num-names}
\bibliography{ref}
}

\end{document}